\PassOptionsToPackage{sort}{natbib}
\PassOptionsToPackage{dvipsnames}{xcolor}

\documentclass{article} 
\usepackage{iclr2027_conference,times}

\usepackage{amsmath,amsfonts,bm}
\usepackage{upgreek}

\def\eqref#1{equation~\ref{#1}}

\def\1{\bm{1}}

\def\rc{{\textnormal{c}}}

\def\rn{{\textnormal{n}}}

\def\rr{{\textnormal{r}}}

\def\rw{{\textnormal{w}}}

\def\rz{{\textnormal{z}}}

\def\rvepsilon{{\bm{\upepsilon}}}

\def\rvc{{\mathbf{c}}}

\def\rvn{{\mathbf{n}}}

\def\rvw{{\mathbf{w}}}
\def\rvx{{\mathbf{x}}}
\def\rvy{{\mathbf{y}}}
\def\rvz{{\mathbf{z}}}

\def\vzero{{\bm{0}}}

\def\vmu{{\bm{\mu}}}

\def\ve{{\bm{e}}}

\def\vu{{\bm{u}}}
\def\vv{{\bm{v}}}

\def\vx{{\bm{x}}}

\def\vz{{\bm{z}}}

\def\mA{{\bm{A}}}

\def\mI{{\bm{I}}}

\def\mV{{\bm{V}}}
\def\mW{{\bm{W}}}
\def\mX{{\bm{X}}}

\def\mSigma{{\bm{\Sigma}}}

\DeclareMathAlphabet{\mathsfit}{\encodingdefault}{\sfdefault}{m}{sl}
\SetMathAlphabet{\mathsfit}{bold}{\encodingdefault}{\sfdefault}{bx}{n}

\def\gA{{\mathcal{A}}}

\def\gC{{\mathcal{C}}}

\def\gI{{\mathcal{I}}}

\def\gM{{\mathcal{M}}}
\def\gN{{\mathcal{N}}}

\def\gS{{\mathcal{S}}}

\def\sR{{\mathbb{R}}}

\newcommand{\E}{\mathbb{E}}

\newcommand{\R}{\mathbb{R}}

\newcommand{\Cov}{\mathrm{Cov}}

\DeclareMathOperator{\Tr}{Tr}

\def\sobj{\gS_\textnormal{obj}}
\def\sbg{\gS_\textnormal{bg}}
\def\dino{DINOv2\xspace}

\makeatletter
\DeclareRobustCommand\onedot{\futurelet\@let@token\@onedot}
\def\@onedot{\ifx\@let@token.\else.\null\fi\xspace}
\def\eg{\emph{e.g}\onedot}

\def\ie{\emph{i.e}\onedot}

\makeatother

\usepackage{graphicx}
\usepackage{booktabs}
\usepackage{amssymb}
\usepackage{multirow}
\usepackage{xspace}
\usepackage{subcaption}
\usepackage[dvipsnames]{xcolor}
\usepackage[backref]{hyperref}
\usepackage{url}
\usepackage{amsthm}
\usepackage{makecell}
\usepackage[capitalize]{cleveref}
\newtheorem{theorem}{Theorem}
\newtheorem{proposition}{Proposition}
\usepackage{empheq}
\usepackage{tikz}
\usetikzlibrary{patterns, patterns.meta, decorations.pathreplacing}
\usepackage{wrapfig}
\newcommand{\cdashedline}{%
    \noalign{\vskip 1.35pt}
    \multispan{3}{\cleaders\hbox to 0.5em{\hss-\hss}\hfill}\cr%
    \noalign{\vskip 1.35pt}
}
\newcommand{\cdashedlinelong}{%
    \noalign{\vskip 1.35pt}
    \multispan{4}{\cleaders\hbox to 0.5em{\hss-\hss}\hfill}\cr%
    \noalign{\vskip 1.35pt}
}

\hypersetup{
  colorlinks=true,
  linkcolor=red,
  citecolor=Blue,
  urlcolor=magenta,
}

\crefname{section}{Sec.}{Secs.}
\Crefname{section}{Section}{Sections}
\crefname{table}{Tab.}{Tabs.}
\Crefname{table}{Table}{Tables}

\DeclareRobustCommand{\brblacktriangle}{%
  \mathrel{\tikz[baseline=0.1ex] \fill (0,0) -- (1.2ex,0) -- (1.2ex,1.2ex) -- cycle;}%
}
\DeclareRobustCommand{\tlblacktriangle}{%
  \mathrel{\tikz[baseline=0.1ex] \fill (0,0) -- (0,1.2ex) -- (1.2ex,1.2ex) -- cycle;}%
}
\DeclareRobustCommand{\diagband}{%
  \mathrel{\tikz[baseline=0.1ex] \fill (0,0.5ex) -- (0.7ex,1.2ex) -- (1.2ex,1.2ex) -- (1.2ex,0.7ex) -- (0.5ex,0) -- (0,0) -- cycle;}%
}

\definecolor{invblue}{HTML}{C1E5F5}
\definecolor{alignedgreen}{HTML}{D9F2D0}
\definecolor{diagpurple}{HTML}{F2CFEE}

\newcommand{\regaligned}{\textcolor{ForestGreen}{$\brblacktriangle$}}
\newcommand{\regunexp}{\textcolor{Purple}{$\diagband$}}
\newcommand{\reginverted}{\textcolor{RoyalBlue}{$\tlblacktriangle$}}

\newcommand{\projecturl}{https://ronensc.github.io/Explainability-Illusion}

\crefname{appendix}{App.}{Apps.}
\Crefname{appendix}{Appendix}{Appendices}

\AddToHook{cmd/appendix/before}{%
    \crefalias{section}{appendix}%
    \crefalias{subsection}{appendix}
}

\newcommand{\beginsupplement}{%
    \renewcommand\thefigure{S\arabic{figure}}    
    \setcounter{figure}{0}
    \renewcommand\thetable{S\arabic{table}}    
    \setcounter{table}{0}  
    \renewcommand\theequation{S\arabic{equation}}    
    \setcounter{equation}{0}
}

\title{The Explainability Illusion: Must ViTs ~~~~~~~~~  Exhibit Patch-Decoding Interpretability?}

\author{Ronen Schaffer\thanks{Denotes equal contribution.},~ Hila Manor\footnotemark[1] ~ \& Tomer Michaeli\\
Faculty of Electrical and Computer Engineering\\
Technion -- Israel Institute of Technology\\
\texttt{\{ronens@campus,hila.manor@campus,tomer.m@ee\}.technion.ac.il}
}

\iclrfinalcopy 
\begin{document}

\maketitle

\begin{abstract}

Transformers are widely assumed to exhibit token-decoding explainability. In particular, in vision transformer (ViT) classifiers, decoding individual patch tokens has become a common diagnostic tool due to its tendency to yield human‑interpretable outputs. In light of the increasing popularity of this approach, an important question is whether token‑decoding explainability is an inherent property of transformers that should be expected to hold across architectures, scales, and training objectives, or merely an artifact of some implicit bias. In this paper we systematically study token‑decoding explainability across diverse ViT settings, spanning architectures, scales, and training objectives, using both controlled experiments and real-world models. We find that while some models indeed exhibit aligned and interpretable token decodings, others exhibit either inverted explainability maps or a complete lack of correlation with the image semantics. As we show, these tendencies toward different explainability behaviors are not intrinsic to the architecture and can vary arbitrarily with changes in the optimization landscape. We provide theoretical support for these observations by demonstrating a setting where a single fixed toy transformer architecture can provably implement the optimal classification rule in multiple ways, which differ in their patch-decoding explainability patterns. Our results show that token‑decoding explainability is not an inherent property of trained transformers, but rather an outcome contingent on the problem setting and thus should not be a-priori assumed to hold. 
Code can be found in our project \href{\projecturl}{website}.
\end{abstract}

\section{Introduction}

Transformers~\citep{vaswani2017attention}, originally presented in the context of natural language processing, have proven powerful across various domains, including images, audio, and video~\citep{dosovitskiy2021an,gong21b_interspeech,arnab2021vivit,yun2019graph}, achieving state-of-the-art performance on countless tasks~\citep{radford2021learning,peebles2023scalable,kirillov2020end,zheng2021rethinking,li2019neural}. Following their growth in popularity, significant attention has been devoted to mechanistically understanding their internal representations.

One of the simplest methods for explainability in transformer language models is Logit Lens~\citep{nostalgebraist2020logitlens}.
This cornerstone approach interprets each intermediate hidden state of the model by projecting it back into the model's vocabulary space. This is done by multiplying the token's embedding by the unembedding matrix, which is normally applied only to the tokens of the final layer. 
Logit Lens has been widely adopted for exposing the evolution of predictions across layers in large language models (LLMs)~\citep{pal2023future,prakash2023layered,wendler2024llamas,li2024understanding, patil2024can,wang2025logitlens4llms,wu2026your}, and inspired numerous derivative approaches for decoding intermediate tokens~\citep{belrose2023eliciting,din2024jump,sakarvadia2023attention}.
It has also expanded to other domains, including images~\citep{joseph2025prisma, neo2025towards}, audio~\citep{yang2025audiolens,glazer2026beyond}, and video~\citep{zhang2025circuitprobe,kim2026map}.

\begin{figure}
    \centering
    \includegraphics[width=\linewidth]{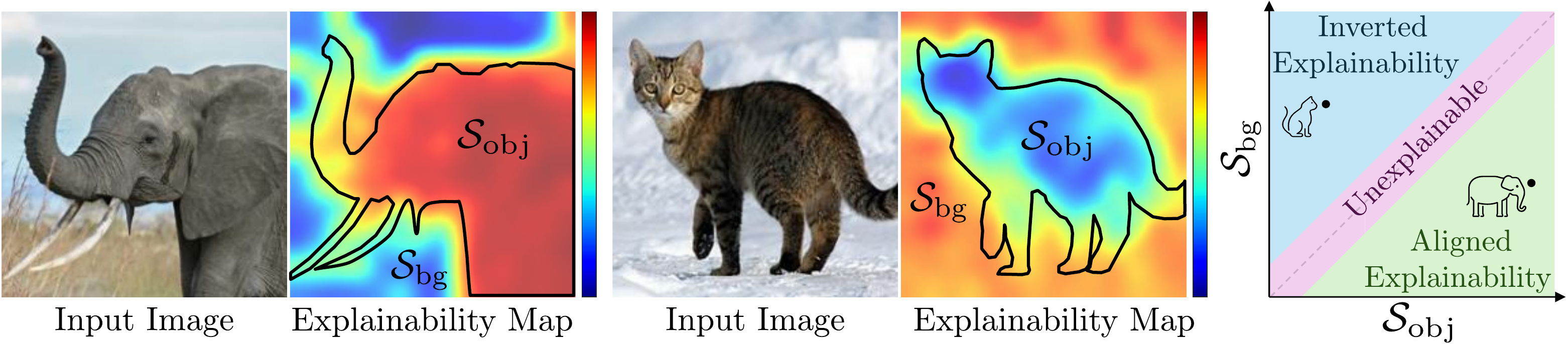}
    \caption{\textbf{Explainability regimes.} 
    For an input image, token-decoding of patches produces explainability maps, where each patch-token receives some score $\gS$. 
    The relation between the average score of object tokens, $\sobj$ (tokens inside the black contour), and the average score of background tokens, $\sbg$ (tokens outside the black contour), determine the explainability of a classified input image in a layer:
    \regaligned~aligned, \regunexp~unexplainable, or \reginverted~inverted.
    Here, the elephant is aligned and the cat is inverted.
    }
    \label{fig:explainable_diag}
\end{figure}

In light of the widespread adoption of these techniques, it seem natural to ask whether we can universally trust the validity of patch-token decoding explainability. Much of the current literature assumes that intermediate representations consistently map directly to interpretable concepts~\citep{joseph2025prisma,vilas2023analyzing,jiang2025interpreting,neo2025towards}. 
For instance, it is commonly expected that when interpreting the classification result of a vision transformer (ViT) on an input image of an elephant, the patch-tokens corresponding to the spatial location of the elephant would exhibit a higher logit for the ``elephant'' class than to any other class. This is illustrated in Fig.~\ref{fig:explainable_diag}, which shows a map of the ``elephant'' logit across all patch tokens of the image. 
Several recent works noted that this might not always be the case, observing that in some transformers token-decoding maps can look inverted~\citep{li2023clipsurgery,kang2025isclipideal,bousselham2024grounding}.
Yet, these observations have 
been treated as a localized effect to be engineered away.

In this paper, we show that there is no guarantee that transformers exhibit token-decoding explainability.
We begin by analyzing dozens of pre-trained ViTs, showing they break the common assumption of \emph{aligned}-explainability. Instead, many models exhibit either \emph{inverted}-explainability or completely \emph{unexplainable} intermediate representations. For example, when applying patch-token decoding explainability techniques to prominent contrastive vision-language models (VLMs) such as CLIP~\citep{radford2021learning} and SigLIP~\citep{zhai2023sigmoid}, the explainability maps are typically inverted. 
Next, through a series of controlled synthetic experiments, we expose how different conditions cause the same architecture to change its explainability behavior. We observe that different tasks and data distributions induce implicit biases toward different explainability regimes. Those biases do not deterministically determine the explainability behavior, however they control the probability of a training run to converge to either an aligned- or an inverted-explainability solution.   

Finally, we prove that aligned-explainability cannot be an inherent property of all transformers.
Specifically, we present a simple classification task for which the optimal classification rule admits an implementation by a toy, single layer transformer architecture in infinitely many ways. We prove that among the sets of weights that implement the optimal solution, some correspond to an aligned-explainable model and some to an inverted-explainable model. 

Our findings demonstrate that the behavior and explainability of decoded representations can vary widely depending on the optimization landscape, from the task, to the training loss, and even the data distribution itself.
This implies that token-decoding explainability is not a fundamental property of transformers, but merely a fragile byproduct of their training conditions.
As such, when training a transformer on a new task, it cannot be a-priori assumed that zero-shot decoding of its internal states would reveal any meaningful or localized information about the semantics encoded in those tokens.

\section{Related work}

\vspace{-4pt}
\paragraph{Explainability techniques for ViTs.}
Explainability methods broadly fall into two families.
The first attributes the ViT's prediction to the input image patches.
Early techniques aggregated attention across layers for standard transformers~\citep{abnar2020quantifying} and were subsequently adapted to ViTs~\citep{dosovitskiy2021an}.
Later methods propagated relevance through self-attention in both vision and language models~\citep{chefer2021transformer,ali2022xai,achtibat2024attnlrp}. Other approaches rely on gradient-based weighting of attention maps or features~\citep{chefer2021gae,zhao2024gradient}.
However, most attribution techniques are computationally intensive, requiring, \eg, heavy backward passes to compute gradients.
The second family maps intermediate tokens directly into the model's output space.
Logit Lens~\citep{nostalgebraist2020logitlens} popularized this idea, decoding intermediate LLM hidden states with the final unembedding matrix.
Tuned Lens~\citep{belrose2023eliciting} refined this by training per-layer translators to correct depth-wise affine drift.
Given its simplicity and gradient-free nature, zero-shot token-decoding became popular for LLM explainability~\citep{pal2023future,prakash2023layered,wendler2024llamas,li2024understanding, patil2024can,wang2025logitlens4llms,wu2026your} and was quickly adopted for ViTs by applying the classification head to decode patch tokens~\citep{vilas2023analyzing,joseph2025prisma}. 
Since then, it has been extended to multimodal language models~\citep{neo2025towards,jiang2025interpreting}, 
video models~\citep{zhang2025circuitprobe,kim2026map}, and audio models~\citep{yang2025audiolens,glazer2026beyond}.
Yet, unlike LLM decoders where unembeddings naturally decode individual tokens, ViT encoder heads are trained only on the class-token.
Despite this divergence, zero-shot patch decoding is assumed to reliably yield aligned maps. 
Here, we show this is highly model-dependent and not guaranteed.

\vspace{-7pt}
\paragraph{Patch misalignment in CLIP.}
Several previous works observed that CLIP's token decoding can be misleading. 
\citet{li2023clipsurgery} identified an ``opposite visualization'' effect where patch similarity maps highlight the background complement. Similar negative contrast was subsequently reported in open-vocabulary segmentation and dense inference~\citep{zhou2022maskclip,bousselham2024grounding,lan2024clearclip,wang2024sclip,shao2024cliptrase}, and analyzed geometrically by \citet{kang2025isclipideal}.
However, since prior art focused on downstream dense prediction, 
they largely approached this from an interventionist perspective,
developing test-time remedies like attention surgery or feature pruning to force spatial alignment. 
Rather than treating inversion as an isolated engineering flaw to be patched, we study it 
through the lens of representation underdetermination and 
ask whether alignment should be expected in the first place.
By systematically analyzing contrastive models across architectures, objectives, and scales,  
we show that inverted-explainability is a pervasive implicit bias of zero-shot patch decoding in VLMs, which can be shaped by the data in controlled settings.

\section{Explainability of vision transformers}

\subsection{Definitions}\label{sec:defs}

Consider an image $I$ of an object located inside a spatial mask $\gM$. The image is given as input to a ViT with $N$ patch-tokens and one class-token. Let $\smash{\{T[I]_i\}_{i=1}^N}$ denote the patch tokens in some layer within the ViT. 
A common approach to interpreting these tokens is to treat each of them as if it were the class token at the final layer.
Crucially, this zero-shot approach examines native representations, rather than training auxiliary probes.
For example, in ViTs for classification, we can apply to each of these tokens the classification head, which is normally applied only to the class-token of the last layer.
We can then inspect the logit value corresponding to the true object class, $\gC$, and treat it as the patch's score $\gS$.
Similarly, in image encoders of contrastive VLMs like CLIP, we can apply the projection head to each patch token and compute its cosine similarity with the text embedding of a template sentence of the form ``\texttt{a photo of a $\gC$}'' obtained from the text encoder.
This provides a score $\gS$ for each patch. 
The map of scores obtained for all patches in a given layer is often assumed to visualize how the model interprets each spatial location of the image, as shown in \cref{fig:explainable_diag}.

To quantitatively measure the quality of an explainability map for image $I$, we calculate the average score of the object's tokens, $\sobj$, and the average score of the background tokens, $\sbg$, as 
\begin{equation}\label{eq:eqxplainability_scores}
    \sobj[I]=\frac{1}{|\gM|}\sum_{j\in\gM} \gS(T[I]_j), \quad \sbg[I]=\frac{1}{|\gM^c|}\sum_{j\notin\gM} \gS(T[I]_j),
\end{equation}
where $\gM^c$ is the complement of $\gM$. The relation between these two mean scores can be used to define the explainability behavior for that image. 
The common interpretation is that if $\sobj>\sbg$ (green triangle in \cref{fig:explainable_diag}) then the object patches better correlate with the object's label than the background tokens, meaning the object tokens are directly aligned with the expected classification.
However, if $\sbg>\sobj$ (blue triangle in \cref{fig:explainable_diag}), 
the background tokens have higher scores than the object tokens, which according to the common interpretation means 
the model's knowledge of the class is localized in opposition to the object's tokens.
Lastly, if $\sbg\approx\sobj$ (purple 
band in \cref{fig:explainable_diag}), 
the token-decoding approach does not separate between background and object tokens. 
We term these three regimes \emph{aligned-explainability}, \emph{inverted-explainability}, and \emph{unexplainability}, respectively. 

The above definitions apply to a single image. To obtain a global measure of a model's explainability, we can examine its behavior over a large set of input images.
Specifically, we propose to measure the fraction $\gA$ of images within a large set $\gI$, which fall in the aligned-explainability regime, \ie,
\begin{equation}\label{eq:aligned_rate}
    \gA = \frac{\left|\left\{I\in\gI : \sobj[I]>\sbg[I]\right\}\right|}{|\gI|}.
\end{equation}
This metric captures the bias of a model toward aligned explainability over the set $\gI$. 
We consider a model aligned (\regaligned) if $66.7\%\le\gA$, unexplainable (\regunexp) if $33.3\%\le\gA<66.7\%$, and inverted (\reginverted) if $\gA<33.3\%$, making for equal regimes.

Explainability analysis can be applied to any layer. 
However, for conciseness, all experiments use the following convention unless specified otherwise. 
For models in which the patch tokens of the last layer were not affected by the loss during training, we always use the layer before the last. 
This includes, \eg, most ViT classifiers, in which the classification head operates only on the class token at the last layer, so that the patch tokens of that layer are not affected by the loss during training. 
For models in which the last layer's patch tokens are affected by the loss, we always show the last layer. 
This includes, \eg, \dino, whose pretrained classification head expects as input the class token concatenated to the mean of the last layer's patch tokens. 
Further details are in \cref{app:technical_details_real,app:last_layer}.

\subsection{Aligned-explainability of real-world vision transformers}\label{sec:vits}

\begin{figure}[t]
    \centering
    \includegraphics[width=\linewidth]{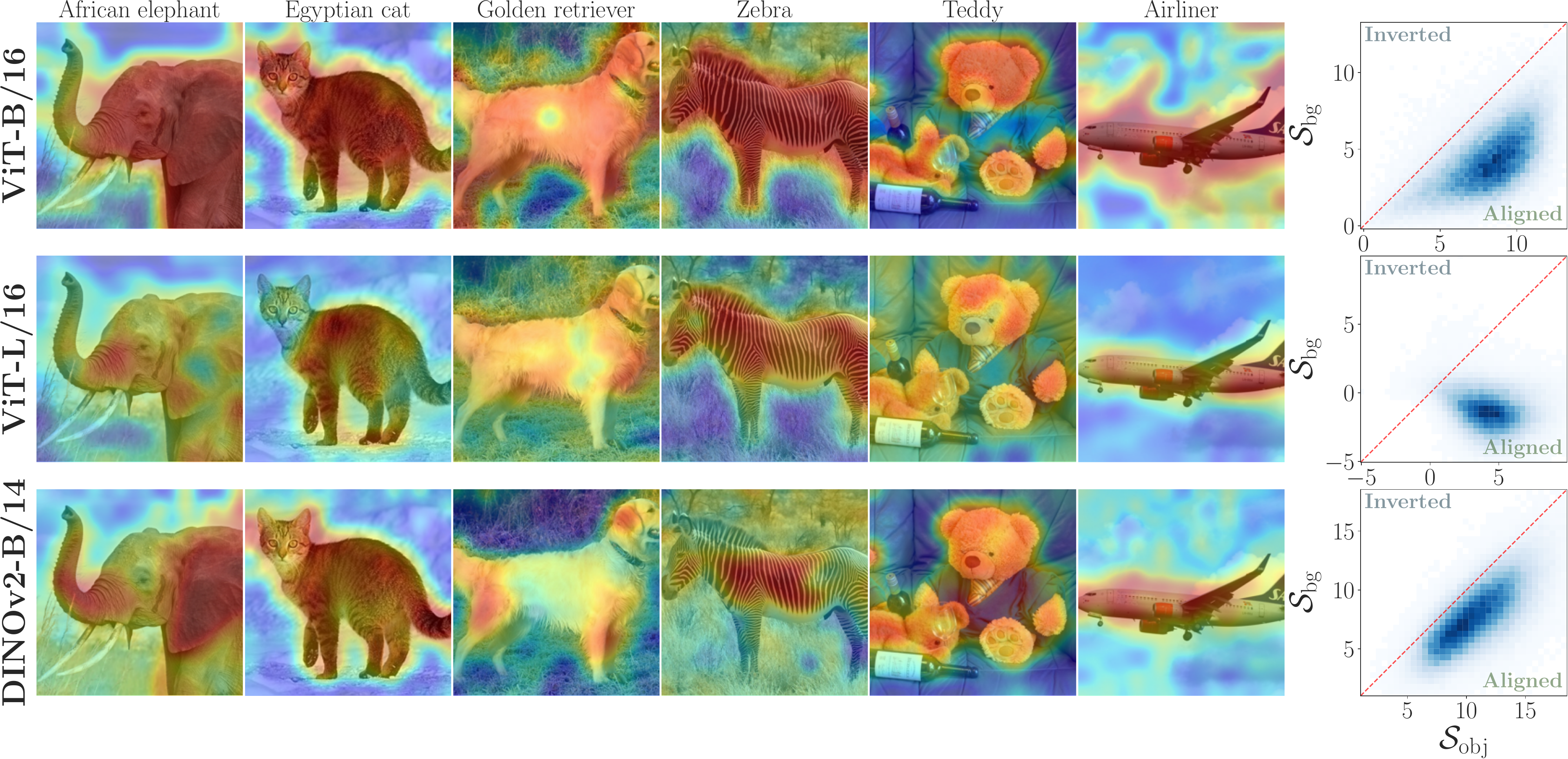}
    \caption{\textbf{Aligned-explainability bias of real-world ViT classifiers.} 2D histograms of $\sbg$ and $\sobj$ scores computed over all of PartImageNet++ (right) and explainability maps of six image inputs (left), on ViT-B/16, ViT-L/16, and \dino. All classifiers exhibit an aligned-explainability bias.
    }
    \label{fig:real_world_explainability}
\end{figure}

\begin{figure}[t]
    \centering
    \includegraphics[width=\linewidth]{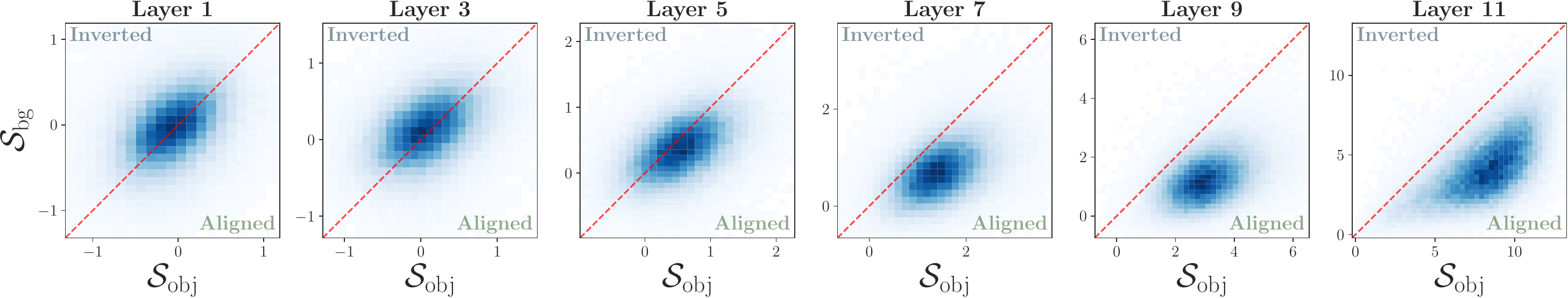}
    \caption{\textbf{Evolution of aligned-explainability in ViT-B/16.} 
    2D histograms of $\sbg$ and $\sobj$ scores computed over all of PartImageNet++ as they evolve through the model's 12 layers. While the early layers are unexplainable, as the features evolve they are more aligned. 
    }
    \label{fig:many_layers_vit}
\end{figure}

It has been well known that the feature maps of the last layers in convolutional neural network based classifiers align with the model's prediction~\citep{zhou2016learning,selvaraju2017grad}.
Therefore, it is natural to assume that the same alignment transfers to transformer-based classifiers.

\begin{table}[t]
    \centering
    \small
    \caption{\textbf{Explainability of real-world ViT classifiers and contrastive VLMs.}
    Scores are computed over all of
    PartImageNet++.
    Explainability regimes are marked according to \cref{sec:defs}, where 
    dashed rules mark the boundary between them.
    Contrastive VLMs exhibit a clear bias toward inverted-explainability, whereas classifiers are largely aligned-explainable.
    OpenCLIP's parentheses
    indicate 
    dataset size and 
    number of samples seen, and ``+reg'' marks models trained with register tokens.
    }
    \label{tab:vits_vlms}
    \begin{tabular}{lcc}
        \toprule
        \textbf{Classifier} & $\mathcal{A}$ (\%) & \textbf{Exp.} \\
        \midrule
ViT-B/16 (in1k) & 98.7 & \regaligned \\
ViT-B/16 (MIIL) & 98.6 & \regaligned \\
ViT-S/16 & 98.4 & \regaligned \\
ViT-B/16 (AugReg2) & 98.4 & \regaligned \\
ViT-B/16 & 98.4 & \regaligned \\
ViT-S/16 (in1k) & 98.4 & \regaligned \\
ViT-B/16 (orig) & 98.2 & \regaligned \\
ViT-L/16 (orig) & 98.1 & \regaligned \\
DINOv2-B/14+reg & 97.9 & \regaligned \\
ViT-L/16 & 97.8 & \regaligned \\
DINOv2-s/14+reg & 97.7 & \regaligned \\
ViT-B/32 (in1k) & 97.7 & \regaligned \\
ViT-B/32 & 97.5 & \regaligned \\
DINOv2-g/14+reg & 96.8 & \regaligned \\
DINOv2-s/14 & 96.7 & \regaligned \\
DINOv2-B/14 & 94.4 & \regaligned \\
DeiT-B/16 & 92.2 & \regaligned \\
DINOv2-g/14 & 75.6 & \regaligned \\
ViT-Ti/16 & 70.5 & \regaligned \\
\cdashedline 
DINOv2-L/14 & 63.5 & \regunexp \\
DINOv2-L/14+reg & 41.1 & \regunexp \\
\cdashedline
ViT-B/32 (SAM) & 14.4 & \reginverted \\
ViT-B/16 (SAM) & 6.4 & \reginverted \\
DeiT-Ti/16 & 4.1 & \reginverted \\
DeiT-S/16 & 1.4 & \reginverted \\
        \bottomrule
    \end{tabular}
    \quad
    \begin{tabular}{lcc}
        \toprule
        \textbf{Contrastive VLM} & $\mathcal{A}$ (\%) & \textbf{Exp.} \\
        \midrule
SigLIP-B/32 & 22.5 & \reginverted \\
OpenCLIP-B/16 (LAION80M-3B) & 13.8 & \reginverted \\
OpenCLIP-L/14 (LAION80M-3B) & 12.1 & \reginverted \\
OpenCLIP-B/16 (LAION400M-3B) & 11.5 & \reginverted \\
OpenCLIP-B/16 (LAION80M-13B) & 8.5 & \reginverted \\
OpenCLIP-L/14 (LAION80M-34B) & 7.0 & \reginverted \\
OpenCLIP-B/16 (LAION80M-34B) & 7.0 & \reginverted \\
OpenCLIP-B/32 (LAION2B-34B) & 6.3 & \reginverted \\
MetaCLIP-H/14 & 4.6 & \reginverted \\
CLIP-L/14 & 4.3 & \reginverted \\
OpenCLIP-B/16 (LAION400M-34B) & 3.5 & \reginverted \\
OpenCLIP-B/16 (LAION2B-3B) & 3.5 & \reginverted \\
OpenCLIP-B/16 (LAION2B-13B) & 3.5 & \reginverted \\
OpenCLIP-L/14 (LAION2B-3B) & 3.4 & \reginverted \\
OpenCLIP-B/16 (LAION400M-13B) & 3.3 & \reginverted \\
MetaCLIP-L/14 & 3.2 & \reginverted \\
MetaCLIP-G/14 & 3.0 & \reginverted \\
OpenCLIP-H/14 (LAION2B-34B) & 2.9 & \reginverted \\
OpenCLIP-L/14 (LAION2B-34B) & 2.8 & \reginverted \\
MetaCLIP-B/32 & 2.6 & \reginverted \\
CLIP-B/32 & 2.3 & \reginverted \\
OpenCLIP-g/14 (LAION2B-34B) & 2.2 & \reginverted \\
OpenCLIP-B/16 (LAION2B-34B) & 2.2 & \reginverted \\
OpenCLIP-g/14 (LAION2B-13B) & 2.0 & \reginverted \\
CLIP-B/16 & 2.0 & \reginverted \\
MetaCLIP-B/16 & 1.5 & \reginverted \\
        \bottomrule
    \end{tabular}
\end{table}

\begin{wrapfigure}{r}{0.37\linewidth}
    \vspace{-0.35cm}
    \centering
    \includegraphics[width=\linewidth]{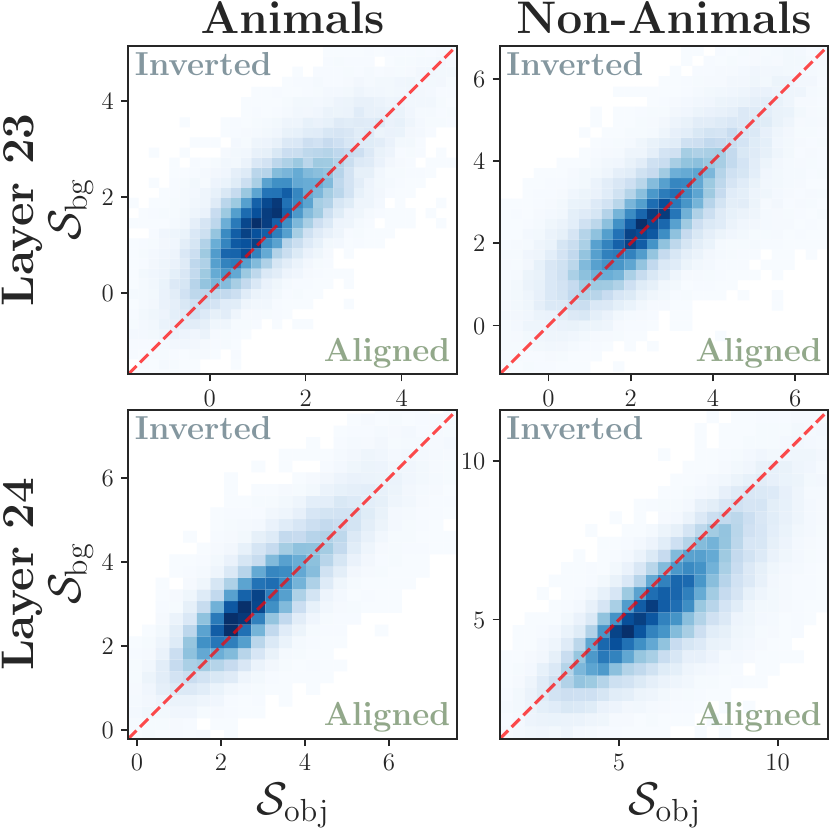}
    \caption{\textbf{Different subsets can lie in different regimes.} 
    2D histograms of $\sbg$ and $\sobj$ scores computed separately for animal and non-animal classes in PartImageNet++ on \dino-L/14.
    In the penultimate layer, animals have inverted-explainability and non-animals are unexplainable. In the last layer, animals are unexplainable and non-animals have aligned-explainability.
    }
    \label{fig:dino-animals}
    \vspace{-1cm}
\end{wrapfigure}
\Cref{fig:real_world_explainability} shows the token-decoding explainability behavior of ViT-B/16, ViT-L/14~\citep{dosovitskiy2021an}, and \dino~\citep{oquab2024dinov}.
We plot explainability maps for six varied image inputs, as well as 2D-histograms of $\sobj$ and $\sbg$ calculated over all of PartImageNet++~\citep{li2024partimagenet++}. 
As can be seen, all models are biased toward the aligned-explainability regime. 
This is true even for \dino, which was initially trained in a self-supervised manner and only later supplemented with a simple classifying head.
Notably, deeper layers are more explainable than shallower ones (see \cref{fig:many_layers_vit}).
Histograms of additional classifiers are in \cref{app:more_results_vits}.

However, while most ViT classifiers have a strong aligned-explainability bias, it is not universal. 
\Cref{tab:vits_vlms} compares explainability scores on PartImageNet++ over a large set of ViT classifiers, varying in architecture, training loss, data augmentation, and optimizer.
Indeed, ViTs trained with sharpness-aware minimization (SAM)~\citep{foret2021sharpnessaware} and some DeiTs~\citep{touvron2021deit} exhibit completely inverted explainability. 
Importantly, this inversion is strictly directional and not an activation-norm artifact, and is also unrelated to register tokens~\citep{darcet2024vision} (details in \cref{app:more_results_vits,app:techincal_dino_details}).
We emphasize that it is also not related to some classifier failure, as all models reach a high top-1 accuracy, with almost all exceeding 88\% (See \cref{app:more_results_vits}).

A more fine-grained analysis reveals that the explainability patterns of a model can differ across classes. 
For example, \cref{fig:dino-animals} depicts the explainability histograms of \dino/L-14 over two distinct subsets: images of animals and images that belong to the non-animal classes.
Surprisingly, animals are inverted in the penultimate layer and lean toward unexplainability in the last layer, whereas non-animals are unexplainable in the penultimate layer and aligned in the last layer. 

These observations suggest that different parameters of the optimization and the loss landscape bias a model toward different explainability regimes, and that this behavior is independent of its accuracy. Aligned explainability is therefore a property that cannot be assumed a-priori for ViT classifiers.

\subsection{Inverted-explainability of real-world vision-language transformers}\label{sec:vlms}

\begin{figure}[t]
    \centering
    \includegraphics[width=\linewidth]{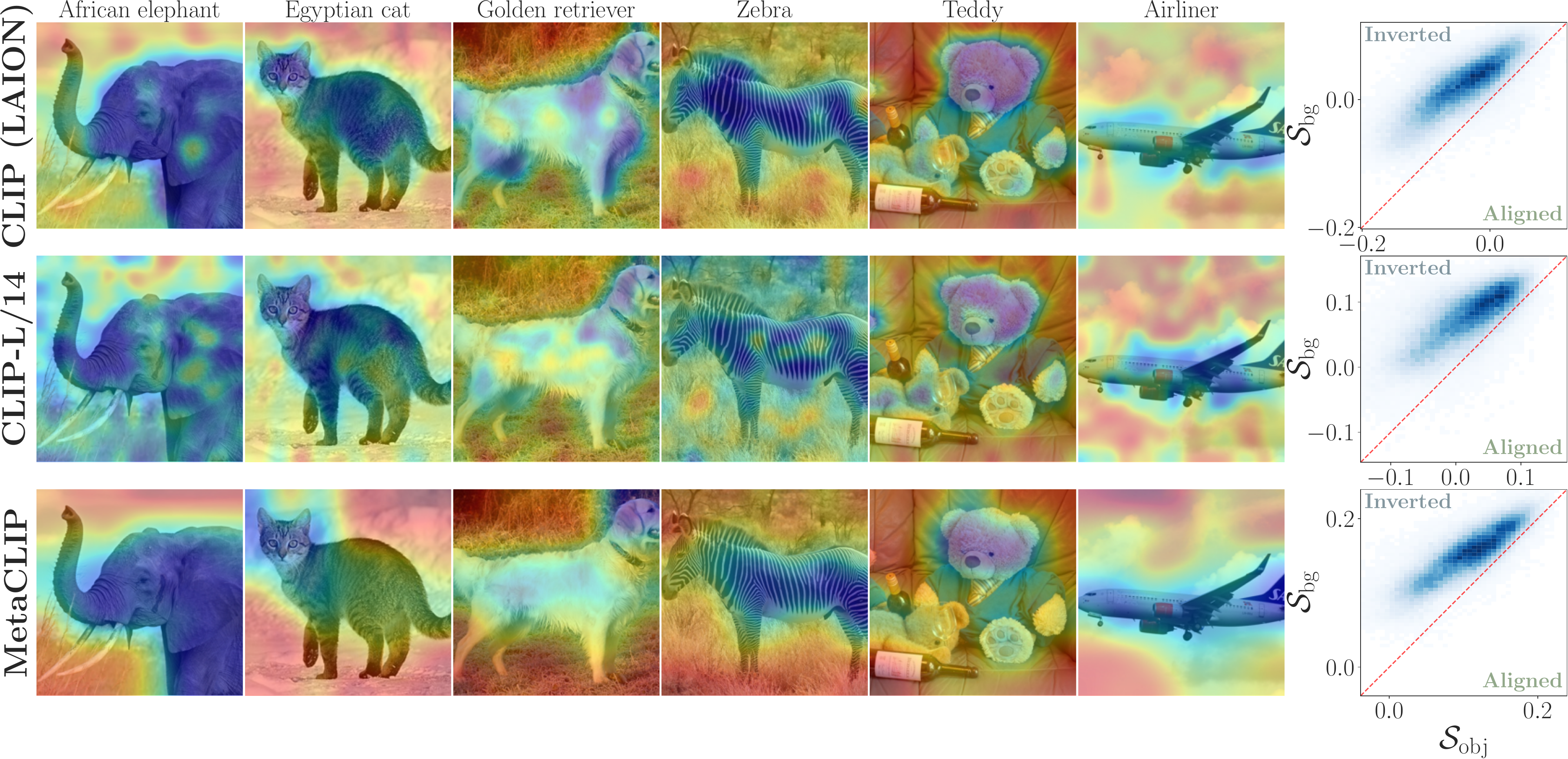}
    \caption{\textbf{Inverted-explainability bias of real-world ViT-based contrastive VLMs.} 2D histograms of $\sbg$ and $\sobj$ scores calculated over PartImageNet++ (right), as well as explainability maps of six image inputs (left), on three models: CLIP-B/16 trained by OpenCLIP on LAION, CLIP-L/14, and MetaCLIP. All contrastive VLMs exhibit an inverted explainability bias.
    }
    \label{fig:real_world_antiexplainability}
\end{figure}

It is tempting to assume that the bias toward aligned-explainability of ViT classifiers would be seen also in contrastive VLMs like CLIP~\citep{radford2021learning}. 
However, consistent with prior localized observations~\citep{li2023clipsurgery,bousselham2024grounding}, it turns out this is not the case.
\Cref{fig:real_world_antiexplainability} shows the explainability behavior of two CLIP models as well as of MetaCLIP~\citep{xu2024demystifying}.
In all three models the patch tokens of the object consistently score lower than those of the background, \ie, the models display inverted-explainability.

As seen in \cref{tab:vits_vlms}, contrastive VLMs are consistently biased toward inverted-explainability. 
A natural suspicion is that the inversion originates in the contrastive objective itself, whose softmax ties every image to every caption in the batch through a single shared normalization.
To examine this, we analyze a SigLIP~\citep{zhai2023sigmoid} model, which opts for a sigmoid loss applied independently to each image-caption pair.
Since the original pre-trained SigLIP image encoders do not employ a class token, we train a class-token ViT-B/32 SigLIP on LAION-400M~\citep{schuhmann2021laion400m} (see \cref{app:siglip} for the full setup).
Nonetheless, this SigLIP exhibits inverted-explainability as well.
We include in \cref{app:more_results_vlms} explainability histograms of additional contrastive VLMs, as well as analyses on additional data scales and training budgets of OpenCLIP models~\citep{cherti2023reproducible}.
All these models lie in the inverted-explainability regime, even at the largest training scales. 
Similarly to the the case in ViT classifiers, the explainability evolves throughout the layers (see \cref{fig:many_layers_clip}).

Several previous works have noted related phenomena in CLIP.
However, this was usually taken as a flaw of the model in need of ``fixing'', by realigning the final attention block~\citep{zhou2022maskclip,li2023clipsurgery,bousselham2024grounding,shao2024cliptrase,wang2024sclip}.
For example, \citet{li2023clipsurgery} noticed inverted similarity map visualizations, and proposed this occurs due to self-attention relations between inconsistent semantic regions. They concluded that this is a negative phenomenon that arises due to a lack of local supervision on the features. Other papers followed a similar logic. 
In particular, all these works inherently assume that the representations of contrastive VLMs \emph{should} have aligned-explainability. 
Yet, there is no apparent reason to assume these representations should exhibit any explainability at all. 
Perhaps the representations do not require ``fixing'', and their inverted explainability is merely an artifact of the task over which the model is trained.

\section{Controlled explainability analysis}\label{sec:controlled}

In \cref{sec:vits,sec:vlms} we demonstrated that real-world ViT classifiers show a bias toward aligned-explainability, whereas contrastive VLMs exhibit inverted-explainability.
To test whether it is indeed the training task that ultimately governs the explainability regimes of ViTs, we conduct a large-scale analysis spanning different design choices over synthetic datasets.
Specifically, we construct two types of synthetic datasets, fit for classification and for contrastive learning, and train ViTs with varying dataset configurations.
We find that for both classifiers and contrastive VLMs, the training can converge to either aligned or inverted explainability.
This demonstrates that beyond the loss, the data distribution also affects the explainability of ViTs.
However, the effect is not deterministic. We show that for a fixed task and data distribution, different training seeds can converge to different explainability regimes.
This shows that the training loss and the data distribution do not impose a strict regime, but rather promote different explainability biases.

\subsection{Setting}\label{sec:controlled_setting}
\paragraph{Data construction.}
We construct both datasets to model the compositional nature of real-world images, requiring a ViT to localize and semantically understand a distinct object embedded within a background.
Accordingly, both datasets consist of $D\times D\times3$ sized images,
where a small, randomly sampled patch $\rvx$ of size $p\times p\times3$, representing the object, is implanted at a uniformly random spatial location within a background of random noise $\rvn\in\sR^{3D^2-3p^2}$ that is statistically independent of $\rvx$. 
For a supervised dataset with $\gC$ classes, $\rvx$ is sampled from a multivariate Gaussian mixture model with $\gC$ equal-weight Gaussians. Specifically, $\rvx=\rvx_\rvc$, where $\rvc$ is a random integer in the set $\{1,\ldots,\gC\}$ and for any $c$ in this set, $\rvx_c\sim\gN(\vmu_c,\Sigma)$. The Gaussians have distinct means, $\smash{\vmu_c\in\sR^{3p^2}}$, but share the same covariance matrix $\Sigma$. Specifically, 
\begin{equation}\label{eq:synth_patch_vit}
    \rvx_c = \vmu_c + \gamma\,\mA\rvz + \rvepsilon_\rvx,
    \qquad
    \rvz\sim\gN(\vzero,\mI_r),
    \quad
    \rvepsilon_\rvx\sim\gN(\vzero,\mI_{3p^2}),
\end{equation}
where $\smash{\mA\in\sR^{3p^2\times r}}$ and the scalar $\gamma$ is a hyper-parameter controling the scale of the covariance matrix. Here, $\rvz$ and $\rvepsilon_\rvx$ are statistically independent. 
Contrastive training requires more delicate modeling to simulate paired multi-modal data (\eg, image-text pairs).
Instead of a label, the visual object patch $\rvx$ and its continuous paired representation $\rvy\in\sR^d$ are jointly sampled from a shared latent variable $\rvw$, and multiplied by scalar factors $\alpha_\rvx,\alpha_\rvy$ as
\begin{equation}\label{eq:synth_patch_vlm}
    \rvx = \alpha_\rvx\rvw + \rvepsilon_\rvx,
    \quad
    \rvy = \alpha_\rvy\mV\rvw + \rvepsilon_\rvy,
    \qquad
    \rvw, \rvepsilon_\rvx \sim\gN(\vzero, \mI_{3p^2}),
    \quad
    \rvepsilon_\rvy\sim\gN(\vzero,\mI_{d}),
\end{equation}
where $\rvw, \rvepsilon_\rvx$, and $\rvepsilon_\rvy$ are statistically independent.
Here, $\smash{\mV\in\R^{d\times 3p^2}}$ has orthonormal rows and $d<3p^2$.
In both datasets, $\rvn$ is sampled from an isotropic Gaussian distribution with variance scaled to control the \emph{object to background-noise ratio} (OBNR), defined as 
$\textnormal{OBNR}=\sqrt{\frac{\textnormal{Tr}\{\textnormal{Cov}(\rvx)\}/\dim(\rvx)}{\textnormal{Tr}\{\textnormal{Cov}(\rvn)\}/\dim(\rvn)}}$, where $\dim(\cdot)$ is the dimension of its argument. 
Specifically, when the OBNR is set to $1$, the background noise variance is the most similar to the variance of the object's patch, leading to a harder separation of the object from its background. Implementation details of the background scaling appear in \cref{app:data_construction}.
In all supervised experiments, $\vmu_c$ and $\mA$ are sampled once by drawing i.i.d.~standard normal entries, ensuring that different training seeds of the same dataset configuration are compared on the same data distribution. The matrix
$\mV$ is constructed by sampling once from the standard normal distribution and orthogonalizing the rows via QR decomposition, setting a single $\mV$ for all contrastive experiments.
In all experiments, we set $D=20, p=8$ for both datasets and $r=38$ and $d=16$ for supervised and contrastive training, respectively.

This construction allows us to explicitly control task difficulty by adjusting the OBNR, $\alpha_\rvx$, $\alpha_\rvy$, or~$\gamma$. 
We sweep $\alpha_\rvx,\alpha_\rvy\!\in\!\{1,2,4\}$, $\gC\!\in\!\{10,20\}$ and $\gamma\!\in\!\{0.5,0.75,1\}$ at varying OBNRs, giving 81 contrastive and 66 supervised configurations, each trained under three seeds. 
\Cref{app:synth_more_results} includes additional results for 
a supervised configuration where $\mA$ differs per class.

\paragraph{Architecture.}
In both types of experiments, we opt for small ViTs that consist of 12 layers. For the contrastive multi-modal case, we do not include tokenization and embedding matrices and instead operate on the sampled $\rvy$ directly. Additionally, we use $\rvy$ as a single token. More details on the architecture and training settings appear in \cref{app:synth_arch_details}.

\subsection{Explainability in synthetic experiments}\label{sec:synth_exp_results}

\begin{figure}[t]
    \centering
    \includegraphics[width=\linewidth]{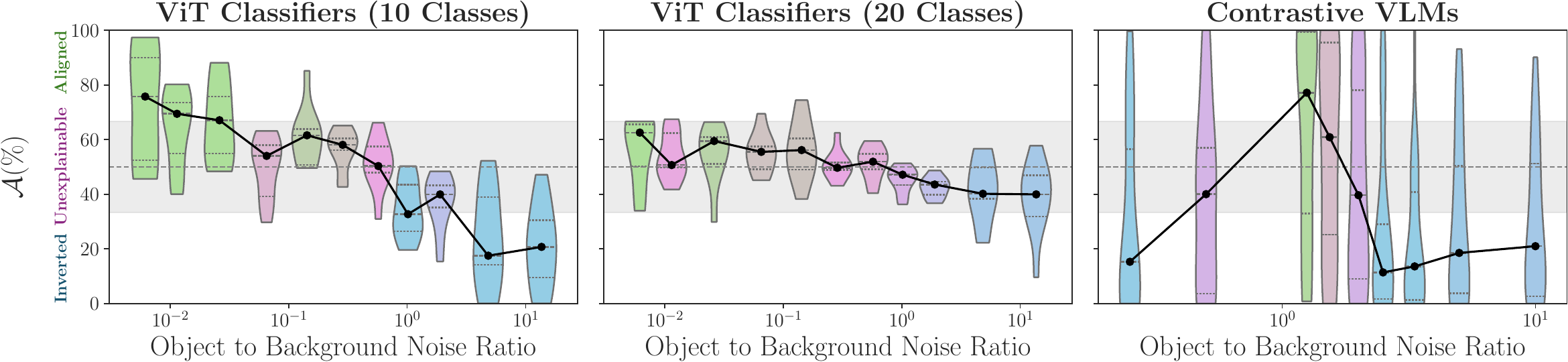}
    \caption{\textbf{Explainability trends of controlled ViT experiments.}
    Violin plots of $\gA$ scores of 99 classifiers with 10 (left) or 20 (middle) classes, and 232 contrastive models (right), trained on our synthetic datasets, for different OBNR values. 
    The violins are colored gradually according to their medians (indicated with black dots), with blue, green, and purple denoting inverted-explainability, aligned-explainability and unexplainability, respectively. 
    The black lines connect the medians, the dashed rules mark $\gA=50\%$, and the gray background indicates the unexplainability regime.
    In ViTs, low or high OBNR values induce aligned or inverted explainability biases, respectively, while $\textnormal{OBNR}\approx1$ promote unexplainability.
    In contrastive VLMs, low and high OBNR values promote inverted explainability, while $\textnormal{OBNR}\approx1$ yields aligned-explainability.
    Despite these biases, ViTs occasionally converge outside the dominant regime for the same data distribution.
    }
    \label{fig:synthetic_totals}
\end{figure}

Among all hyper-parameters, we find the OBNR to have the most prominent effect on the explainability regime to which the ViTs converge.
We plot in \cref{fig:synthetic_totals} the distribution of $\gA$ across our trained models for varying OBNR values. Each model performs similarly well on its task (see \cref{app:synth_perf}).

For ViT-classifiers, very high or very low OBNR values induce some sort of explainability: 
lower OBNR levels cause alignment bias, whereas higher OBNR levels tend to lead to inversion. OBNR values that are close to 1, where the object is similar in variance to the background and the classification task is harder, tend to lie in the unexplainable regime. 
Additionally, a smaller number of classes leads to a sharper explainability trend.

For contrastive VLMs, we see a slightly different behavior.
While high OBNR values promote an inverted-explainability bias similarly to ViT classifiers, low values also induce inverted-explainability. 
Interestingly, aligned-explainability emerges when the OBNR is close to 1, and unexplainability emerges in intermediate OBNR values.

We note that in both cases, while models are biased to converge to some regime, as indicated by the width of the violins and their medians (connected by the black line), some can still converge outside the biased regime for the same data distribution (the violins' tail). This aligns with the spurious inverted-explainable ViTs samples in \cref{tab:vits_vlms}. In fact, different regimes can occur for the same setting using different training seeds (see \cref{app:synth_seeds}). 
Further experiments in \cref{app:synth_more_results}, including different data distributions and smaller ViTs, show the same trend: 
explainability is an implicit bias determined by the problem setting and not a fixed property of the training task or data distribution.

\section{Theoretical analysis of a toy model}
\label{sec:theory}

Previous sections showed that the explainability regime changes across models, objectives, and training seeds.
However, they leave a core question open: can a model be optimal yet fail to be explainable?
To answer this, we study a problem simple enough to solve analytically. 
We derive a closed form expression for the optimal classifier, identify a toy attention architecture that precisely implements it, and show that its token-decoding map can be made aligned, unexplainable, or inverted.
All three versions produce identical logits, making them indistinguishable under any logit-based loss.

\vspace{-4pt}
\paragraph{The classification task.}
We consider a one-dimensional image $\rvz=(\rz_1,\ldots,\rz_N)\in\R^N$, a vector of length $N$ whose entry $\rz_n$ is the content of position $n$.
One position $n_\textnormal{obj}$, drawn uniformly at random, holds an object, and every other position holds background $\smash{\rz_n\sim\gN(0,\sigma^2)}$ drawn independently of the class.
The object belongs to a class $\rc\in\{1,\ldots,K\}$, which equals $i$ with prior probability $\pi_i$, and its content follows a Gaussian mixture over the $K$ classes: given $\rc=i$, $\rz_{n_\textnormal{obj}}\sim\gN(\mu_i,\sigma_i^2)$.

\begin{proposition}[Optimal classifier]\label{prop:theory_optimal}
The optimal classifier for this data distribution is
\begin{equation}\label{eq:theory_lse}
    p_{\rc|\rvz}(i|\vz) = \mathrm{softmax}_i\!\left(\log\left(\sum_{n=1}^{N}e^{\psi_{i,n}}\right)+\beta_i\right),
    \qquad
    \psi_{i,n}=\frac{z_n^2}{2\sigma^2}-\frac{(z_n-\mu_i)^2}{2\sigma_i^2},
\end{equation}
where $\beta_i=\log\pi_i-\log\sigma_i$.
\end{proposition}

\begin{figure}[t]
    \centering
    \includegraphics[width=\linewidth]{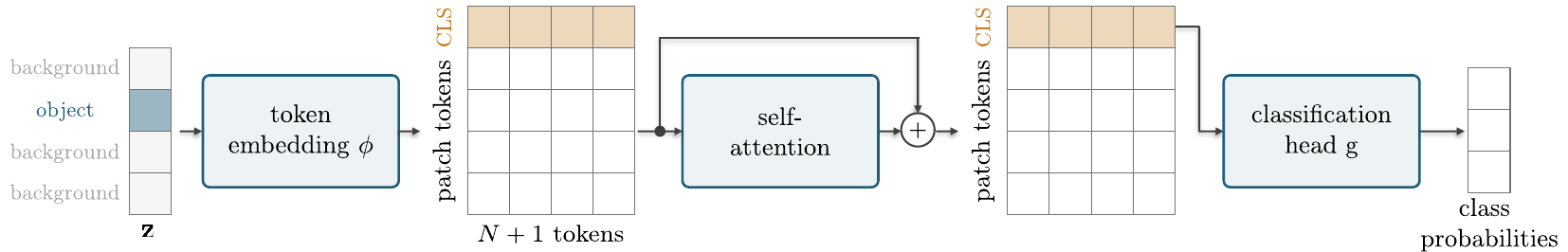}
    \caption{\textbf{The toy transformer model.}
    Each token in the 1D $\rvz$ image input is embedded using $\phi$.
    After prepending a class token, a single self-attention block with a residual connection produces representations. 
    The classification head $g$ is applied to the class token to output predictions.
    }
    \label{fig:toy_model}
\end{figure}

\paragraph{The toy model.}
\Cref{fig:toy_model} shows the model's architecture: token embedding, followed by a single-head self-attention, a residual connection, and finally a classification head.
It treats every entry of $\rvz$ as a token, and prepends a learnable class token, giving $N+1$ tokens in total.
At a high level, every patch token carries a score for each class, and that is what a token-level explanation reads.

\begin{theorem}[One optimum, three regimes]\label{thm:theory_regimes}
There are weights for the toy model, with a self-attention value matrix $\mW_V=(N+1)\alpha\mI$ for a scalar $\alpha$, such that for every $\alpha>0$ and every $\alpha\le-1$ the class token computes the optimal classifier of \cref{prop:theory_optimal} exactly, with the same logits for every $\alpha$.
For these weights and every class $i$, the scores $\sobj$ and $\sbg$ of class $i$ satisfy:
\begin{enumerate}
    \item if $\alpha>0$, then $\mathbb{E}[\sobj-\sbg|\rc=i]>0$, so the map is aligned-explainable in expectation;
    \item if $\alpha\le-1$, then $\mathbb{E}[\sobj-\sbg|\rc=i]<0$, so the map is inverted-explainable in expectation;
    \item if $|\alpha|\to\infty$, then $\sobj-\sbg\to0$ in every image, so the map becomes unexplainable.
\end{enumerate}
Items 1 and 2 hold when the distribution of class $i$ differs from the background ($\mu_i\neq0$ or $\sigma_i\neq\sigma$).
\end{theorem}
The weights and the proofs are given in \cref{app:theory}.
Importantly, when the same logits are produced using different weights, the training loss alone cannot tell the weight settings apart.

\Cref{thm:theory_regimes} yields two key conclusions that challenge the assumption of inherent explainability. 
First, it shows that an inverted token-decoding map or an unexplainable map incurs no loss penalty.
Thus, these non-aligned maps do not inherently indicate a defective or suboptimal model. 
Second, since the same loss value can be reached with various weights, it does not dictate the convergence regime.
Therefore, the token-decoding explainability bias of a model must stem from other factors, such as the architecture, the optimization, or the data distribution.

Naturally, our toy model has limitations.
The setting is simplified, using one dimension, a single implanted object patch, and one attention block.
It also decodes patch tokens unaffected by the loss, and requires an embedding dimension of at least $K+1$, while real ViTs are often narrower relative to the number of classes. 
Furthermore, while it demonstrates that an optimal model can converge to any of the three regimes, it does not explain the specific regime adopted by any real-world model.
Nevertheless, we verify in \cref{app:last_layer} that decoding patch tokens unaffected by the loss of real-world ViTs also exhibits the model's regime. 
We leave the study of narrower embeddings to future work.

\section{Conclusion}
\label{sec:conclusion}

We examined whether token-decoding explainability is an inherent property of ViTs, and found that it is not.
Most real-world ViT classifiers are aligned-explainable, yet some are inverted while still classifying accurately. Furthermore, contrastive vision-language models are consistently inverted. 
We presented controlled experiments which showed that the explainability regime of a model shifts with its data and training setting. Finally, we presented a toy problem that illustrated that even an optimal transformer can be aligned, unexplainable, or inverted. 
While our analysis spans many real-world models, it focuses on gradient-free decoding of patch-tokens. Extending this study to gradient-based explainability techniques is an interesting direction for future work. 
Our synthetic models are deliberately small, and our toy model establishes that every regime is attainable at optimality without predicting which regime a given model reaches. Nevertheless, our work exposes an important observation, which is that aligned-explainability in patch-token decoding is a property to verify for each model, not one to assume.

\subsection*{AI use statement}

During the writing of this paper, generative AI tools (Claude 5~\citep{claude} and Gemini 3~\citep{gemini}) were mainly used to polish the text, and refine its clarity and fluency, including assisting with translation. 
In addition, they were used to generate drafts of some sections in the paper (specifically, sections relating to the proofs in \cref{sec:theory,app:theory}), as well as to assist in formulating our mathematical claims and in writing proofs.
They also assisted when creating figures in this paper and when coding our experiments.
Generative AI tools were not used for retrieval or discovery of related work, for research ideation, for generating synthetic data sets, for developing the theoretical models or conceptual frameworks of our work or providing the critical ingredients in the proofs, for designing or providing feedback on research methodology or experiments, for proposing or refining hypotheses, for implementing methods, for cleaning and reformatting datasets, for supporting qualitative and thematic data analysis, or for interpreting results.
We have reviewed all AI-assisted work, including all proofs, code, and figures they helped in creating, using manual verification by two authors. We take full responsibility for the content of this work, including all text, claims, or artifacts generated with the help of generative AI.

\subsection*{Ethics statement}

In this work, we analyze pre-trained, real-world ViTs and demonstrate that  common assumptions regarding their explainability do not hold, both theoretically and in practice. 
Specifically, we show that token-decoding explainability methods are not guaranteed to provide meaningful semantics, so reliance on such explainability maps to audit model behavior can be misleading.
By exposing and disproving these assumptions, we aim to prevent potentially harmful reliance on flawed explainability metrics or misplaced trust in AI tools.
Additionally, all our experiments rely only on publicly available data.
We use PartImageNet++ for our analysis of real-world models and synthetic datasets for our controlled experiments, and to train SigLIP with a class-token we used the publicly available LAION-400M dataset.
Lastly, while our theoretical model in \cref{sec:theory} uses a shallow toy network, which is non-practical for real-world models, we hope even this relatively modest setting can serve as a basis for future work extending our results to deeper networks. 

\subsection*{Reproducibility statement}

As explainability of ViTs helps create transparency of these models, so do we aim to have such transparency with regard to our paper.
Therefore, we disclose all information needed for reproducing our results and analysis.
Specifically, we include full and detailed proofs for all theoretical claims in \cref{app:theory}.
For our real-world experiments, we use only publicly available checkpoints, except for the SigLIP model we trained on a publicly available dataset, for which we  release weights in our project \href{\projecturl}{website}. 
For our controlled experiments, all details needed to recreate the experiments appear in \cref{sec:controlled_setting,app:technical_details_synth} and we release the code needed to train all models on our project \href{\projecturl}{website}.
The supplementary material includes code that computes $\sobj$, $\sbg$, and $\gA$ on PartImageNet++ for any \texttt{timm} ViT, \dino, or OpenCLIP model.

\subsubsection*{Acknowledgments}
This research was made possible through a GPU compute resource grant funded
by the Association of University Heads, the Council for Higher Education and
the AI Research Compute Center. The research was additionally supported by the Israel Science Foundation (ISF) under Grant  no.~2318/22. TM acknowledges support from the Joseph and Sadie Danciger Chair in Engineering.
The research of HM was supported by the Planning and Budgeting Committee of the Israeli Council for Higher Education. 
The authors are grateful to Matan Kleiner for his insightful input to this work.

\bibliography{0_citations}
\bibliographystyle{iclr2027_conference}

\clearpage
\beginsupplement
\appendix

\section{Technical details of analyzing real-world models}\label{app:technical_details_real}

\paragraph{Token grid.}
In all experiments, since the object is positioned at pixel resolution it generally straddles token boundaries.
Therefore, before computing $\sobj$ and $\sbg$, we assign each token to $\gM$ if more than half of its pixels lie inside the object mask, and to $\gM^c$ otherwise.
Thus, every token contributes to exactly one of $\sobj$ and $\sbg$.

\paragraph{Applying normalization layers.}
In \cref{sec:defs} we describe how the score $\gS$ of a patch is computed by the token-decoding approach of passing it through the model's classification head.
We note that some models are trained such that the class-token of the final layer is first normalized via some normalization layer before being passed to the classification head. 
In all experiments, whenever a model includes such a normalization layer, we always apply the same processing to the patch tokens before propagating them through the classification head.
This is also the case for contrastive VLMs, where usually the class-token is passed through a normalization layer before being projected onto the common image-text space.
Therefore, we always pass the patch tokens through the final normalization layer and only then apply the classification head.

\subsection{Explainability analysis of \dino}\label{app:techincal_dino_details}

\paragraph{Classification head.}
\dino for classification is trained in two stages. First, it is pretrained in a self-supervised manner, and later supervised training is used to train its classification head. We use this classifying head in all \dino experiments.

In standard ViTs, the classification head operates on the class-token alone, and therefore to analyze the explainability of a model using token-decoding, we can reuse the classification head directly on patch tokens.
However, in \dino, the classification head expects as input a concatenation of the class-token and the mean of all patch tokens. 
Therefore, to apply token-decoding on \dino, we replace the class-token input of the classification head with the currently analyzed patch-token, and keep the concatenated patch-tokens mean untouched.

To verify the validity of this approach, we train for each \dino model variant a second classification head that operates on the class-token alone and apply token-decoding to it exactly as in standard ViTs.
Specifically, we keep each pretrained \dino backbone frozen and train for it a linear classification head on its normalized final-layer class-token over the ImageNet-1k training set~\citep{deng2009imagenet}.
We follow the linear evaluation protocol of~\citet{oquab2024dinov}, and train for $10$ epochs using SGD with momentum $0.9$, no weight decay, a batch size of $1{,}024$, and a cosine learning-rate schedule.
As in their protocol, we train the head with $13$ learning rates between $10^{-5}$ and $0.1$, and keep the one with the highest top-1 accuracy on the ImageNet-1k validation set.
Training images are resized to $256$ pixels and center-cropped to $224$, with no further augmentation.

\Cref{tab:dino_cls_head} compares the two heads.
All new heads, which accept only a class-token, 
are within one point of the top-1 accuracy of the official head for the corresponding model.
Importantly, every \dino model remains in the same explainability regime when using the new head and $\mathcal{A}$ changes by at most $5.1$ points.
In particular, \dino-L/14 and \dino-L/14+reg remain unexplainable.
Therefore, their unexplainability does not stem from our patch-token substitution.

\paragraph{Register tokens.}

\citet{darcet2024vision} have shown that ViTs can recycle background tokens as registers that store global image information. 
Specifically, they showed that without adding explicit training register tokens, background patches tend to have high-norm activations.
Therefore, considering how inverted-explainability is affected by the relation between $\sbg$ and $\sobj$, one might expect all models trained with registers to exhibit aligned-explainability and all models trained without them to exhibit inverted-explainability. 
However, we demonstrate in \cref{tab:vits_vlms} that models trained without registers exhibit a clear bias towards aligned-explainability, despite background patches having higher activation norms.
In fact, it seems that the existence of registers does not dictate the regime of a model at all.
We include in \cref{tab:vits_vlms,tab:acc} \dino models trained with and without token registers over the same architectures. 
Indeed, the results clearly indicate that models trained with or without token register can still exhibit aligned-explainability or unexplainability, irrespective of the existence of the registers.

\begin{table}[t]
    \centering
    \small
    \caption{\textbf{Comparing the official classification heads of \dino with a class-token head.}
    We compare the original \dino heads used in \cref{tab:vits_vlms} (``Official'') with new trained versions of these heads (``CLS-only'').
    Specifically, the original heads expect as input a patch-token concatenated to the mean value of all patch-tokens, and the new heads are trained on a single input of the class-token.
    In both, we apply the token-decoding approach of replacing the class-token with patch-tokens of the final layer.
    Top-1 accuracy is computed over the ImageNet-1k validation set, and $\mathcal{A}$ over PartImageNet++ at the last layer.
    Explainability regimes are marked according to \cref{sec:defs}.
    Both heads yield the same regime for every model.
    }
    \label{tab:dino_cls_head}
    \begin{tabular}{lcccccc}
        \toprule
        & \multicolumn{2}{c}{\textbf{Top-1 (\%)}} & \multicolumn{2}{c}{$\mathcal{A}$ (\%)} & \multicolumn{2}{c}{\textbf{Exp.}} \\
        \cmidrule(lr){2-3} \cmidrule(lr){4-5} \cmidrule(lr){6-7}
        \textbf{Classifier} & Official & CLS-only & Official & CLS-only & Official & CLS-only \\
        \midrule
        DINOv2-s/14 & 81.1 & 80.3 & 96.7 & 97.2 & \regaligned & \regaligned \\
        DINOv2-s/14+reg & 80.8 & 80.2 & 97.7 & 97.8 & \regaligned & \regaligned \\
        \midrule
        DINOv2-B/14 & 84.3 & 83.8 & 94.4 & 95.9 & \regaligned & \regaligned \\
        DINOv2-B/14+reg & 84.4 & 83.8 & 97.9 & 97.7 & \regaligned & \regaligned \\
        \midrule
        DINOv2-L/14 & 86.1 & 85.6 & 63.5 & 65.0 & \regunexp & \regunexp \\
        DINOv2-L/14+reg & 86.5 & 85.9 & 41.1 & 39.2 & \regunexp & \regunexp \\
        \midrule
        DINOv2-g/14 & 86.6 & 86.1 & 75.6 & 70.5 & \regaligned & \regaligned \\
        DINOv2-g/14+reg & 87.0 & 86.6 & 96.8 & 96.5 & \regaligned & \regaligned \\
        \bottomrule
    \end{tabular}
\end{table}

\subsection{Training details of the class-token SigLIP}\label{app:siglip}

To analyze whether the inverted-explainability bias in contrastive VLMs comes from the softmax in the contrastive objective, we opted to analyze a SigLIP~\citep{zhai2023sigmoid} model, as discussed in \cref{sec:vlms}.
However, the original pre-trained SigLIP encoders do not employ a class token, which is required for token-decoding explainability methods.
We therefore train OpenCLIP's ViT-B/32 image encoder, which uses a class-token, on LAION-400M~\citep{schuhmann2021laion400m} under the sigmoid loss, and report explainability of the resulting model as SigLIP-B/32.

The architecture of this SigLIP method is thus identical to the OpenCLIP ViT-B/32 models we analyze in \cref{sec:vlms,tab:vits_vlms,tab:scaling}, and only the loss differs.
We retain the same text tower from OpenCLIP, including the tokenizer.
Any difference from the CLIP models of \cref{sec:vlms,tab:vits_vlms,tab:scaling} is therefore attributable to the objective rather than to the architecture.
We train for 32 epochs on 32 NVIDIA B200 GPUs across four nodes, with a batch size of $4{,}096$ per device.
We set the learning rate to $10^{-3}$, and use $2{,}000$ warm-up steps and bfloat16 mixed precision.
Following~\citet{zhai2023sigmoid}, we also set weight decay to $10^{-4}$ and Adam's $\beta_2$ to $0.95$, in order to suppress the loss spikes that appear at this batch size.

The resulting trained model achieves $57.45\%$ top-1 and $82.92\%$ top-5 zero-shot accuracy on the ImageNet-1k validation set.

\subsection{Source of pretrained checkpoints}\label{app:source_of_ckpts}

In \cref{sec:vits,sec:vlms} we analyze pretrained ViT-based classifiers and contrastive VLMs and report their results in \cref{tab:vits_vlms}.
The \dino classifiers are the official backbones with ImageNet-1k linear heads (e.g., \texttt{dinov2\_vitb14\_lc} and \texttt{dinov2\_vitb14\_reg\_lc} for \dino-B/14 and \dino-B/14+reg respectively), loaded through \texttt{torch.hub}.
All other ViT-based classifiers were obtained from the \texttt{timm} library~\citep{rw2019timm}, with the identifiers listed in \cref{tab:timm_tags}.
For DeiT~\citep{touvron2021deit} we use the non-distilled checkpoints, which carry no distillation token and whose only classification head operates on the class-token.
For contrastive VLMs, we use the official checkpoints of CLIP~\citep{radford2021learning} and MetaCLIP~\citep{xu2024demystifying}, as well as one official OpenCLIP checkpoint~\citep{ilharco_gabriel_2021_5143773}, all loaded through the OpenCLIP library with the identifiers listed in \cref{tab:openclip_tags}.
All other OpenCLIP models are the scaling-law checkpoints of~\citet{cherti2023reproducible}, for which we use the final checkpoint of each run from \href{https://huggingface.co/laion/scaling-laws-openclip}{their Hugging Face repository}.

\begin{table}[t]
    \centering
    \caption{\textbf{\texttt{timm} checkpoints.}
    Names used in \cref{tab:vits_vlms,tab:acc} and the corresponding \texttt{timm}~\citep{rw2019timm} model identifiers (\texttt{timm} v1.0.22).
    $^\dagger$The original ViT-L/16 weights~\citep{dosovitskiy2021an} have no identifier in v1.0.22, so we use the file \href{https://github.com/huggingface/pytorch-image-models/releases/tag/v0.1-vitjx}{\texttt{jx\_vit\_large\_p16\_224-4ee7a4dc.pth}} distributed by early \texttt{timm} versions.}
    \label{tab:timm_tags}
    \small
    \begin{tabular}{ll}
        \toprule
        \textbf{Name} & \textbf{\texttt{timm} identifier} \\
        \midrule
        ViT-Ti/16 & \texttt{vit\_tiny\_patch16\_224.augreg\_in21k\_ft\_in1k} \\
        ViT-S/16 & \texttt{vit\_small\_patch16\_224.augreg\_in21k\_ft\_in1k} \\
        ViT-B/16 & \texttt{vit\_base\_patch16\_224.augreg\_in21k\_ft\_in1k} \\
        ViT-L/16 & \texttt{vit\_large\_patch16\_224.augreg\_in21k\_ft\_in1k} \\
        ViT-B/16 (orig) & \texttt{vit\_base\_patch16\_224.orig\_in21k\_ft\_in1k} \\
        ViT-B/16 (AugReg2) & \texttt{vit\_base\_patch16\_224.augreg2\_in21k\_ft\_in1k} \\
        ViT-B/16 (MIIL) & \texttt{vit\_base\_patch16\_224\_miil.in21k\_ft\_in1k} \\
        ViT-L/16 (orig)$^\dagger$ & -- \\
        ViT-S/16 (in1k) & \texttt{vit\_small\_patch16\_224.augreg\_in1k} \\
        ViT-B/16 (in1k) & \texttt{vit\_base\_patch16\_224.augreg\_in1k} \\
        ViT-B/16 (SAM) & \texttt{vit\_base\_patch16\_224.sam\_in1k} \\
        \midrule
        ViT-B/32 & \texttt{vit\_base\_patch32\_224.augreg\_in21k\_ft\_in1k} \\
        ViT-B/32 (in1k) & \texttt{vit\_base\_patch32\_224.augreg\_in1k} \\
        ViT-B/32 (SAM) & \texttt{vit\_base\_patch32\_224.sam\_in1k} \\
        \midrule
        DeiT-Ti/16 & \texttt{deit\_tiny\_patch16\_224.fb\_in1k} \\
        DeiT-S/16 & \texttt{deit\_small\_patch16\_224.fb\_in1k} \\
        DeiT-B/16 & \texttt{deit\_base\_patch16\_224.fb\_in1k} \\
        \bottomrule
    \end{tabular}
\end{table}

\begin{table}[t]
    \centering
    \caption{\textbf{OpenCLIP checkpoints.}
    Names used in \cref{tab:vits_vlms} and the corresponding OpenCLIP~\citep{ilharco_gabriel_2021_5143773} model names and pretrained tags.}
    \label{tab:openclip_tags}
    \small
    \begin{tabular}{lll}
        \toprule
        \textbf{Name} & \textbf{Model} & \textbf{Pretrained tag} \\
        \midrule
        CLIP-B/32 & \texttt{ViT-B-32} & \texttt{openai} \\
        CLIP-B/16 & \texttt{ViT-B-16} & \texttt{openai} \\
        CLIP-L/14 & \texttt{ViT-L-14} & \texttt{openai} \\
        \midrule
        MetaCLIP-B/32 & \texttt{ViT-B-32} & \texttt{metaclip\_fullcc} \\
        MetaCLIP-B/16 & \texttt{ViT-B-16} & \texttt{metaclip\_fullcc} \\
        MetaCLIP-L/14 & \texttt{ViT-L-14} & \texttt{metaclip\_fullcc} \\
        MetaCLIP-H/14 & \texttt{ViT-H-14} & \texttt{metaclip\_fullcc} \\
        MetaCLIP-G/14 & \texttt{ViT-bigG-14} & \texttt{metaclip\_fullcc} \\
        \midrule
        OpenCLIP-g/14 (LAION2B-34B) & \texttt{ViT-g-14} & \texttt{laion2b\_s34b\_b88k} \\
        \bottomrule
    \end{tabular}
\end{table}
\clearpage

\section{Additional results of real-world models}\label{app:more_results_real}

\subsection{ViT classifiers}\label{app:more_results_vits}

\begin{figure}
    \centering
    \includegraphics[width=\linewidth]{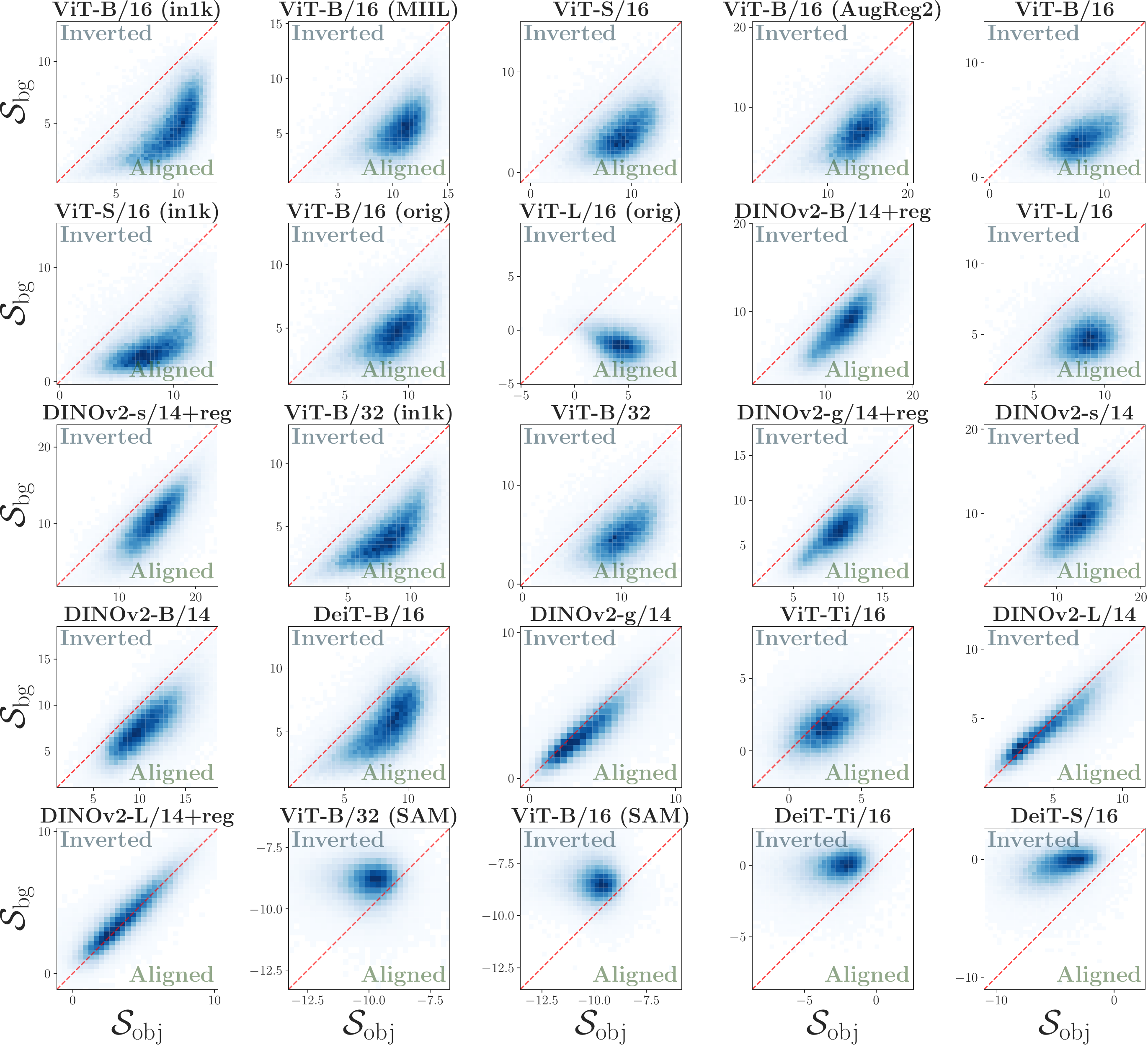}
    \caption{
    \textbf{Aligned-explainability bias of real-world ViT-based classifiers.} 
    We plot 2D histograms of $\sbg$ and $\sobj$ scores calculated over all of PartImageNet++~\citep{li2024partimagenet++} for 25 ViT classifiers of different architectures. ViT classifiers exhibit a strong bias toward aligned-explainability. Some models still converge to the unexplainability or even the inverted-explainability regime, irrespective of accuracy or scale.
    }
    \label{fig:more_real_word_vits}
\end{figure}

\paragraph{Additional 2D histograms.}
We include in \cref{fig:more_real_word_vits} 2D-histograms of $\sobj$ and $\sbg$ calculated over all of PartImageNet++~\citep{li2024partimagenet++} for all ViT-based classifiers that appear in \cref{tab:vits_vlms}. As can be seen, ViTs have a strong aligned-explainability bias. Still, some models, such as \dino-L/14, ViT-B/16 trained with sharpness-aware minimization or some DeiT configurations exhibit non-aligned explainability.

\paragraph{Accuracy of ViTs.}
Furthermore, we include in \cref{tab:acc} the top-1 classification accuracy over PartImageNet++~\citep{li2024partimagenet++} of the ViTs from \cref{tab:vits_vlms}. All models have high top-1 accuracy, even in the inverted-explainability regime or the unexplainable regime.

\begin{table}[t]
    \centering
    \caption{\textbf{Top-1 accuracy of real-world ViT classifiers across explainability regimes.}
    Scores and accuracy values are computed over 
    PartImageNet++~\citep{li2024partimagenet++}.
    Explainability regimes are marked according to \cref{sec:defs}.
    Dashed rules mark the boundary between explainability regimes.
    The aligned-explainability bias of ViT classifiers is unrelated to classification accuracy, as all models achieve high top-1 accuracy values.}
    \label{tab:acc}
    \begin{tabular}{lccc}
        \toprule
        \textbf{Classifier} & \textbf{Top-1 Accuracy (\%)} & $\mathcal{A}$ (\%) & \textbf{Exp.} \\
        \midrule
         ViT-B/16 (in1k) & 91.9 & 98.7 & \regaligned \\
        ViT-B/16 (MIIL) & 93.5 & 98.6 & \regaligned \\
        ViT-S/16 & 88.3 & 98.4 & \regaligned \\
        ViT-B/16 (AugReg2) & 93.2 & 98.4 & \regaligned \\
        ViT-B/16 & 91.2 & 98.4 & \regaligned \\
        ViT-S/16 (in1k) & 89.7 & 98.4 & \regaligned \\
        ViT-B/16 (orig) & 90.3 & 98.2 & \regaligned \\
        ViT-L/16 (orig) & 93.2 & 98.1 & \regaligned \\
        DINOv2-B/14+reg & 90.9 & 97.9 & \regaligned \\
        ViT-L/16 & 93.2 & 97.8 & \regaligned \\
        DINOv2-s/14+reg & 88.1 & 97.7 & \regaligned \\
        ViT-B/32 (in1k) & 87.3 & 97.7 & \regaligned \\
        ViT-B/32 & 89.1 & 97.5 & \regaligned \\
        DINOv2-g/14+reg & 93.5 & 96.8 & \regaligned \\
        DINOv2-s/14 & 88.4 & 96.7 & \regaligned \\
        DINOv2-B/14 & 92.1 & 94.4 & \regaligned \\
        DeiT-B/16 & 95.6 & 92.2 & \regaligned \\
        DINOv2-g/14 & 93.1 & 75.6 & \regaligned \\
        ViT-Ti/16 & 82.2 & 70.5 & \regaligned \\
        \cdashedlinelong
        DINOv2-L/14 & 92.6 & 63.5 & \regunexp \\
        DINOv2-L/14+reg & 93.2 & 41.1 & \regunexp \\
        \cdashedlinelong
        ViT-B/32 (SAM) & 89.1 & 14.4 & \reginverted \\
        ViT-B/16 (SAM) & 92.4 & 6.4 & \reginverted \\
        DeiT-Ti/16 & 78.9 & 4.1 & \reginverted \\
        DeiT-S/16 & 90.6 & 1.4 & \reginverted \\
        \bottomrule
    \end{tabular}
\end{table}

\paragraph{Effect of activation norms on inverted classifiers.}

For contrastive VLMs, our explainability scores are computed by applying the projection matrix directly on the patch tokens (after the normalization layer, as described in \cref{app:technical_details_real}) and computing the cosine similarity to the embedded text prompt.
However, for classifiers, after applying the classification head on the patch tokens we use the logits directly as our scores, as is standard for token-decoding explainability.
Thus, one might suspect that the inverted-explainability results observed in \cref{tab:vits_vlms} stem from activation norm irregularities.
Specifically, when logits are negative, as observed in SAM-trained ViTs and DeiT-Ti/S in \cref{fig:more_real_word_vits}, 
lower $\gA$ scores could hypothetically be caused by higher activation norms of object patches, instead of due to true directional misalignment.

To rule this out, we evaluate normalized cosine similarity scores for these checkpoints.
Specifically, we compute the cosine similarity $\rw_c^\top\vx / (\|\rw_c\|\,\|\vx\|)$ between the patch tokens (after the normalization layer) and the label class weight vector, over all of PartImageNet++.
This measures only the angular alignment of the classification head to the token representation. 
We note that here we omit the scalar bias $b_c$ to isolate directional alignment, and that all DeiT models, as described in \cref{app:source_of_ckpts}, are non-distilled.
The results in \cref{tab:sam_norm_audit} show that all four models remain strongly inverted-explainable under cosine similarity.
Particularly, the mean norms of both object and background patches are very similar in all models.
Therefore, the observed explainability regime of these models is indeed directional, rather than an artifact of token norm scaling.

\begin{table}
\centering
\small
\caption{\textbf{Effect of activation norm on inverted ViT-based classifiers.} The mean activation norms for both object and background patch tokens are very similar, which means our $\gS$ computation approach of directly using the logits leads to similar results as using normalized cosine similarity decoding. In particular, both exhibit the same inverted-explainability, which rules out norm-scaling artifacts.}
\label{tab:sam_norm_audit}
\begin{tabular}{llcccc}
\toprule
\makecell{\textbf{Model}} & 
\makecell{\textbf{$\gS(\cdot)$ Approach}} & 
\makecell{\textbf{Mean $\sobj$}} & 
\makecell{\textbf{Mean $\sbg$}} & 
\makecell{\textbf{$\mathcal{A}$ (\%)}} & 
\makecell{\textbf{Mean Norm} \\ \textbf{$(\|\vx_{\textnormal{obj}}\| / \|\vx_{\textnormal{bg}}\|)$}} \\
\midrule
\multirow{2}{*}{ViT-B/16 (SAM)} 
 & Logits              & $-10.07$  & $-8.33$  & $6.4\%$ & \multirow{2}{*}{$10.99 / 10.77$} \\
 & Cosine Similarity   & $-0.35$  & $-0.30$  & $8.4\%$ &  \\
\midrule
\multirow{2}{*}{ViT-B/32 (SAM)} 
 & Logits              & $-10.09$  & $-8.87$  & $14.4\%$ & \multirow{2}{*}{$13.22 / 13.16$} \\
 & Cosine Similarity   & $-0.33$  & $-0.29$  & $15.2\%$ &  \\
 \midrule
\multirow{2}{*}{DeiT-Ti/16} 
 & Logits              & $-2.99$  & $-0.13$  & $4.1\%$ & \multirow{2}{*}{$25.88 / 24.44$} \\
 & Cosine Similarity   & $-0.17$  & $-0.002$  & $4.2\%$ &  \\
 \midrule
\multirow{2}{*}{DeiT-S/16} 
 & Logits              & $-4.73$  & $-0.57$  & $1.4\%$ & \multirow{2}{*}{$23.06 / 21.54$} \\
 & Cosine Similarity   & $-0.23$  & $-0.02$  & $1.3\%$ &  \\
\bottomrule
\end{tabular}
\end{table}

\subsection{Contrastive VLMs}\label{app:more_results_vlms}

\begin{figure}
    \centering
    \includegraphics[width=\linewidth]{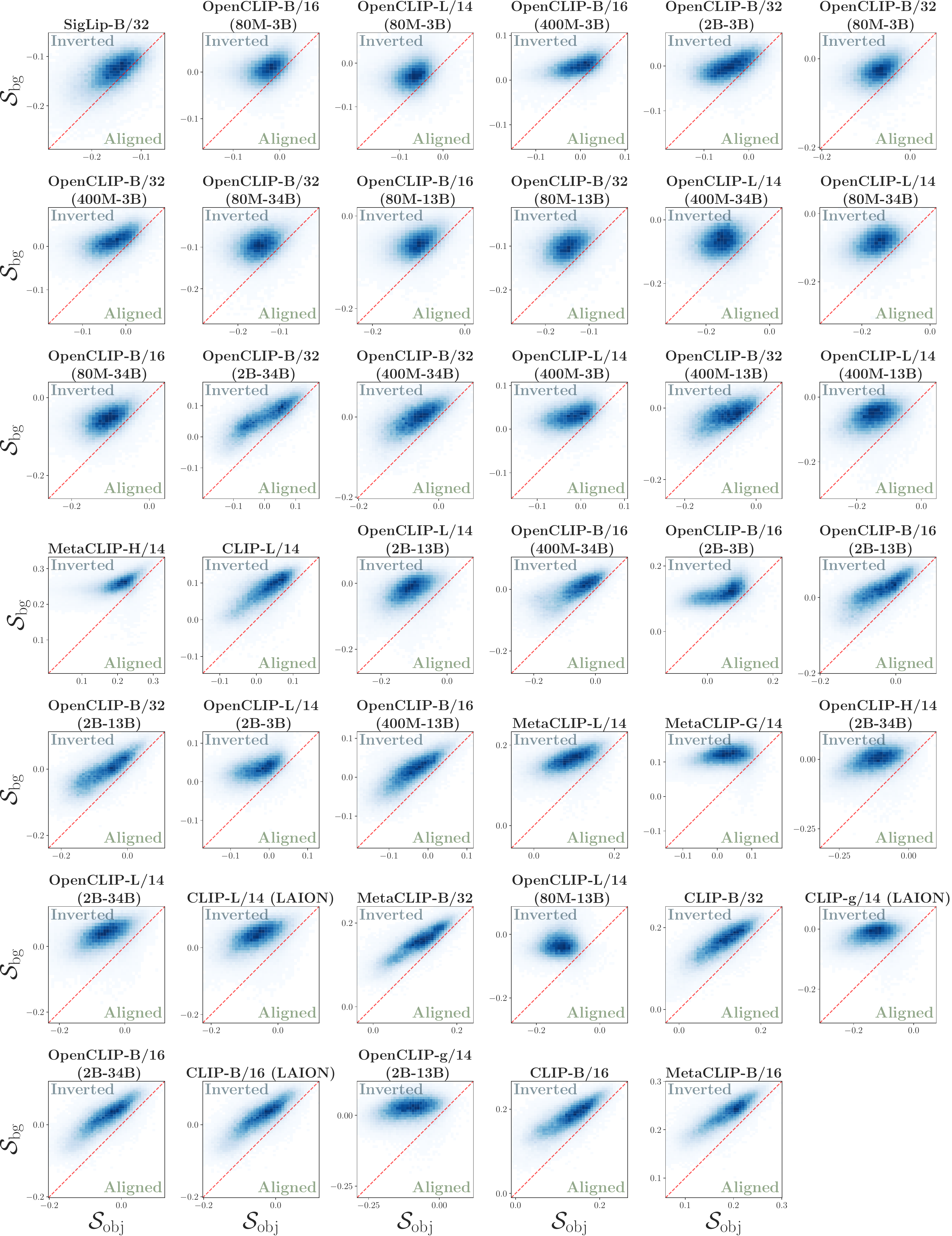}
    \caption{\textbf{Inverted-explainability bias of real-world contrastive VLMs.} 
    We plot 2D histograms of $\sbg$ and $\sobj$ scores calculated over all of PartImageNet++~\citep{li2024partimagenet++} for 41 contrastive VLMs of different architectures. 
    Contrastive VLMs exhibit a clear bias toward inverted-explainability, irrespective of accuracy or scale.}
    \label{fig:more_real_world_vlms}
\end{figure}

\paragraph{Scaling analysis.}
In \cref{sec:vlms,app:siglip} we verified that inverted-explainability is not a simple consequence of the loss. Here, we further check whether explainability might be a product of scale.
Specifically, we analyze the $29$ pre-trained checkpoints of \citet{cherti2023reproducible}.
These checkpoints were trained from scratch across varying training-set sizes, different numbers of samples seen during training (equivalent to different numbers of epochs), and various architecture sizes. 
We report the results of $\gA$ in \cref{tab:scaling}. 
Indeed, every one of the models exhibits inverted-explainability, with $\mathcal{A}$ ranging from $2.0\%$ to $13.8\%$.
More striking is the apparent direction of the inverted-explainability trend: within a specific architecture, larger training sets and longer training promote lower $\gA$, which means a stronger bias toward inverted-explainability in the model.
We note that models marked with a dagger (\dag) used smaller learning rate and batch sizes in their training. These runs diverge from the clear scaling trend by a small margin. For example, in B-16, the single model trained with a smaller learning rate and batch size has lower $\gA$ than its counterparts that were trained with more seen samples, or with a larger training set.
The explainability 2D histograms of all models appear in \cref{fig:more_real_world_vlms}.

\begin{table}[t]
    \centering
    \small
    \caption{\textbf{Inverted-explainability across the CLIP scaling grid.}
    We compute $\gA$ scores for all models of \citet{cherti2023reproducible} over all of
    PartImageNet++~\citep{li2024partimagenet++}.
    The models vary in architecture size and training-set size (rows) and in the number of samples seen during training (columns), which is equivalent to varying training epochs.
    $\dag$ marks the seven runs trained at $5\times10^{-4}$ rather than the $10^{-3}$ used by the other $22$ runs, and with a smaller batch-size.
    All $29$ models exhibit inverted-explainability.
    Moreover, for a specific architecture, larger training sets and longer training promote a stronger inverted-explainability bias.
    }
    \label{tab:scaling}
    \begin{tabular}{llrrr}
        \toprule
        & & \multicolumn{3}{c}{Samples seen} \\
        \cmidrule(lr){3-5}
        Arch. & Data & 3B & 13B & 34B \\
        \midrule
        B-32 & 80M & 9.7\dag & 8.3\dag & 9.1 \\
        & 400M & 9.4 & 5.7 & 6.2\dag \\
        & 2B & 11.3 & 3.4\dag & 6.3 \\
        \midrule
        B-16 & 80M & 13.8 & 8.5 & 7.0 \\
        & 400M & 11.5 & 3.3\dag & 3.5 \\
        & 2B & 3.5 & 3.5 & 2.2 \\
        \midrule
        L-14 & 80M & 12.1 & 2.5 & 7.0 \\
        & 400M & 6.1 & 5.2 & 7.7 \\
        & 2B & 3.4 & 3.5 & 2.8 \\
        \midrule
        H-14 & 2B & -- & -- & 2.9\dag \\
        \midrule
        g-14 & 2B & -- & 2.0\dag & -- \\

        \bottomrule
    \end{tabular}
\end{table}

\paragraph{Additional 2D histograms.}
We include in \cref{fig:more_real_world_vlms} 2D-histograms of $\sobj$ and $\sbg$ calculated over all of PartImageNet++~\citep{li2024partimagenet++} for all contrastive VLMs that appear in \cref{tab:vits_vlms,tab:scaling}. As can be seen, contrastive VLMs all exhibit inverted-explainability, even at the largest training scales.

\begin{figure}
    \centering
    \includegraphics[width=\linewidth]{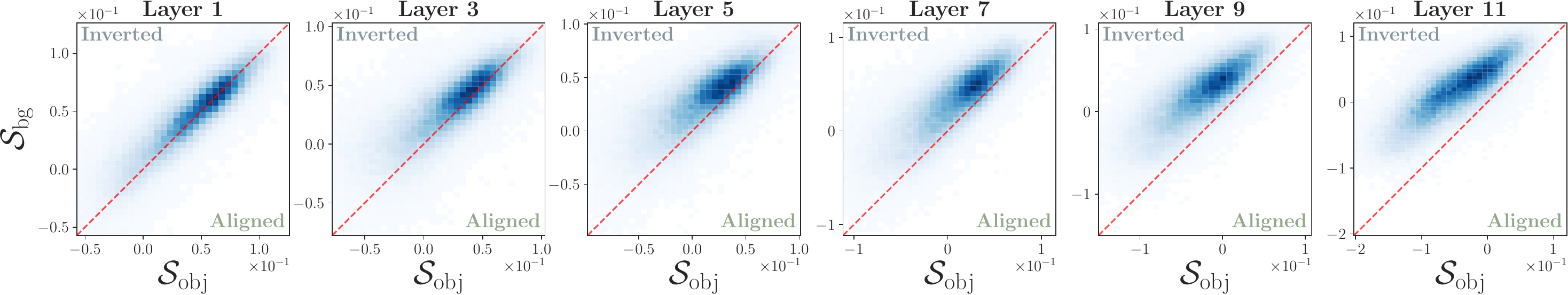}
    \caption{\textbf{Evolution of inverted explainability in CLIP.} 
    We plot 2D histograms of $\sbg$ and $\sobj$ scores calculated over all of PartImageNet++~\citep{li2024partimagenet++} as they evolve through the model's 12 layers. While the early layers are unexplainable, as the features evolve they exhibit more inversion. 
    }
    \label{fig:many_layers_clip}
\end{figure}

\paragraph{Evolution of inverted explainability.}
We include in \cref{fig:many_layers_clip} a plot depicting how inverted explainability evolves through the layers of a CLIP model. Deeper layers are more inverted-explainable than shallower ones, similarly to the phenomenon seen in ViT classifiers in \cref{fig:many_layers_vit}.

\subsection{Explainability of the last layer}\label{app:last_layer}

For models whose last-layer patch tokens are not affected by the loss, all our analyses use the penultimate layer (\cref{sec:defs}).
In these models, the classification head or the contrastive projection, for classifiers and contrastive models respectively, usually reads only the class token of the last layer.
Thus, the training loss places no constraint on the last-layer patch tokens.
Specifically, the patch tokens at the end of the last layer receive no gradient from the loss.
They are computed by the last layer, whose attention and MLP weights are trained only through the class token.
In contrast, the patch tokens at the end of the penultimate layer are optimized directly, since the class token attends to them in the last layer.
However, the decoded tokens of our toy model in \cref{sec:theory} are of the same kind, as its head is also applied only to the class token.
Therefore, for completeness, we examine here the explainability regimes of the last layer patch tokens.

\Cref{tab:last_layer} compares $\gA$ at the penultimate and last layers for all ViT classifiers and contrastive VLMs of \cref{tab:vits_vlms} whose last-layer patch tokens are not affected by the loss.
Specifically, all models from \cref{tab:vits_vlms} are included with the exception of \dino, whose head reads the last layer patch tokens and for which we already reported the last layer in \cref{tab:vits_vlms}.
Indeed, the aligned-explainability and inverted-explainability exhibited in all models remain the same when examining the last layer.
This indicates that the explainability regime is a property of the representation that carries through to the final representations in the last layer, even though these representations do not participate in the loss.
It also supports our analysis of tokens outside the loss path in the toy model, since real-world models exhibit a clear regime on such tokens as well.

\begin{table}[t]
    \centering
    \small
    \setlength{\tabcolsep}{3.5pt}
    \caption{\textbf{Explainability at the penultimate and last layers.}
    $\gA$ is computed over all of PartImageNet++ at the penultimate layer (Penult.), which we use throughout the paper, and at the last layer (Last), whose patch tokens are not affected by the loss.
    Explainability regimes are marked according to \cref{sec:defs}.
    Each model is in the same regime at both layers.
    Dashed rules mark the boundary between regimes.
    \dino is excluded, as its head reads the last-layer patch tokens, and thus our results in \cref{tab:vits_vlms} already use its last layer decoding.
    }
    \label{tab:last_layer}
    \begin{tabular}[t]{lccc}
        \toprule
        & \multicolumn{2}{c}{$\gA$ (\%)} & \\
        \cmidrule(lr){2-3}
        \textbf{Classifier} & Penult. & Last & \textbf{Exp.} \\
        \midrule
ViT-B/16 (in1k)    & 98.7 & 98.8 & \regaligned \\
ViT-B/16 (MIIL)    & 98.6 & 96.3 & \regaligned \\
ViT-S/16           & 98.4 & 95.5 & \regaligned \\
ViT-B/16 (AugReg2) & 98.4 & 88.0 & \regaligned \\
ViT-B/16           & 98.4 & 93.8 & \regaligned \\
ViT-S/16 (in1k)    & 98.4 & 98.3 & \regaligned \\
ViT-B/16 (orig)    & 98.2 & 92.0 & \regaligned \\
ViT-L/16 (orig)    & 98.1 & 97.8 & \regaligned \\
ViT-L/16           & 97.8 & 98.2 & \regaligned \\
ViT-B/32 (in1k)    & 97.7 & 97.7 & \regaligned \\
ViT-B/32           & 97.5 & 94.5 & \regaligned \\
DeiT-B/16          & 92.2 & 95.1 & \regaligned \\
ViT-Ti/16          & 70.5 & 70.5 & \regaligned \\
\cdashedline
ViT-B/32 (SAM) & 14.4 & 9.8 & \reginverted \\
ViT-B/16 (SAM) & 6.4 &  11.2 & \reginverted \\
DeiT-Ti/16     & 4.1 &  4.5 & \reginverted \\
DeiT-S/16      & 1.4 &  2.2 & \reginverted \\
        \bottomrule
    \end{tabular}
    \,
    \begin{tabular}[t]{lccc}
        \toprule
        & \multicolumn{2}{c}{$\gA$ (\%)} & \\
        \cmidrule(lr){2-3}
        \textbf{Contrastive VLM} & Penult. & Last & \textbf{Exp.} \\
        \midrule
SigLIP-B/32                   & 22.5 & 14.6 & \reginverted \\
OpenCLIP-B/16 (LAION80M-3B)   & 13.8 & 16.5 & \reginverted \\
OpenCLIP-L/14 (LAION80M-3B)   & 12.1 & 12.2 & \reginverted \\
OpenCLIP-B/16 (LAION400M-3B)  & 11.5 & 13.0 & \reginverted \\
OpenCLIP-B/16 (LAION80M-13B)  & 8.5  & 11.0 & \reginverted \\
OpenCLIP-L/14 (LAION80M-34B)  & 7.0  & 11.9 & \reginverted \\
OpenCLIP-B/16 (LAION80M-34B)  & 7.0  & 16.8 & \reginverted \\
OpenCLIP-B/32 (LAION2B-34B)   & 6.3  & 12.5 & \reginverted \\
MetaCLIP-H/14                 & 4.6  & 4.1 & \reginverted \\
CLIP-L/14                     & 4.3  & 10.6 & \reginverted \\
OpenCLIP-B/16 (LAION400M-34B) & 3.5  & 8.9 & \reginverted \\
OpenCLIP-B/16 (LAION2B-3B)    & 3.5  & 4.4 & \reginverted \\
OpenCLIP-B/16 (LAION2B-13B)   & 3.5  & 15.0 & \reginverted \\
OpenCLIP-L/14 (LAION2B-3B)    & 3.4  & 3.8 & \reginverted \\
OpenCLIP-B/16 (LAION400M-13B) & 3.3  & 15.7 & \reginverted \\
MetaCLIP-L/14                 & 3.2  & 2.6 & \reginverted \\
MetaCLIP-G/14                 & 3.0  & 3.0 & \reginverted \\
OpenCLIP-H/14 (LAION2B-34B)   & 2.9  & 4.3 & \reginverted \\
OpenCLIP-L/14 (LAION2B-34B)   & 2.8  & 13.3 & \reginverted \\
MetaCLIP-B/32                 & 2.6  & 6.4 & \reginverted \\
CLIP-B/32                     & 2.3  & 9.9 & \reginverted \\
OpenCLIP-g/14 (LAION2B-34B)   & 2.2  & 3.1 & \reginverted \\
OpenCLIP-B/16 (LAION2B-34B)   & 2.2  & 15.2 & \reginverted \\
OpenCLIP-g/14 (LAION2B-13B)   & 2.0  & 1.9 & \reginverted \\
CLIP-B/16                     & 2.0  & 14.3 & \reginverted \\
MetaCLIP-B/16                 & 1.5  & 8.5 & \reginverted \\
        \bottomrule
    \end{tabular}
\end{table}

\clearpage
\section{Technical details of the controlled experiments}\label{app:technical_details_synth}

\subsection{Data construction details}\label{app:data_construction}

\paragraph{OBNR details.}
As described in \cref{sec:controlled_setting}, the background noise sampled in both datasets is scaled to control the OBNR.
Specifically, in each case we first normalize the background noise according to the covariance of the object patch. This way, the OBNR is equal to $1$. Then, we can control the OBNR simply by multiplying it by constant factors, $\alpha_\rvn$.

For the supervised case (\cref{eq:synth_patch_vit}), the initial scaling of the background noise is $\Tr\{\Cov(\rvx_c)\}=\sqrt{\gamma^2 r + 1}$, since $\mA$ is a randomly sampled $r$-rank matrix. Then, we multiply the noise by $\alpha_\rvn\in[0.05,0.1,0.3,0.6,1,2,4,8,20,50,100]$.
In the contrastive case (\cref{eq:synth_patch_vlm}), the background noise is similarly normalized to $\Tr\{\Cov(\rvx)\}=\alpha_\rvx^2 + 1$.  Here, we multiply by $\alpha_\rvn\in[0.1,0.2,0.3,0.4,0.5,0.65,0.8,2,4]$.

\subsection{Synthetic architecture details}\label{app:synth_arch_details}

In both experiments, we use small ViTs with $12$ layers and a patch size of $2$. As the input images are of size $20\times20$, this results in a $10\times10$ token grid. Additionally, the model always uses an embedding dimension of $24$, with $3$ heads.

\paragraph{ViT classifiers.} All classifiers were trained using Adam~\citep{kingma2015adam} 
with a learning rate of $0.02$, $\beta_1=0.9,\beta_2=0.999$, no weight decay, and a cosine learning rate scheduler.
Additionally, all models were trained with a batch size of $2{,}048$ for $300$ epochs.
As the dataset samples a completely new sample at each training step, we set $100$ training steps per epoch, which means an entire training run covers over $61$ million samples.

\paragraph{Contrastive VLMs.} All contrastive runs were trained using AdamW~\citep{loshchilov2018decoupled} with a learning rate of $0.01$, $\beta_1=0.9,\beta_2=0.98$, a weight decay of $0.025$, and a cosine learning rate scheduler. 
Additionally, all models were trained with a batch size of $50{,}000$  for $2{,}000$ epochs, where $100$ steps served as warm-up.
The dataset samples new samples at each training step here as well, so we set $20$ training steps per epoch, meaning the entire training run covers $2$ billion samples. 
For the text tower, we use $4$ layers with $2$ heads, and an embedding dimension of $16$.
The contrastive loss, the projections, and the learned logit scale are unmodified from OpenCLIP~\citep{ilharco_gabriel_2021_5143773}.

\clearpage

\section{Controlled experiments detail}\label{app:results_synth}

\subsection{Explainability regime variance for different training seeds}\label{app:synth_seeds}

\begin{figure}
    \centering
    \begin{subfigure}{0.49\textwidth}
        \includegraphics[width=\linewidth]{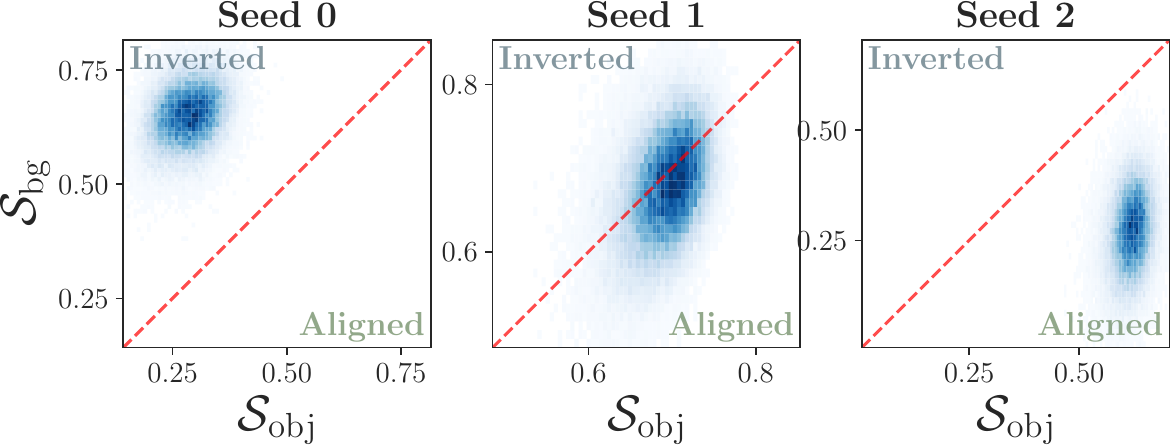}
        \caption{\textbf{$\alpha_\rvx=1, \alpha_\rvy=1, \alpha_\rvn=2$}}
        \label{fig:training_seeds_vlms_2}
    \end{subfigure}
    \hfill
    \begin{subfigure}{0.49\textwidth}
        \includegraphics[width=\linewidth]{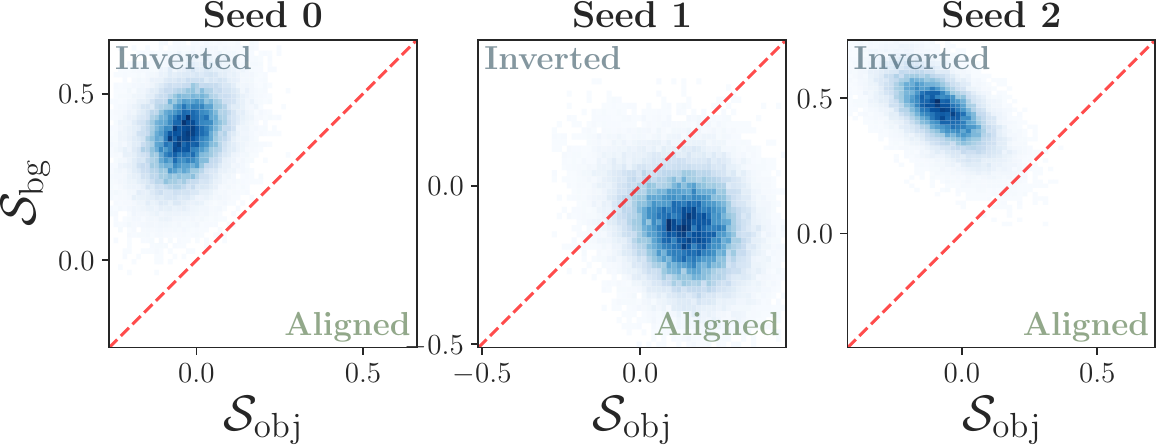}
        \caption{\textbf{$\alpha_\rvx=4, \alpha_\rvy=1, \alpha_\rvn=0.2$}}
        \label{fig:training_seeds_vlms_0.2}
    \end{subfigure}
    \hfill
    \begin{subfigure}{0.49\textwidth}
        \includegraphics[width=\linewidth]{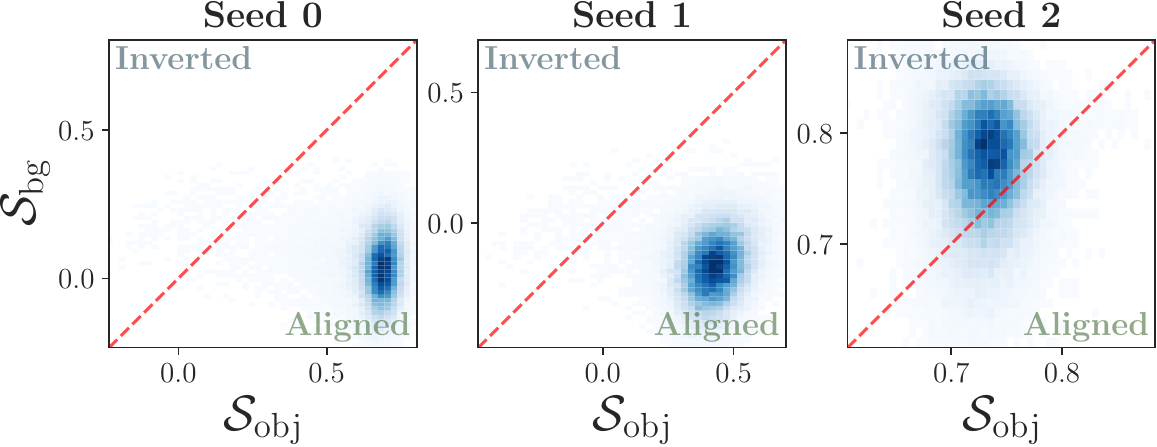}
        \caption{\textbf{$\alpha_\rvx=2, \alpha_\rvy=2, \alpha_\rvn=0.8$}}
        \label{fig:training_seeds_vlms_0.8}
    \end{subfigure}
    \hfill
    \begin{subfigure}{0.49\textwidth}
        \includegraphics[width=\linewidth]{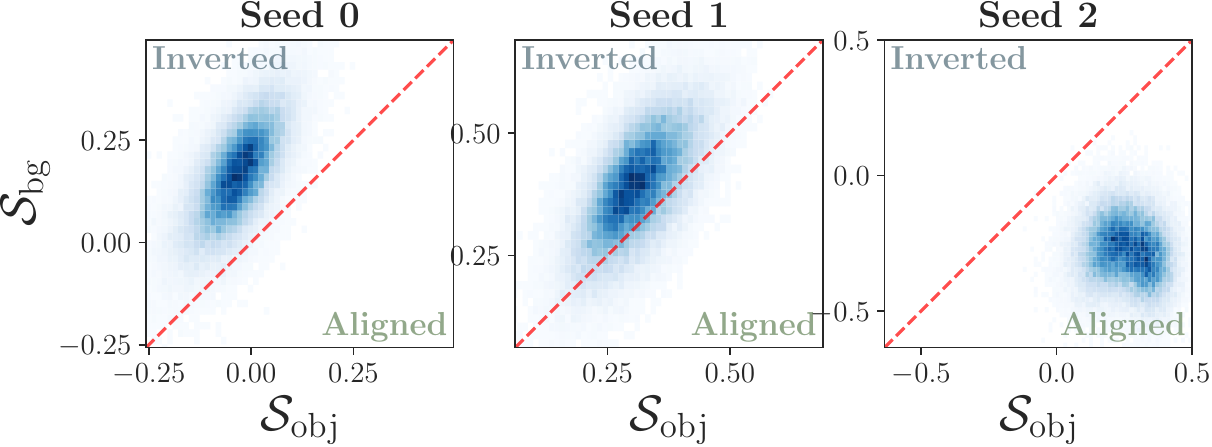}
        \caption{\textbf{$\alpha_\rvx=2, \alpha_\rvy=1, \alpha_\rvn=0.3$}}
    \end{subfigure}
    \hfill
    \begin{subfigure}{0.49\textwidth}
        \includegraphics[width=\linewidth]{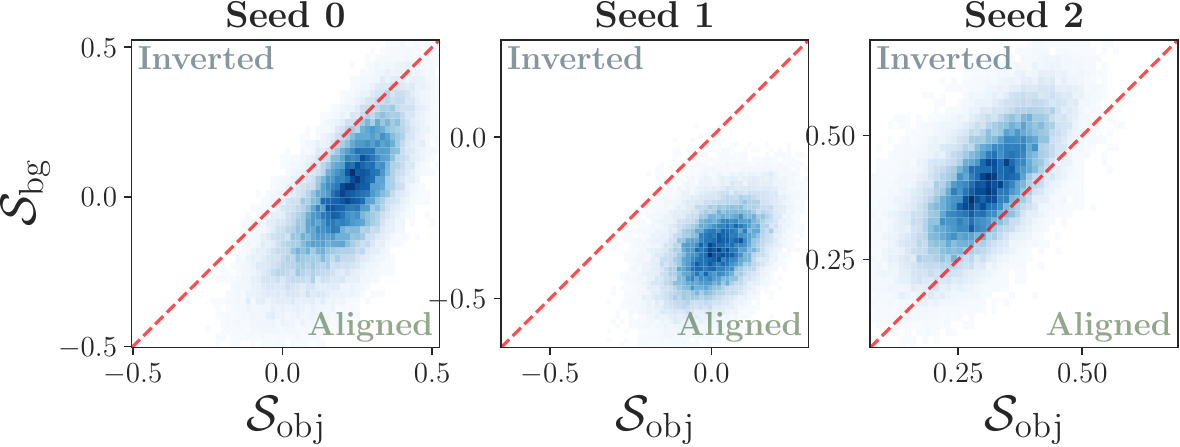}
        \caption{\textbf{$\alpha_\rvx=4, \alpha_\rvy=4, \alpha_\rvn=0.65$}}
    \end{subfigure}
    \hfill
    \begin{subfigure}{0.49\textwidth}
        \includegraphics[width=\linewidth]{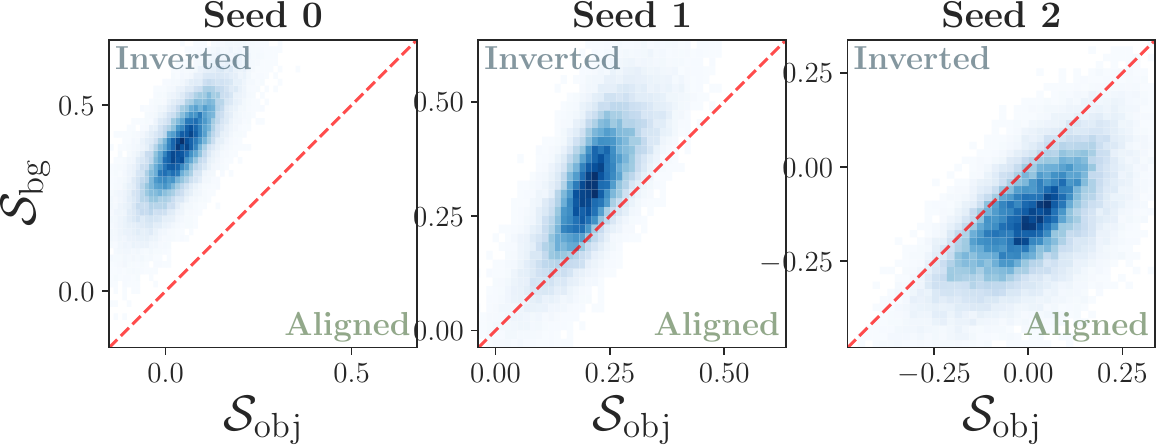}
        \caption{\textbf{$\alpha_\rvx=1, \alpha_\rvy=4, \alpha_\rvn=0.4$}}
    \end{subfigure}
    \caption{
    \textbf{Different training seeds of the exact same configuration exhibiting different explainability regimes.}
     2D histograms of $\sbg$ and $\sobj$ scores of $6$ different contrastive configurations, spanning different $\alpha_\rvx,\alpha_\rvy,\alpha_\rvn$ (and thus different OBNRs), trained over our synthetic dataset. 
     While all configurations induce a bias toward some explainability regime, as seen in \cref{fig:synthetic_totals},
     different training seeds of the exact same data distribution can still yield very different explainability regimes.
     For example, when $\alpha_\rvn=0.8$ then $\textnormal{OBNR}\approx1$ and the contrastive models have a bias toward aligned-explainability, yet one of the three training seeds of \cref{fig:training_seeds_vlms_0.8} leads to inverted-explainability.
     Additionally, when $\alpha_\rvn=0.2$ the OBNR is high and the natural bias is toward inverted-explainability, however one of the seeds in \cref{fig:training_seeds_vlms_0.2} cause the model to converge to aligned-explainability.
     In some configurations, such as in \cref{fig:training_seeds_vlms_2}, the three seeds even led to the three possible regimes: inverted-explainability, aligned-explainability, and unexplainability.
     Indeed, explainability is not an inherent property of transformers, but an outcome contigent on the problem setting and optimization.
    }
    \label{fig:training_seeds_vlms}
\end{figure}

\Cref{sec:controlled} demonstrates that in our controlled experiments, the bias toward different explainability regimes is heavily dictated by the OBNR. However, we note that this is a bias and not a strict deterministic outcome. 
In \cref{fig:training_seeds_vlms} we include 2D histograms of 6 different contrastive configurations, each for 3 different seeds.
While, as we showed in \cref{fig:synthetic_totals}, different OBNR values lead to a bias of some explainability regime,
different training seeds of the exact same data distribution can still yield very different explainability regimes.
For example, for contrastive models, when the OBNR is close to $1$ (\eg, when $\alpha_\rvn=0.8$) models have a bias toward aligned-explainability.
However, when training on three different seeds, one converged to an inverted-explainable solution (\cref{fig:training_seeds_vlms_0.8}).
In the opposite case, when the OBNR is high (\eg when $\alpha_\rvn=0.2$) the natural bias is toward inverted-explainability.
Still, one of the training seeds for such a configuration causes the model to converge to aligned-explainability (\cref{fig:training_seeds_vlms_0.2}).
In fact, we can even find configurations such as \cref{fig:training_seeds_vlms_2} where the three seeds lead to all three possible regimes: inverted-explainability, aligned-explainability, and unexplainability.

\begin{figure}
    \centering
    \begin{subfigure}{0.49\textwidth}
        \includegraphics[width=\linewidth]{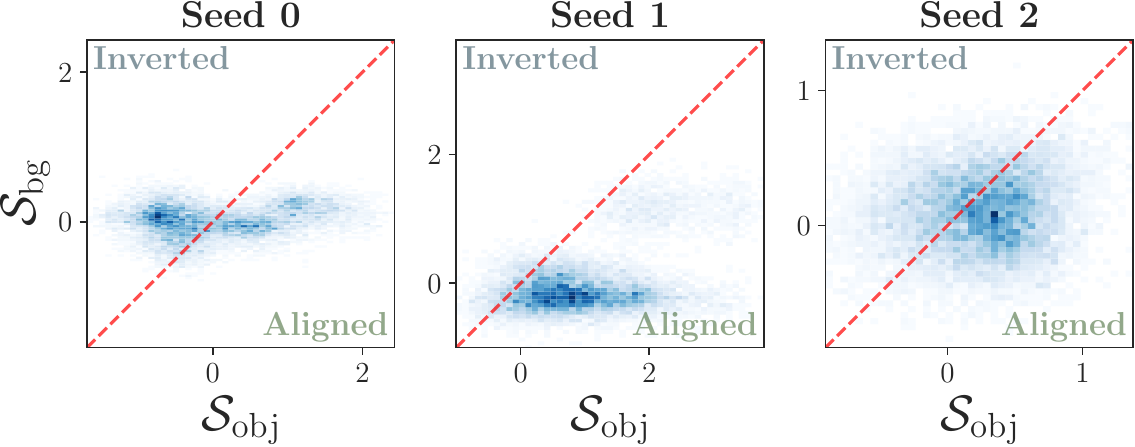}
        \caption{\textbf{$\gC=10, \gamma=0.75, \alpha_\rvn=100$}}
        \label{fig:training_seeds_vits_100}
    \end{subfigure}
    \hfill
    \begin{subfigure}{0.49\textwidth}
        \includegraphics[width=\linewidth]{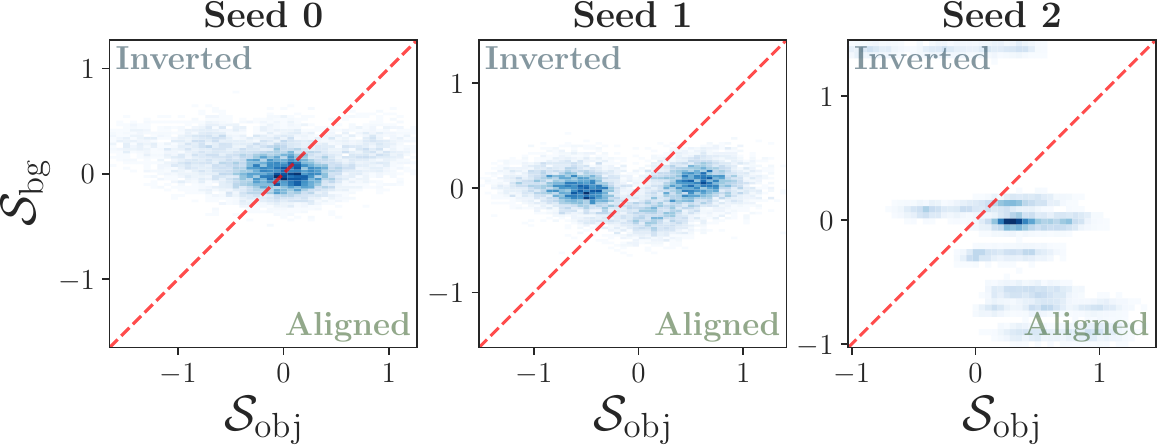}
        \caption{\textbf{$\gC=10, \gamma=0.5, \alpha_\rvn=50$}}
        \label{fig:training_seeds_vits_50}
    \end{subfigure}
    \caption{
    \textbf{Different training seeds of the exact same configuration exhibiting different explainability regimes.}
     2D histograms of $\sbg$ and $\sobj$ scores of $2$ different ViT-classifier configurations, spanning different $\gC,\gamma,\alpha_\rvn$ (and thus different OBNRs), trained over our synthetic dataset. 
     While a configuration induces a bias toward some explainability regime, as seen in \cref{fig:synthetic_totals},
     different training seeds of the exact same data distribution can still yield very different explainability patterns.
     For example, in \cref{fig:training_seeds_vits_100}, as the OBNR is low the bias is toward aligned-explainability (\cref{fig:synthetic_totals}). However, one seed exhibits at least partial inverted-explainability.
     Additionally, in \cref{fig:training_seeds_vits_50} we can see some seeds exhibit different explainability for different classes of the model, similar to the phenomenon in \dino-L/14 (\cref{fig:dino-animals}). Here, the model might cluster classes, and converge to have aligned or inverted explainability depending on the classes cluster, similarly to the animals and non-animals clusters of \dino.
     This reinforces how explainability is not a static property of transformers.
    }
    \label{fig:training_seeds_vits}
\end{figure}

In addition, in \cref{fig:training_seeds_vits} we include 2D histograms of 2 different ViT-based classifiers, each for 3 different seeds.
Here, as well, different training seeds can lead to very different and interesting explainability patterns.
For instance, \cref{fig:training_seeds_vits_50} demonstrates a similar phenomenon as seen in \dino-L/14 (\cref{fig:dino-animals}). 
Specifically, different class clusters can converge to have different explainability regime, similar to the animals and non-animals clusters of \dino.
This reinforces our claim on the elusive nature of explainability. It is not some inherent property of transformers, but an outcome that is not to be assumed a-prior.

\subsection{Additional results}\label{app:synth_more_results}

We conduct additional experiments on more complicated tasks for the ViTs, both in the supervised and the contrastive configurations.
For ViT classifiers, we repeat our experiment of \cref{sec:controlled_setting}, while this time also randomly sampling the rank-$r$ $\mA$ matrix per class. As the covariances of the objects can vary widely between each class, we limit $\gamma$ to $[0.5,0.75]$, yielding a total of 44 supervised configurations, each trained under three seeds. 
For the self-supervised experiments, we reduce the number of layers in the ViT to 6, and test on $\alpha_\rvn\in[0.05, 0.1, 0.2, 0.3, 0.5, 0.65, 0.8, 4.0]$, which yields 72 configurations, each trained under three seeds.

\begin{figure}
    \centering
    \includegraphics[width=\linewidth]{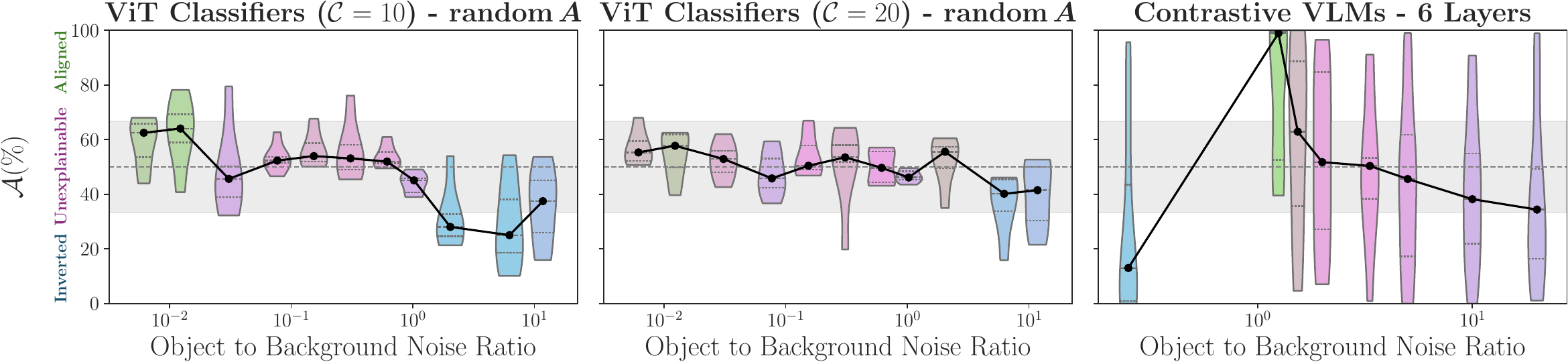}
    \caption{\textbf{Explainability trends of additional controlled ViT experiments.}
    Violin plots of $\gA$ scores of 66 classifiers with 10 (left) or 20 (middle) classes, and 144 contrastive models (right), trained on our synthetic datasets, for different OBNR values. 
    The ViT classifiers are trained on a synthetic dataset variation where $\mA$ is sampled differently per class.
    The contrastive models are trained using ViTs with 6 layers. Both make the task harder for the ViT to accomplish.
    The violins are colored gradually according to their medians (indicated with black dots), with blue, green, and purple denoting inverted-explainability, aligned-explainability and unexplainability, respectively. 
    The black lines connect the medians, the dashed rules mark $\gA=50\%$, and the gray background indicates the unexplainability regime.
    The trends of explainability biases seen in \cref{fig:synthetic_totals} are present here as well.
    Specifically, in ViTs, low or high OBNR values induce aligned or inverted explainability biases, respectively, while $\textnormal{OBNR}\approx1$ promotes unexplainability.
    Contrastive VLMs instead converge to inverted-explainability at both extremes, and to aligned-explainability at $\textnormal{OBNR}\approx1$.
    Despite these biases, ViTs occasionally converge outside the dominant regime for the same data distribution.
    }
    \label{fig:synth_totals_more}
\end{figure}

In \cref{fig:synth_totals_more} we plot the distribution of $\gA$ across these trained models for varying OBNR values. 
The explainability trends are exactly as in \cref{fig:synthetic_totals}. Specifically, in ViT-classifiers very high or very low OBNR values induce explainability, with higher background noise levels causing alignment bias and lower background noise levels leading to inversion, while $\textnormal{OBNR}\approx1$ lie in the unexplainable regime. 
Here, as well a smaller choice classes amount leads to a sharper explainability trend.

The explainability behavior seen in \cref{fig:synthetic_totals} seen for contrastive VLMs, also remains the same for a smaller ViT. Low and high OBNR values promote an inverted-explainability bias, and $\textnormal{OBNR}\approx1$ promotes aligned-explainability.

\subsection{Performance of trained models}\label{app:synth_perf}

\begin{figure}
    \centering
    \includegraphics[width=\linewidth]{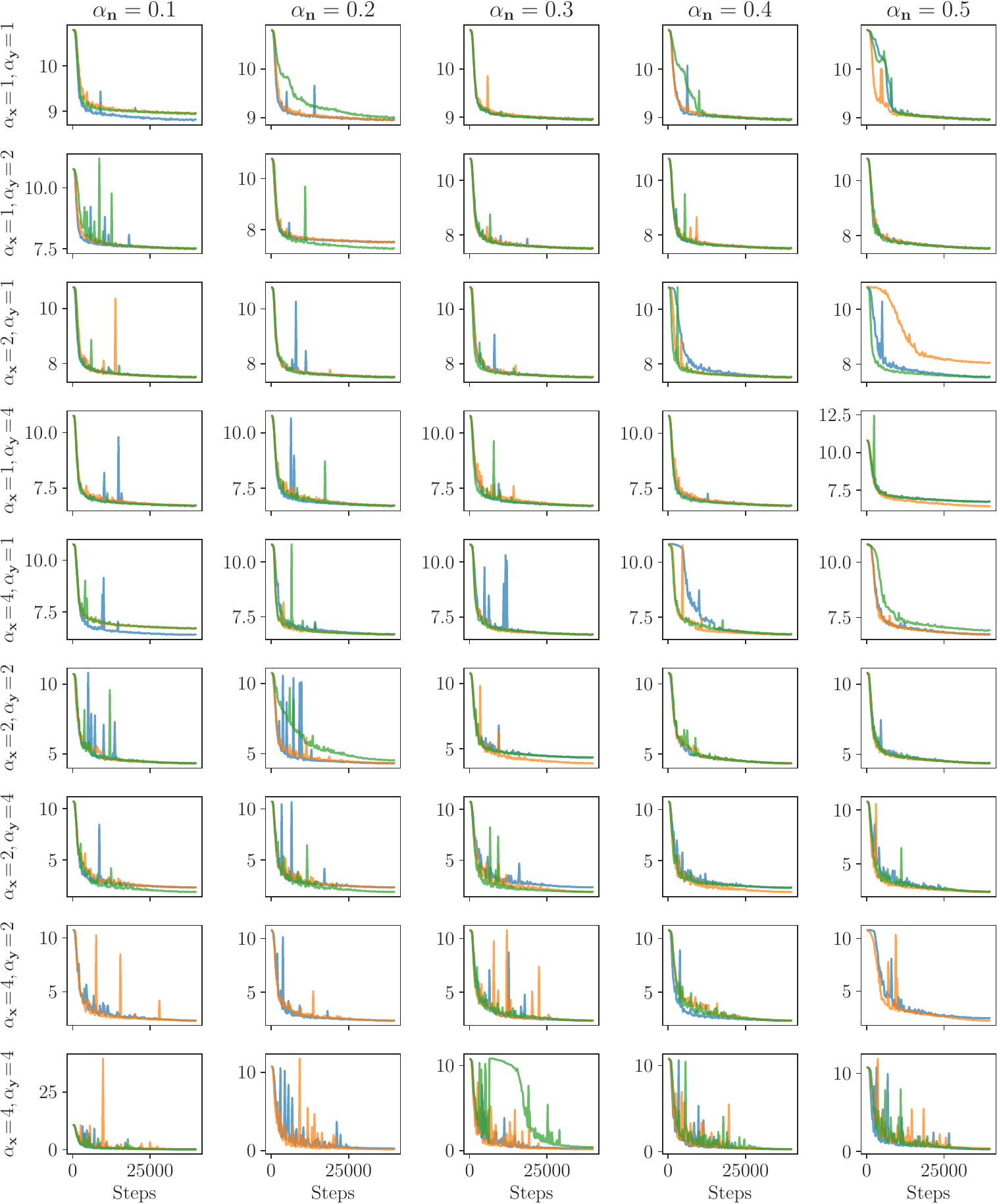}
    \caption{\textbf{Training loss plots of the contrastive VLMs from \cref{fig:synthetic_totals}, for $\alpha_\rvn\in[0.1,0.5]$.}
    All reported VLMs converge similarly when trained on the same task. Therefore, the trends of explainability biases  we see are not a product of some failure of the model, but rather an implicit bias of the problem setting.
    }
    \label{fig:vlm_acc_loss1}
\end{figure}

\begin{figure}
    \centering
    \includegraphics[width=0.9\linewidth]{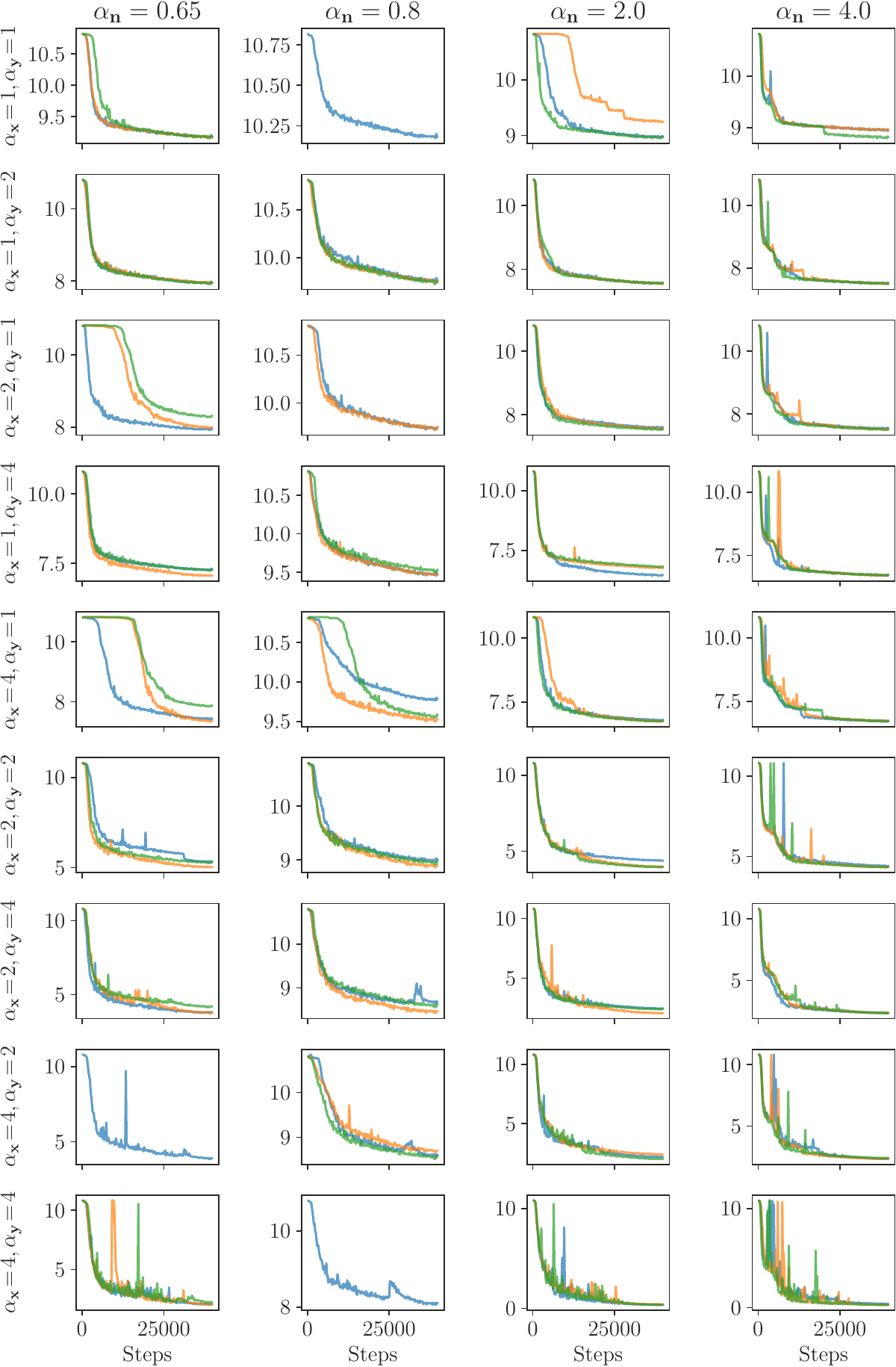}
    \caption{\textbf{Training loss plots of the contrastive VLMs from \cref{fig:synthetic_totals}, for $\alpha_\rvn\in[0.65,4.0]$.}
    All reported VLMs converge similarly when trained on the same task. Therefore, the trends of explainability biases  we see are not a product of some failure of the model, but rather an implicit bias of the problem setting.
    }
    \label{fig:vlm_acc_loss2}
\end{figure}

\begin{figure}
    \centering
    \includegraphics[width=\linewidth]{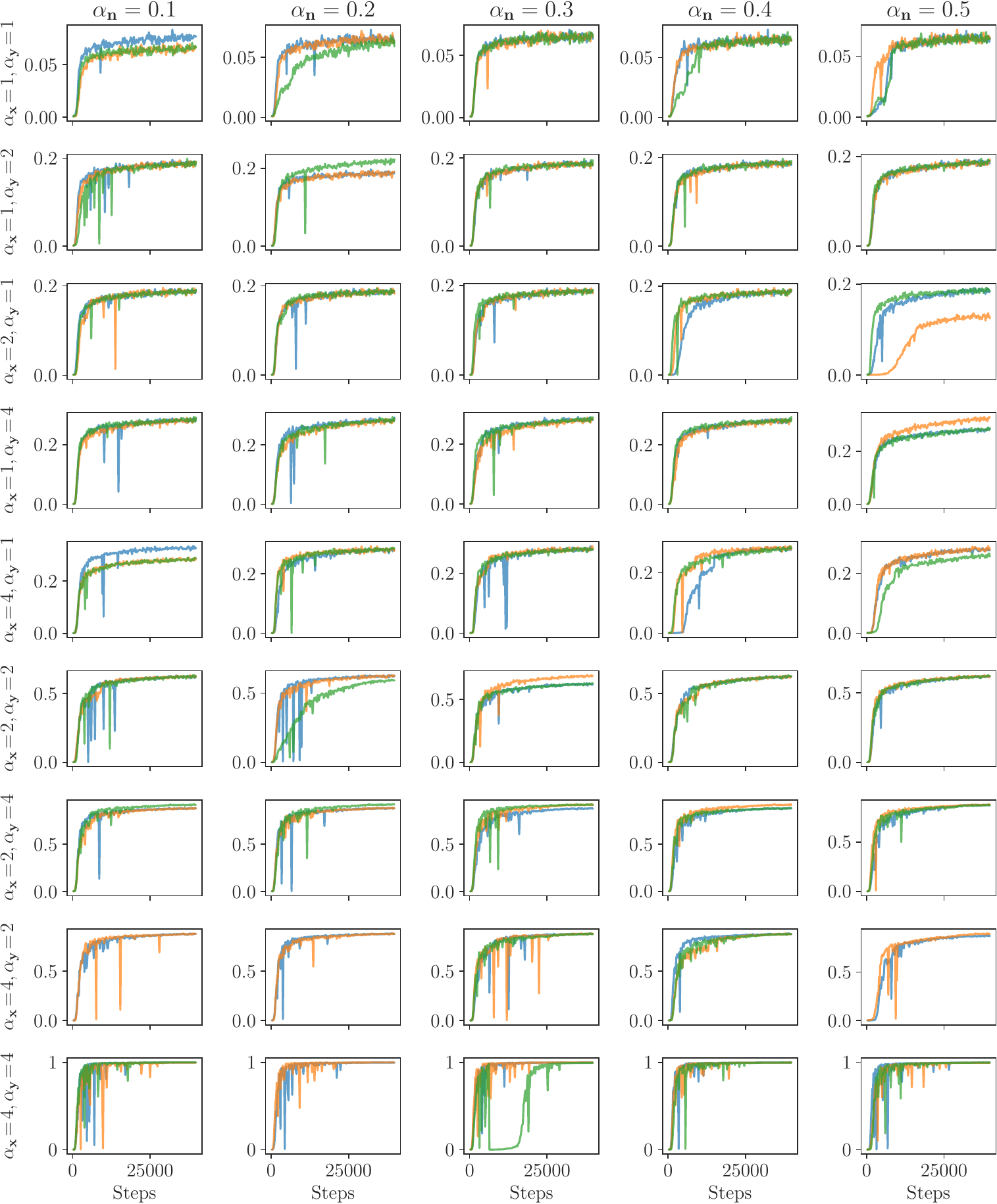}
    \caption{\textbf{Text-to-image retrieval recall at $5$ (R@5) plots of the contrastive VLMs from \cref{fig:synthetic_totals}, for $\alpha_\rvn\in[0.1,0.5]$, during training.}
    All reported VLMs converge to perform similarly when trained on the same task. Therefore, the trends of explainability biases  we see are not a product of some failure of the model, but rather an implicit bias of the problem setting.
    }
    \label{fig:vlm_acc_r51}
\end{figure}

\begin{figure}
    \centering
    \includegraphics[width=0.9\linewidth]{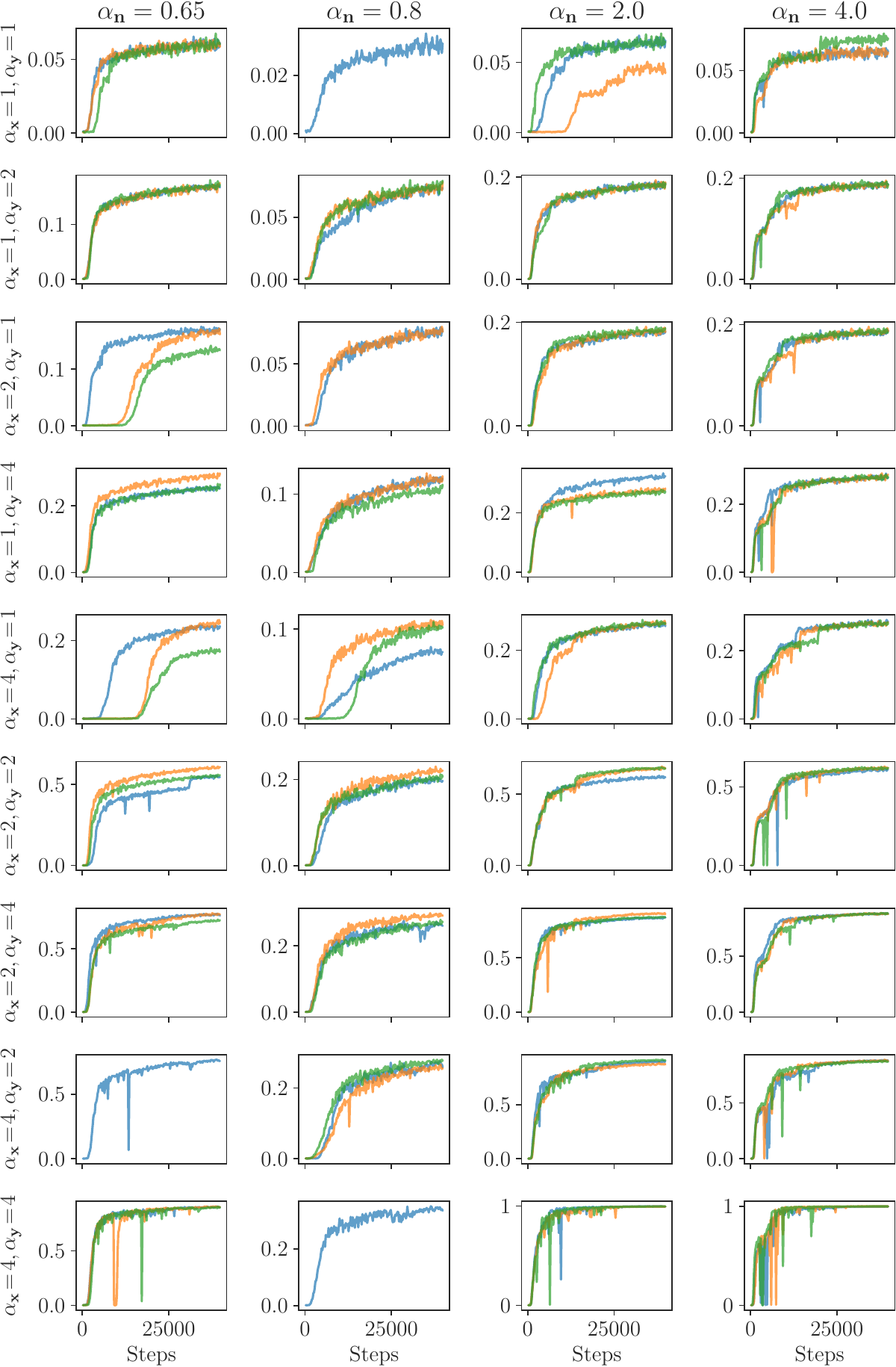}
    \caption{\textbf{Text-to-image retrieval recall at $5$ (R@5) plots of the contrastive VLMs from \cref{fig:synthetic_totals}, for $\alpha_\rvn\in[0.65,4.0]$, during training.}
    All reported VLMs converge to perform similarly when trained on the same task. Therefore, the trends of explainability biases  we see are not a product of some failure of the model, but rather an implicit bias of the problem setting.
    }
    \label{fig:vlm_acc_r52}
\end{figure}

\begin{figure}
    \centering
    \includegraphics[width=\linewidth]{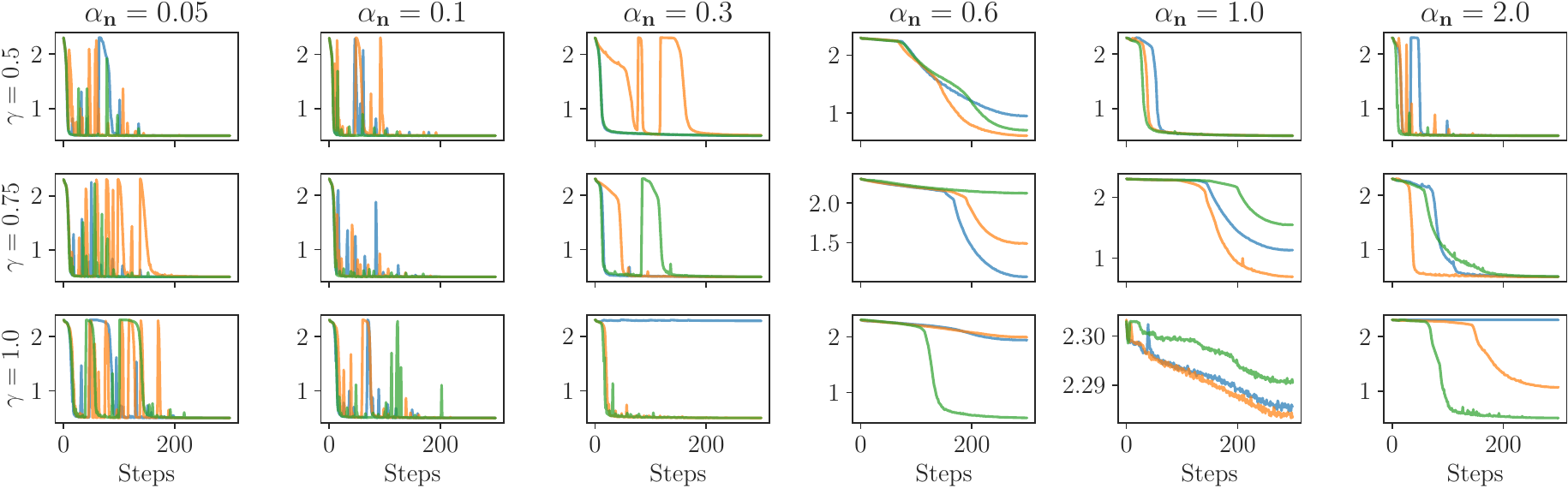}
    \caption{\textbf{Training loss plots of the ViT classifiers with $10$ classes from \cref{fig:synthetic_totals}, for $\alpha_\rvn\in[0.05,2.0]$.}
    All reported ViT classifiers converge similarly when trained on the same task. Therefore, the trends of explainability biases  we see are not a product of some failure of the model, but rather an implicit bias of the problem setting.
    }
    \label{fig:vit10_acc_loss1}
\end{figure}

\begin{figure}
    \centering
    \includegraphics[width=\linewidth]{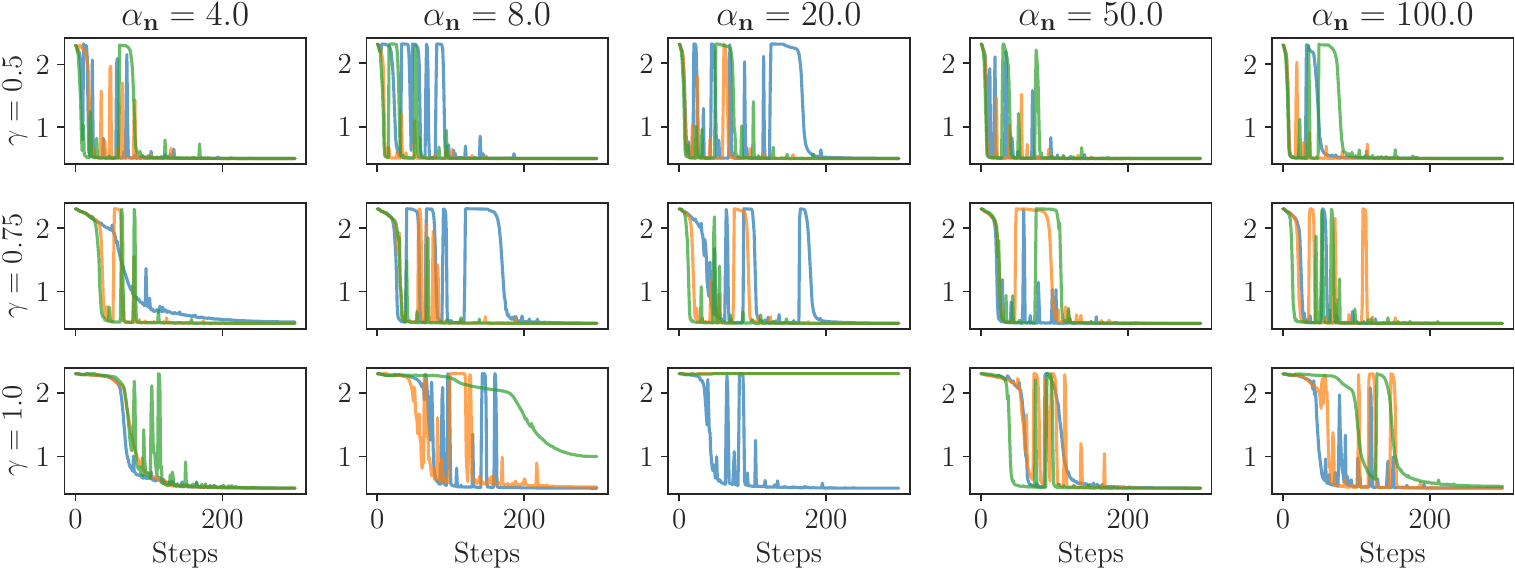}
    \caption{\textbf{Training loss plots of the ViT classifiers with $10$ classes from \cref{fig:synthetic_totals}, for $\alpha_\rvn\in[4,100]$.}
    All reported ViT classifiers converge similarly when trained on the same task. Therefore, the trends of explainability biases  we see are not a product of some failure of the model, but rather an implicit bias of the problem setting.
    }
    \label{fig:vit10_acc_loss2}
\end{figure}

\begin{figure}
    \centering
    \includegraphics[width=\linewidth]{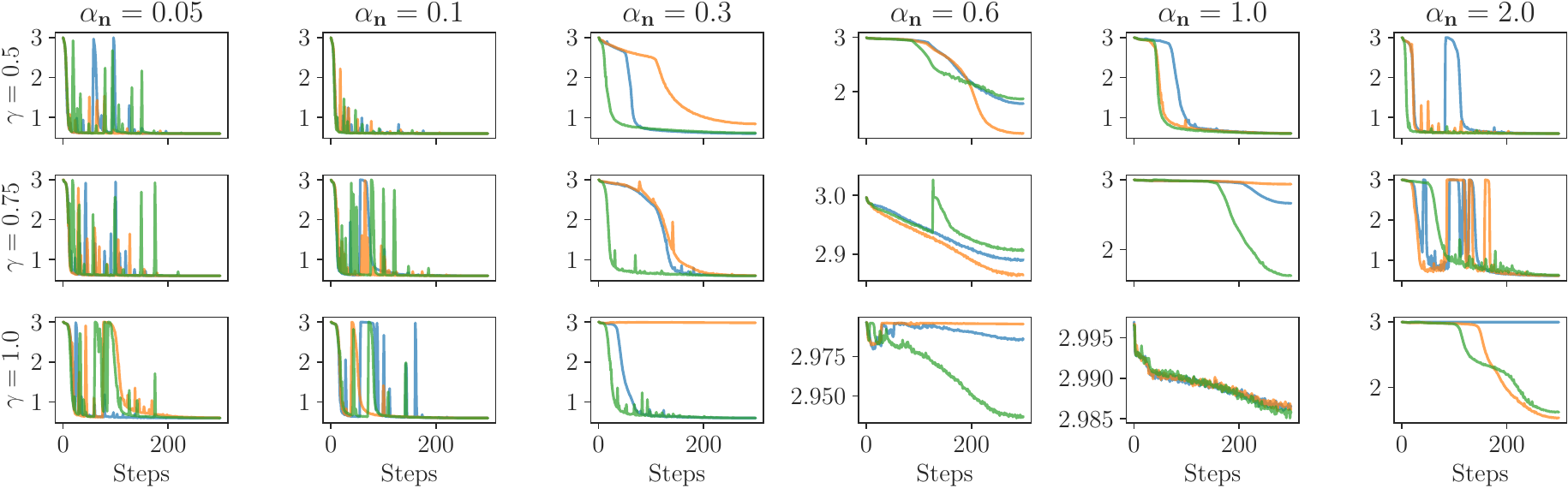}
    \caption{\textbf{Training loss plots of the ViT classifiers with $20$ classes from \cref{fig:synthetic_totals}, for $\alpha_\rvn\in[0.05,2.0]$.}
    All reported ViT classifiers converge similarly when trained on the same task. Therefore, the trends of explainability biases  we see are not a product of some failure of the model, but rather an implicit bias of the problem setting.
    }
    \label{fig:vit20_acc_loss1}
\end{figure}

\begin{figure}
    \centering
    \includegraphics[width=\linewidth]{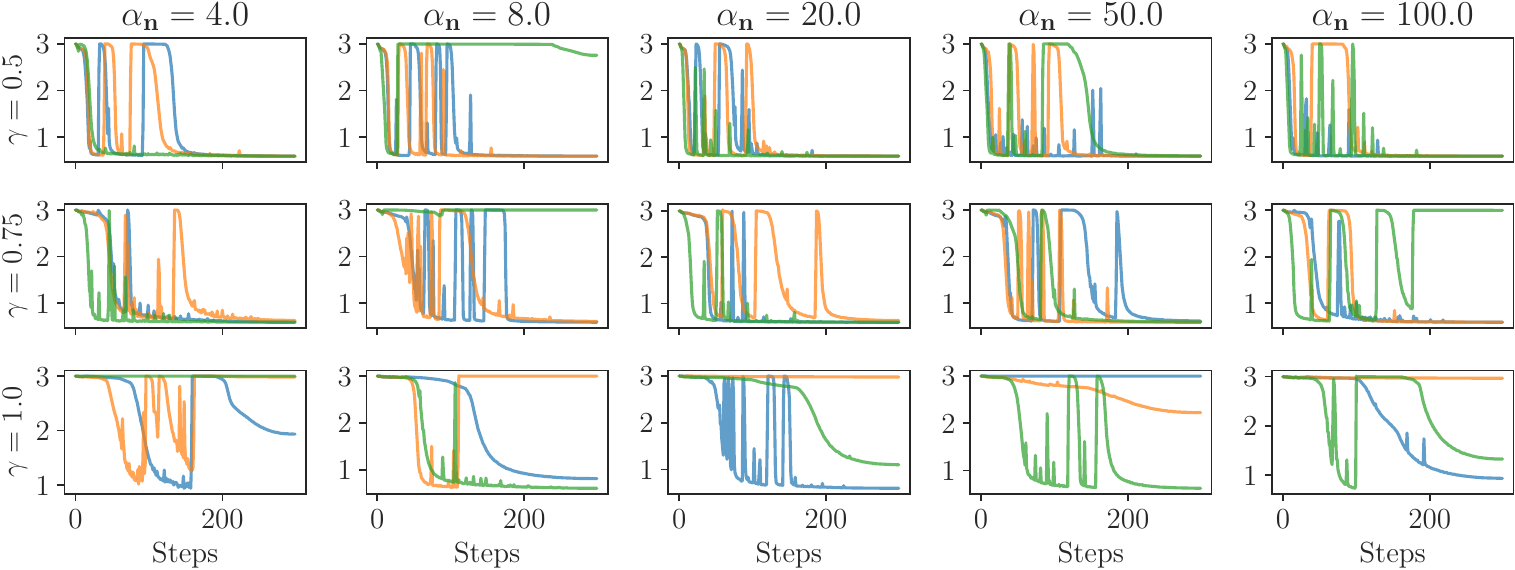}
    \caption{\textbf{Training loss plots of the ViT classifiers with $20$ classes from \cref{fig:synthetic_totals}, for $\alpha_\rvn\in[4,100]$.}
    All reported ViT classifiers converge similarly when trained on the same task. Therefore, the trends of explainability biases  we see are not a product of some failure of the model, but rather an implicit bias of the problem setting.
    }
    \label{fig:vit20_acc_loss2}
\end{figure}

\begin{figure}
    \centering
    \includegraphics[width=\linewidth]{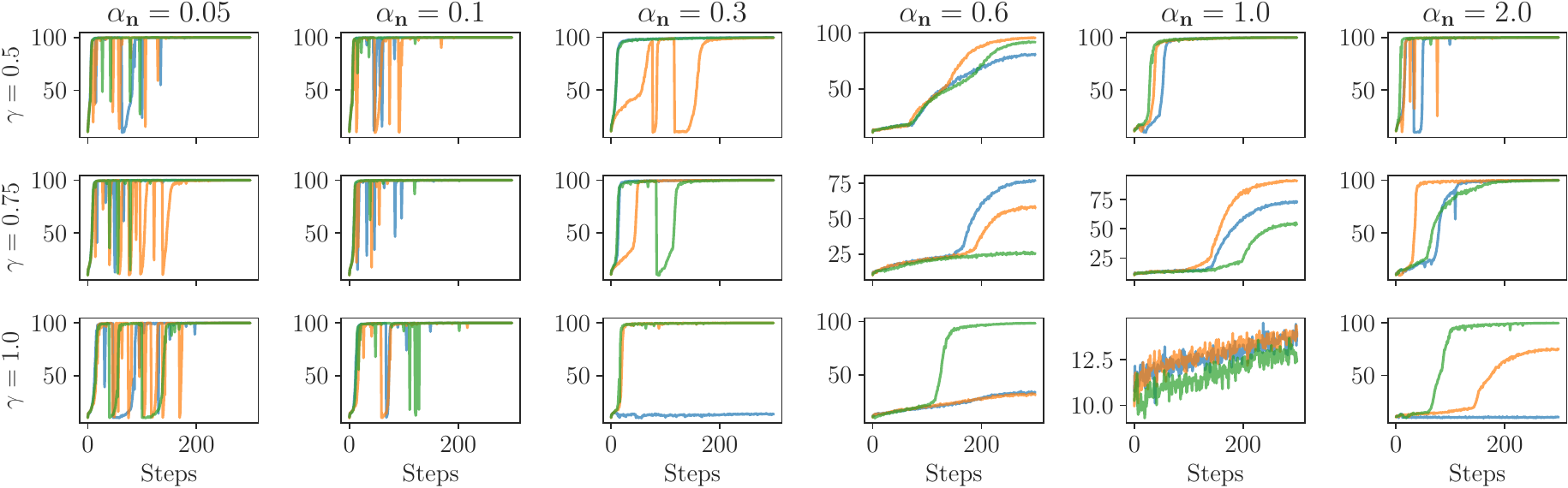}
    \caption{\textbf{Top-1 accuracy plots of the ViT classifiers with $10$ classes from \cref{fig:synthetic_totals}, for $\alpha_\rvn\in[0.05,2.0]$, during training.}
    All reported ViT classifiers converge similarly to high accuracy when trained on the same task. Therefore, the trends of explainability biases  we see are not a product of some failure of the model, but rather an implicit bias of the problem setting.
    }
    \label{fig:vit10_acc_top1}
\end{figure}

\begin{figure}
    \centering
    \includegraphics[width=\linewidth]{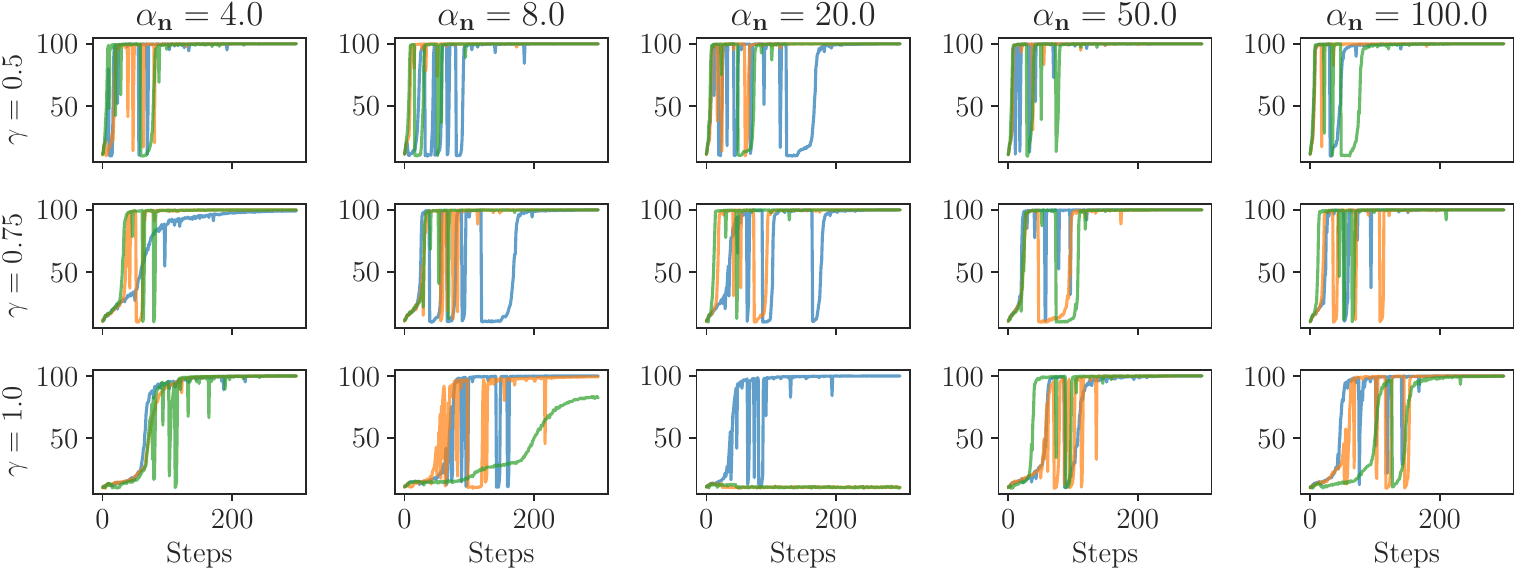}
    \caption{\textbf{Training loss plots of the ViT classifiers with $10$ classes from \cref{fig:synthetic_totals}, for $\alpha_\rvn\in[4,100]$, during training.}
    All reported ViT classifiers converge similarly to high accuracy when trained on the same task. Therefore, the trends of explainability biases  we see are not a product of some failure of the model, but rather an implicit bias of the problem setting.
    }
    \label{fig:vit10_acc_top2}
\end{figure}

\begin{figure}
    \centering
    \includegraphics[width=\linewidth]{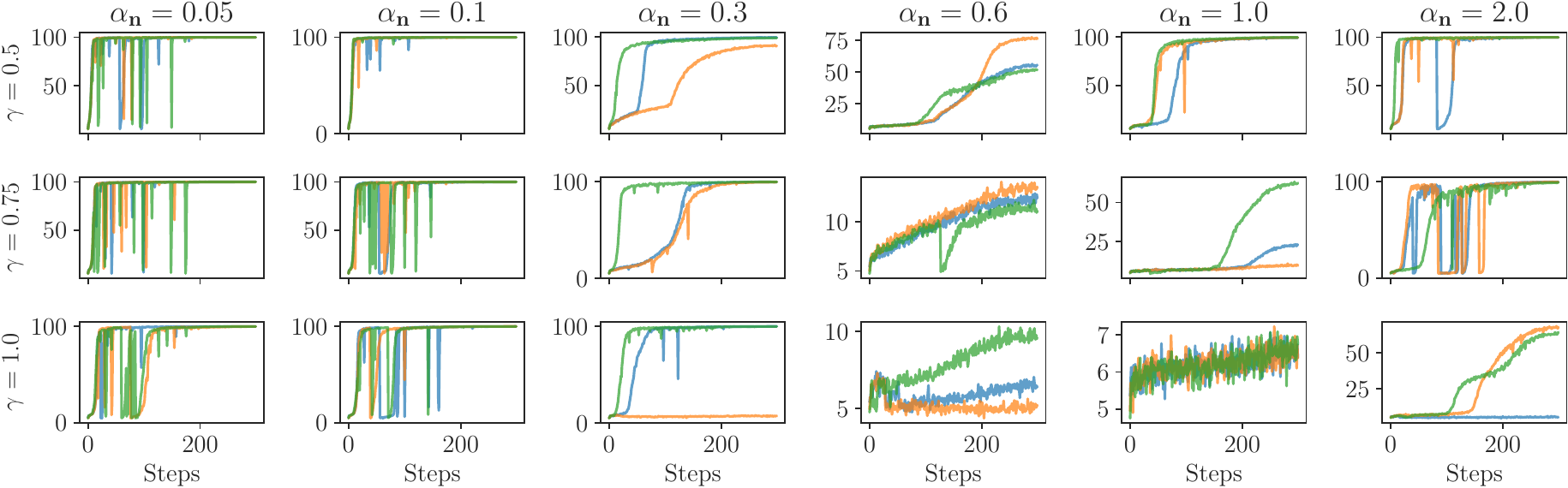}
    \caption{\textbf{Training loss plots of the ViT classifiers with $20$ classes from \cref{fig:synthetic_totals}, for $\alpha_\rvn\in[0.05,2.0]$, during training.}
    All reported ViT classifiers converge similarly to high accuracy when trained on the same task. Therefore, the trends of explainability biases  we see are not a product of some failure of the model, but rather an implicit bias of the problem setting.
    }
    \label{fig:vit20_acc_top1}
\end{figure}

\begin{figure}
    \centering
    \includegraphics[width=\linewidth]{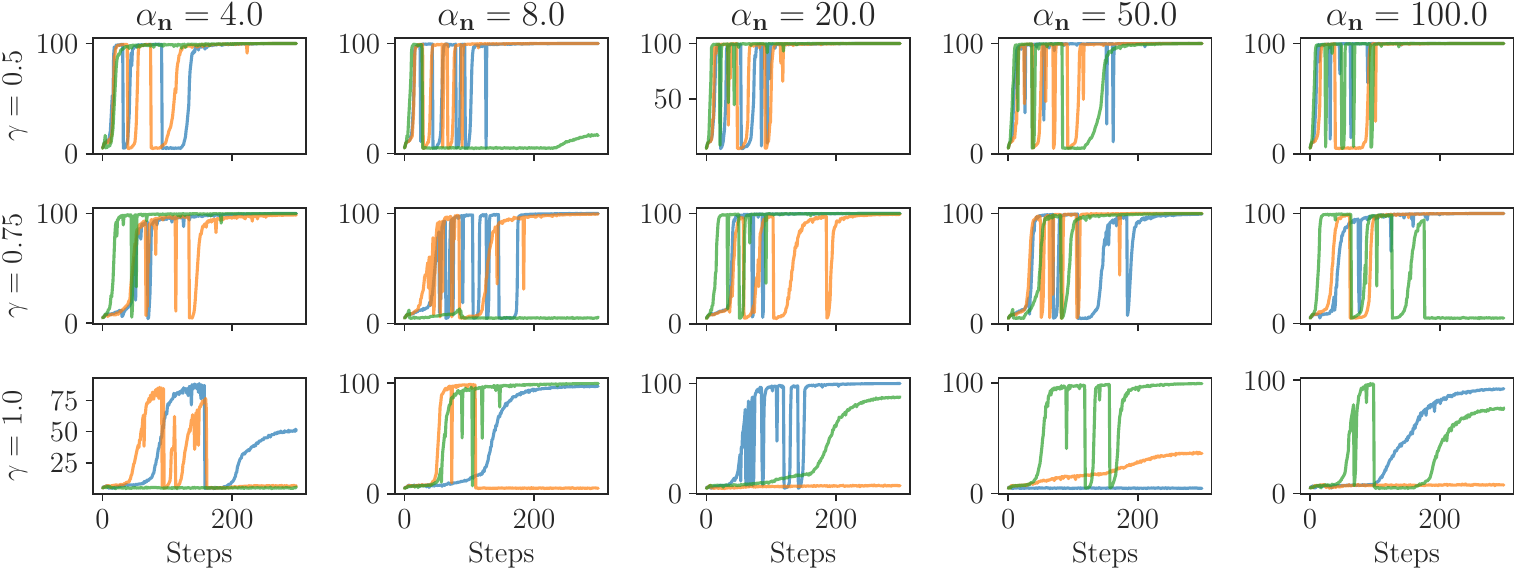}
    \caption{\textbf{Training loss plots of the ViT classifiers with $20$ classes from \cref{fig:synthetic_totals}, for $\alpha_\rvn\in[4,100]$, during training.}
    All reported ViT classifiers converge similarly to high accuracy when trained on the same task. Therefore, the trends of explainability biases  we see are not a product of some failure of the model, but rather an implicit bias of the problem setting.
    }
    \label{fig:vit20_acc_top2}
\end{figure}

In \cref{sec:synth_exp_results} we discuss the results of our controlled experiments, and demonstrate that the bias toward different explainability regimes can be controlled in our setting via adjusting the OBNR.
However, as the OBNR is also dominant in setting the complexity of the problem, it is important to validate that all compared runs converge and each perform well on its respective task. We therefore include in \cref{fig:vlm_acc_loss1,,fig:vlm_acc_loss2,,fig:vlm_acc_r51,,fig:vlm_acc_r52,,fig:vit10_acc_loss1,,fig:vit10_acc_loss2,,fig:vit10_acc_top1,,fig:vit10_acc_top2,,fig:vit20_acc_loss1,,fig:vit20_acc_loss2,,fig:vit20_acc_top1,,fig:vit20_acc_top2} plots depicting the performance of all trained models shown in \cref{fig:synthetic_totals}, separated according to their $\alpha_\rvx, \alpha_\rvy, \gamma, \alpha_\rvn$ configurations.

\paragraph{Contrastive VLMs.} For contrastive VLMs we plot the training loss and the text-to-image retrieval recall at $5$ (R@5) during training of all models in \cref{fig:vlm_acc_loss1,,fig:vlm_acc_loss2,,fig:vlm_acc_r51,,fig:vlm_acc_r52}.
R@5 is measured on a held-out set of $10{,}000$ pairs, where each text embedding ranks all $10{,}000$ image embeddings by cosine similarity, and it is the fraction of texts whose paired image ranks among the top $5$~\citep{radford2021learning}.
The image-to-text R@5 is nearly identical in all runs, so we report only the text-to-image direction.
It is important to note that each combination of $\alpha_\rvx,\alpha_\rvy$ and OBNR value constitutes a different data distribution of different difficulty.
For instance, when the OBNR is close to $1$ and $\alpha_\rvx=\alpha_\rvy=1$, then the object patch is much less separable from the background than in the case when the OBNR is close to $10$ and $\alpha_\rvx=\alpha_\rvy=4$.
This means that even an optimal model will achieve a higher training loss and lower R@5 for the more difficult case.
Therefore, we separate the different configurations in our plots.
As can be seen, all reported models converge and perform similarly well when trained on the same task.
Therefore, the trends of explainability biases we see in \cref{fig:synthetic_totals} are indeed not a product of some failure of the model, but an implicit bias of the problem setting.

We note that we did not include in \cref{fig:synthetic_totals,,fig:vlm_acc_loss1,,fig:vlm_acc_loss2,,fig:vlm_acc_r51,,fig:vlm_acc_r52} $11$ trained models which did not converge during training, and report only the remaining $232$ runs of the $243$ trained models for the $81$ configurations in the paper. 

\paragraph{ViT-based classifiers.} For the ViT classifiers, we plot the training losses and the top-1 classification accuracy during training of all models in 
\cref{fig:vit10_acc_loss1,,fig:vit10_acc_loss2,,fig:vit10_acc_top1,,fig:vit10_acc_top2} for $\gC=10$ and in
\cref{fig:vit20_acc_loss1,,fig:vit20_acc_loss2,,fig:vit20_acc_top1,,fig:vit20_acc_top2} for $\gC=20$.
Here, similarly to the contrastive case, each combination of $\gamma$ and OBNR value constitutes a different data distribution with different difficulty.
Therefore, we separate the different configurations in our plots.
As can be seen, all reported models converge, and perform similarly well when trained on the same task.
In fact, almost all classifiers converge to a top-1 accuracy close to $100\%$ when the explainability regimes are clearest, that is, when the OBNR is much larger or smaller than $1$. 
Thus, the trends of explainability biases  we see in \cref{fig:synthetic_totals} are indeed not a product of some failure of the classifier, but, here as well, an implicit bias of the problem setting.

\clearpage
\section{Details on the theoretical analysis of a toy model}
\label{app:theory}

This appendix covers the setting of the toy model of \cref{sec:theory} in full and provides the proofs for \cref{prop:theory_optimal} and \cref{thm:theory_regimes}.
We begin with describing the synthetic data distribution of the problem in \cref{sec:problem_setup}.
Then we derive in \cref{app:theory_optimal} the optimal classifier for this distribution in closed form, as summarized in \cref{prop:theory_optimal}.
Next, we show that the optimum has the structure of a transformer forward pass and identify weights for which a single-layer transformer implements the optimal solution precisely in \cref{app:theory_transformer}, which proves the first part of \cref{thm:theory_regimes}.
Finally, we provide proof for the specific $\alpha$ values for which the weights achieve the three regimes of \cref{thm:theory_regimes} in \cref{app:theory_signs,app:theory_flat}.

We note that the classification head of our toy ViT applies a logarithm, whereas standard, real-world ViTs use a linear head.
Therefore, we verify that the explainability regimes do not stem from this logarithm in \cref{app:theory_log}.
Specifically, we show that dropping it leaves a linear head that changes neither the predicted class nor the regime, and costs only the exactness of the predicted probabilities obtained by applying softmax to the logits
In addition, for completeness, we derive in \cref{app:theory_size} a closed form for the expected difference of the scores, $\E[\sobj-\sbg|\rc=i]$, in the aligned regime as $\alpha\to0$.

\subsection{Problem Setup}\label{sec:problem_setup}

\textbf{Notation.}
For a discrete random variable $\rc$, $p$ denotes its probability mass function, so $p_\rc(i)$ is the probability that $\rc=i$.
For a continuous random variable $\rvz$, $f$ denotes its probability density function, so $f_\rvz(\vz)$ is the density of $\rvz$ at $\vz$.

\textbf{Inputs.}
We consider a one-dimensional image $\rvz=(\rz_1,\ldots,\rz_N)\in\R^N$, a vector of length $N$ whose entry $\rz_n$ is the content of position $n$.
As depicted in the left part of \cref{fig:toy_model}, exactly one position $\rn_\textnormal{obj}$ holds an object, while the remaining positions in the image hold background. The object's distribution depends on which of the $K$ classes it belongs to:
\begin{itemize}
    \item Class: $p_\rc(i)=\pi_i$, for $i=1,\ldots,K$.
    \item Object position: $\rn_\textnormal{obj}\sim U(\{1,\ldots,N\})$, statistically independent of the class.
    \item Object content: $f_{\rz_{n_\textnormal{obj}}|\rc}(z|i)=\gN(z;\mu_i,\sigma_i^2)$.
    \item Background: $f_{\rz_n}(z)=\gN(z;0,\sigma^2)$ independently for every $n\neq n_\textnormal{obj}$.
\end{itemize}

The probability density of an image vector $\rvz$ conditioned on the case where an object of class $i$ is located in the $n$\textsuperscript{th} element is given by the Gaussian distribution:
\begin{equation}
    f_{\rvz|\rc,\rn_\textnormal{obj}}(\vz|i,n) \;=\; \gN\!\left(\vz;\vmu_{ni},\mSigma_{ni}\right),
    \quad
    \vmu_{ni}=\mu_i\ve^{(n)},
    \quad
    \mSigma_{ni}=\sigma^2\mI+(\sigma_i^2-\sigma^2)\ve^{(n)}\ve^{(n)\top},
    \label{eq:toy_joint}
\end{equation}
where $\ve^{(n)}$ is the standard basis vector with a one in coordinate $n$ and zeros elsewhere.
The covariance is a scaled identity matrix, perturbed in the single entry that carries the object.

Here, we assume the background content is drawn independently of the class. Thus, all information about the expected classification of the image resides in the $n_\textnormal{obj}$\textsuperscript{th} element.
Specifically, a token-decoding approach must highlight this exact token for an aligned-explainable solution.

\subsection{Proof of \texorpdfstring{\cref{prop:theory_optimal}}{Proposition 1} (The optimal classifier)}
\label{app:theory_optimal}

We now compute the posterior $p_{\rc|\rvz}(i|\vz)$ in closed form.
First, we apply Bayes' rule and the law of total probability:
\begin{align}
    p_{\rc|\rvz}(i|\vz) &= \frac{f_{\rvz|\rc}(\vz| i)\,p_\rc(i)}{f_\rvz(\vz)} \nonumber\\
    &= \frac{\pi_i\,f_{\rvz|\rc}(\vz| i)}{\sum_{j=1}^{K}\pi_j\,f_{\rvz|\rc}(\vz| j)}.
    \label{eq:toy_bayes}
\end{align}
Then, we use the law of total probability over the unknown object position for $f_{\rvz|\rc}(\vz|i)$:
\begin{align}
    f_{\rvz|\rc}(\vz|i) &= \sum_{n=1}^{N} f_{\rvz|\rc,\rn_\textnormal{obj}}(\vz|i,n)\,p_{\rn_\textnormal{obj}}(n) \nonumber\\
    &= \frac{1}{N}\sum_{n=1}^{N}\gN(\vz;\vmu_{ni},\mSigma_{ni}).
    \label{eq:toy_positionmix}
\end{align}
The second step in the last equality uses $p_{\rn_\textnormal{obj}}(n)=1/N$, since the object position is uniform.
When substituting \cref{eq:toy_positionmix} into \cref{eq:toy_bayes}, the factor $1/N$ cancels and we obtain:
\begin{equation}
    p_{\rc|\rvz}(i|\vz) \;=\; \frac{\pi_i\sum_{n=1}^{N}\gN\!\left(\vz;\vmu_{ni},\mSigma_{ni}\right)}{\sum_{j=1}^{K}\pi_j\sum_{n=1}^{N}\gN\!\left(\vz;\vmu_{nj},\mSigma_{nj}\right)} .
    \label{eq:toy_marginal}
\end{equation}

The expressions in the numerator and the denominator are given by the multivariate Gaussian distribution,
\begin{equation}
    \gN\!\left(\vz;\vmu_{ni},\mSigma_{ni}\right)
    \;=\; \frac{1}{\sqrt{(2\pi)^N\lvert\mSigma_{ni}\rvert}}
      \exp\!\left(-\tfrac{1}{2}(\vz-\vmu_{ni})^{\top}\mSigma_{ni}^{-1}(\vz-\vmu_{ni})\right).\label{eq:toy_gaussian}
\end{equation}
The covariance $\mSigma_{ni}$ is diagonal, with $\sigma^2$ in every diagonal entry except the $n$\textsuperscript{th}, which is $\sigma_i^2$.
Its inverse is therefore also diagonal, with the reciprocal of each entry,
\begin{align}
    \mSigma_{ni}^{-1} &= \frac{1}{\sigma^2}\mI+ \frac{1}{\sigma_i^2}\ve^{(n)}\ve^{(n)\top} - \frac{1}{\sigma^2}\ve^{(n)}\ve^{(n)\top} \nonumber \\
     &= \frac{1}{\sigma^2}\mI+\left(\frac{1}{\sigma_i^2}-\frac{1}{\sigma^2}\right)\ve^{(n)}\ve^{(n)\top}.\label{eq:toy_inverted_sigma}
\end{align}
Substituting this and $\vmu_{ni}=\mu_i\ve^{(n)}$ in the quadratic form in the exponent, it expands as
\begin{align}
    (\vz-\vmu_{ni})^{\top}\mSigma_{ni}^{-1}(\vz-\vmu_{ni})
    &\;=\; \frac{1}{\sigma^2}(\vz-\mu_i\ve^{(n)})^{\top}(\vz-\mu_i\ve^{(n)}) \nonumber \\
    &\qquad\;+\; \left(\frac{1}{\sigma_i^2}-\frac{1}{\sigma^2}\right)(\vz-\mu_i\ve^{(n)})^{\top}\ve^{(n)}\ve^{(n)\top}(\vz-\mu_i\ve^{(n)}) \nonumber\\
    &\;=\; \frac{1}{\sigma^2}\lVert\vz-\mu_i\ve^{(n)}\rVert^2
      + \left(\frac{1}{\sigma_i^2}-\frac{1}{\sigma^2}\right)(z_n-\mu_i)^2 \nonumber\\
    &\;=\; \frac{1}{\sigma^2}\left(\lVert\vz\rVert^2-2\mu_i z_n+\mu_i^2\right)
      + \left(\frac{1}{\sigma_i^2}-\frac{1}{\sigma^2}\right)(z_n-\mu_i)^2 \nonumber\\
    &\;=\; \frac{1}{\sigma^2}\left(\lVert\vz\rVert^2-z_n^2+(z_n-\mu_i)^2\right)
      + \left(\frac{1}{\sigma_i^2}-\frac{1}{\sigma^2}\right)(z_n-\mu_i)^2 \nonumber\\
    &\;=\; \frac{1}{\sigma^2}\left(\lVert\vz\rVert^2-z_n^2\right) + \frac{1}{\sigma_i^2}(z_n-\mu_i)^2 ,
    \label{eq:toy_quadform}
\end{align}
where the second line uses $\ve^{(n)\top}(\vz-\mu_i\ve^{(n)})=z_n-\mu_i$, the fourth uses $-2\mu_i z_n+\mu_i^2=-z_n^2+(z_n-\mu_i)^2$, and the fifth collects the two copies of $(z_n-\mu_i)^2$.

Finally, since $\mSigma_{ni}$ is diagonal, its determinant is the product of its diagonal entries, $\sigma^{2(N-1)}\sigma_i^2$, and only the factor $\sigma_i$ depends on the class.

Substituting the determinant and \cref{eq:toy_quadform} into \cref{eq:toy_gaussian} and rearranging gives
\begin{align}
    \gN\!\left(\vz;\vmu_{ni},\mSigma_{ni}\right)
    &= \frac{1}{\sqrt{(2\pi)^N\,\sigma^{2(N-1)}\sigma_i^2}}\,
      \exp\!\left(\frac{z_n^2-\lVert\vz\rVert^2}{2\sigma^2}-\frac{(z_n-\mu_i)^2}{2\sigma_i^2}\right) \nonumber\\
    &= \underbrace{\frac{e^{-\lVert\vz\rVert^2/2\sigma^2}}{(2\pi)^{N/2}\,\sigma^{N-1}}}_{\textnormal{same for every } i \textnormal{ and } n}
      \;\cdot\; \frac{1}{\sigma_i}\,\exp\!\left(\frac{z_n^2}{2\sigma^2}-\frac{(z_n-\mu_i)^2}{2\sigma_i^2}\right) .
    \label{eq:toy_density}
\end{align}
The first multiplier does not depend on the class $i$ or the object position $n$, which means when substituting \cref{eq:toy_density} back into \cref{eq:toy_marginal} we obtain
\begin{equation}
    p_{\rc|\rvz}(i|\vz) = \frac{\pi_i\sum_{n=1}^{N}
    \frac{1}{\sigma_i}\,\exp\!\left(\frac{z_n^2}{2\sigma^2}-\frac{(z_n-\mu_i)^2}{2\sigma_i^2}\right)
    }{\sum_{j=1}^{K}\pi_j\sum_{n=1}^{N}
    \frac{1}{\sigma_j}\,\exp\!\left(\frac{z_n^2}{2\sigma^2}-\frac{(z_n-\mu_j)^2}{2\sigma_j^2}\right)
    }.
    \label{eq:toy_marginal_expanded}
\end{equation}
The exponent term depends on the image only through $z_n$, and is the only factor in the expression that couples the class and the object's position.
We therefore call it the \emph{class feature} $\psi_{i,n}$, and combine the leftover $1/\sigma_i$ with the prior $\pi_i$ into a single \emph{class bias},
\begin{equation}
    \psi_i(z) \;=\; \frac{z^2}{2\sigma^2}-\frac{(z-\mu_i)^2}{2\sigma_i^2},
    \qquad
    \psi_{i,n}\;\equiv\;\psi_i(z_n),
    \qquad
    \beta_i \;=\; \log \pi_i-\log\sigma_i .
    \label{eq:toy_feature}
\end{equation}
Substituting \cref{eq:toy_density} into \cref{eq:toy_marginal_expanded}, 
\begin{align}
    p_{\rc|\rvz}(i|\vz)
    &= \frac{\frac{\pi_i}{\sigma_i}\sum_{n=1}^{N}e^{\psi_{i,n}}}{\sum_{j=1}^{K}\frac{\pi_j}{\sigma_j}\sum_{n=1}^{N}e^{\psi_{j,n}}} \nonumber\\
    &= \frac{\exp\!\left(\log\left(\sum_{n=1}^{N}e^{\psi_{i,n}}\right)+\beta_i\right)}{\sum_{j=1}^{K}\exp\!\left(\log\left(\sum_{n=1}^{N}e^{\psi_{j,n}}\right)+\beta_j\right)} \nonumber\\
    &= \mathrm{softmax}_i\!\left(\log\left(\sum_{n=1}^{N}e^{\psi_{i,n}}\right)+\beta_i\right).
    \label{eq:toy_lse}
\end{align}

This concludes the proof of \cref{prop:theory_optimal}.
This posterior is the optimal classifier: no other classifier achieves a lower cross-entropy over the data distribution, and it is what a model trained on unlimited data converges to.
Any weights that make a model compute \cref{eq:toy_lse} are therefore optimal, and the loss gives no reason to prefer one setting over another.

Importantly, \cref{eq:toy_lse} has the structure of some forward pass of classifier. First, features are extracted from patches of an image. Then, an aggregation over the features passes through a non-linear layer. Finally, together with an added bias, a softmax over the classes produces a classification.

\subsection{A single-layer transformer that computes the optimum (first part of \texorpdfstring{\cref{thm:theory_regimes}}{Theorem 1})}
\label{app:theory_transformer}

We now show it is possible to choose weights for the toy model of \cref{fig:toy_model} so that its class-token output equals the optimal classifier of \cref{eq:toy_lse} exactly, for every input $\vz$.

\paragraph{Token embedding.}
We consider the image patches to be already tokenized.
Therefore, we only apply token embedding.
Our embedding acts on each patch separately, without mixing, and raises the dimension of a scalar patch to a vector of dimension $K+1$:
\begin{equation}
    \phi:\R\to\R^{K+1},
    \qquad
    \phi(z) \;=\; \bigl(\,\underbrace{1}_{\textnormal{anchor}},\;\underbrace{e^{\psi_1(z)},\ldots,e^{\psi_K(z)}}_{K\textnormal{ class channels}}\,\bigr).
    \label{eq:toy_embed}
\end{equation}
For the learnable class token, we choose the learned weights such that the already-embedded class token is $\ve^{(0)}$.
Stacking it on top of the $N$ embedded patches gives the token matrix
\begin{equation}
    \mX \;=\;
    \begin{pmatrix} \vx_\textnormal{cls}^{\top} \\[3pt] \vx_1^{\top} \\[2pt] \vdots \\[2pt] \vx_N^{\top} \end{pmatrix}
    \;=\;
    \begin{pmatrix}
        1 & 0 & \cdots & 0 \\
        1 & e^{\psi_{1,1}} & \cdots & e^{\psi_{K,1}} \\
        \vdots & \vdots & & \vdots \\
        1 & e^{\psi_{1,N}} & \cdots & e^{\psi_{K,N}}
    \end{pmatrix}
    \in\R^{(N+1)\times(K+1)},
    \label{eq:toy_tokenmatrix}
\end{equation}
whose rows are the tokens $\vx_m$, with $\vx_\textnormal{cls}=\ve^{(0)}$ and $\vx_n=\phi(z_n)$ for $n=1,\ldots,N$.

\paragraph{Self-attention.}
For the attention matrices, we set as the weights of the self-attention the rank-one choice $\mW_Q\mW_K^{\top}=\ve^{(0)}\ve^{(0)\top}$ and $\mW_V=(N+1)\alpha\mI_{K+1}$ for a nonzero scalar $\alpha$.
With this choice, the attention score between tokens $m$ and $m'$ before the softmax, $R_{m,m'}$, is the product of their anchors, which are all equal $1$:
\begin{align}
    R_{m,m'} &\;=\; \vx_m^{\top}\mW_Q\mW_K^{\top}\vx_{m'} \;=\; \bigl(\vx_m^{\top}\ve^{(0)}\bigr)\bigl(\ve^{(0)\top}\vx_{m'}\bigr) \;=\; 1.
\end{align}
Then, the softmax over the $N+1$ tokens equalizes all attention scores:
\begin{align}
    A_{m,m'} &\;=\; \frac{e^{R_{m,m'}}}{\sum_{m''}e^{R_{m,m''}}} \;=\; \frac{1}{N+1}.
    \label{eq:toy_score_entry}
\end{align}
Notably, the attention scores remain equal under any attention scaling inside the softmax, such as the common choice of $1/\sqrt{d}$. 

Therefore, after multiplying the softmax result by $\mW_V$ and adding the residual connection, we get
\begin{equation}
    \tilde{\mX} = \mX + \mA\mX\mW_V.
\end{equation}
Specifically, for an input token $\vx_m$ where $m=\textnormal{cls},1,\ldots,N$, its output token $\tilde{\vx}_m$ becomes
\begin{align}
    \tilde{\vx}_m &= \vx_m + (N+1)\alpha\sum_{m'}A_{m,m'}\vx_{m'} \nonumber \\
    &= \underbrace{\vx_m}_{\textnormal{residual}} + \underbrace{\alpha \sum_{m'}\vx_{m'}}_{\textnormal{attention output} \equiv \alpha\vv}.
    \label{eq:toy_residual}
\end{align}
Importantly, the attention output equals $\alpha \sum_{m'}\vx_{m'}$ for all tokens, a constant which we denote $\alpha\vv$.
Thus, the residual element keeps the content specific to each token. 

For the class token, the residual stream $\ve^{(0)}$ is zero for all elements except for the first, such that
\begin{equation}
    \tilde{\vx}_\textnormal{cls} \;=\; \ve^{(0)}+\alpha\vv ,
    \qquad
    \bigl(\tilde{\vx}_\textnormal{cls}\bigr)_0 \;=\; 1+\alpha(N+1) ,
    \qquad
    \underbrace{\bigl(\tilde{\vx}_\textnormal{cls}\bigr)_i \;=\; \alpha\sum_{n=1}^{N} e^{\psi_{i,n}} \;\equiv\; \alpha v_i}_{i\,=\,1,\ldots,K} .
    \label{eq:toy_pool}
\end{equation}

\paragraph{Head.}
For our learned head, we choose weights that ignore the anchor channel of the class-token and act on the remaining $K$ coordinates as
\begin{equation}
    g:\R^{K+1}\to\R^{K},
    \qquad
    g(\vu)_i \;=\; \log\left(\frac{u_i}{\alpha}\right)+\beta_i,
    \qquad i=1,\ldots,K ,
    \label{eq:toy_head}
\end{equation}
followed by a softmax over classes.
The factor $1/\alpha$ undoes the factor $\alpha$ in the attention output, so by \cref{eq:toy_pool} the class-token logit is $\log(v_i)+\beta_i$ for every $\alpha$.
This head implementation includes the outer logarithm and bias of \cref{eq:toy_lse}. The argument of the logarithm, which includes summation over exponents, is exactly the result of the self-attention with our chosen weights.
Therefore, our chosen embedding, self-attention weights, and the head implementation compute together the optimal classifier exactly,
\begin{equation}\mathrm{softmax}_i\,g\bigl(\tilde{\vx}_\textnormal{cls}\bigr) \;=\; p_{\rc|\rvz}(i|\vz) \qquad \textnormal{for every } \vz\in\R^N .
    \label{eq:toy_exact}
\end{equation}

Importantly, the class-token logits $\log(v_i)+\beta_i$ do not depend on $\alpha$, so they will compute the same classification for every chosen nonzero $\alpha$.
This proves the first part of \cref{thm:theory_regimes}.

\subsection{Proofs for the aligned and inverted regimes (items 1 and 2 of \texorpdfstring{\cref{thm:theory_regimes}}{Theorem 1})}
\label{app:theory_signs}

A common token-decoding explainability approach, which we use throughout the paper, applies the classification head, which is usually reserved for the class token, to the patch tokens as well.
We denote the logit output for class $i$ of the head when applied to a patch token as the score $\gS(\cdot)_i$ from \cref{sec:defs}.
Here, $i$ is the true label of the object.

From \cref{eq:toy_residual}, for our choice of weights, the class-$i$ channel of the patch token $\tilde{\vx}_n$, after the attention and the residual connection and before the classification head, is
\begin{equation}
    (\tilde{\vx}_n)_i = e^{\psi_{i,n}}+\alpha v_i.
\end{equation}
Therefore, its score after passing through the classification head from \cref{eq:toy_head} is 
\begin{equation}
    \gS(\tilde{\vx}_n)_i = g(\tilde{\vx}_n)_i =\log\!\left(\frac{e^{\psi_{i,n}}}{\alpha}+v_i\right)+\beta_i .
    \label{eq:toy_readout}
\end{equation}
We note that $\tilde{\vx}_n$ is dependent on all the patch-tokens of the original input $\vz$, as it is the output of the self-attention layer.

We now prove items 1 and 2 of \cref{thm:theory_regimes}:
for every $\alpha>0$ and every $\alpha\le-1$, the expectation of the difference of the scores, $\E_\rvz[\sobj-\sbg|\rc=i]$, has the sign of $\alpha$, unless class $i$ has the same distribution as the background.
We show this in five steps, which are outlined below.
First, we verify that the patch logit is defined and inspect how it changes with $\alpha$. 
Next, we reduce the expectation of the difference from being computed over all patches in the image, to a single pair of patches. 
Then, we prove two properties of that difference of a pair of patches. 
Following that, we rewrite the expectation of the pair in a symmetric form.
Finally, we analyze the symmetric form to understand the sign of the expression.

\paragraph{Step 1: the patch logit.}
Take any $\alpha>0$ or any $\alpha\le-1$, with the head of \cref{eq:toy_head}.
For $\alpha>0$ the argument of the logarithm in \cref{eq:toy_readout} is a sum of positive terms.
For $\alpha\le-1$ it stays positive as well, since $1/\alpha\ge-1$ gives
\begin{equation}
    v_i + \frac{e^{\psi_{i,n}}}{\alpha} \ge v_i-e^{\psi_{i,n}}=\sum_{m\neq n}e^{\psi_{i,m}}>0.
\end{equation}
Therefore, the patch logit is defined in both cases. 
Additionally, in both cases the class token still returns the same logit $\log(v_i)+\beta_i$, so by \cref{eq:toy_exact} the model remains optimal.
Crucially, for a specific image $\vz$, $v_i$ is the same for all patches, so the patch logit increases with $\psi_{i,n}$ when $\alpha>0$ and decreases when $\alpha<0$.

\paragraph{Step 2: reduce the expectation of the difference of scores from all patches in the image to one pair of patches.}
To consider the explainability of the image $\rvz$, we inspect the difference between $\sobj$ and $\sbg$ as defined in \cref{eq:eqxplainability_scores}, for the object of class $i$ that appears in $\vz$.
Here, since the object occupies a single token, $\sobj$ is the logit at the position of the object patch $n_\textnormal{obj}$, whereas $\sbg$ is the mean logit over the $N-1$ background patches:
\begin{align}
    \sobj-\sbg &= \gS(\tilde{\vx}_{n_\textnormal{obj}})_i - \frac{1}{N-1}\sum_{b\neq n_\textnormal{obj}}\gS(\tilde{\vx}_b)_i \nonumber \\
    &= \frac{1}{N-1}\sum_{b\neq n_\textnormal{obj}}\Bigl(\gS(\tilde{\vx}_{n_\textnormal{obj}})_i-\gS(\tilde{\vx}_b)_i\Bigr) .
    \label{eq:toy_bg_mean}
\end{align}

To derive the expectation of these differences, $\E_{\rvz,\rn_\textnormal{obj}}\left[\sobj-\sbg|\rc=i\right]$, we first consider the case when 
we condition on $\rn_\textnormal{obj}=n_\textnormal{obj}$.
Now, when the object position is fixed, the background values are independent and identically distributed, and every logit depends on the other patches only through the sum $v_i=\sum_{n=1}^{N} e^{\psi_{i,n}}$ (\cref{eq:toy_pool}), which does not depend on their order.
Exchanging two background patches therefore leaves the joint distribution of the logits unchanged, so $\E_\rvz\left[\gS(\tilde{\vx}_{n_\textnormal{obj}})_i-\gS(\tilde{\vx}_b)_i|\rc=i,\rn_\textnormal{obj}=n_\textnormal{obj}\right]$ is the same for every background patch $b$.
By linearity, and since all the terms in the sum are equal, for any single background patch $b$,
\begin{align}
    \E_\rvz\left[\sobj-\sbg|\rc=i,\rn_\textnormal{obj}=n_\textnormal{obj}\right] &= 
    \E_\rvz\left[ \frac{1}{N-1}\sum_{b\neq n_\textnormal{obj}}\Bigl(\gS(\tilde{\vx}_{n_\textnormal{obj}})_i-\gS(\tilde{\vx}_b)_i\Bigr) \,\Big|\, \rc=i,\rn_\textnormal{obj}=n_\textnormal{obj} \right] \nonumber \\
    &= \frac{1}{N-1}\sum_{b\neq n_\textnormal{obj}}\E_\rvz\left[ \gS(\tilde{\vx}_{n_\textnormal{obj}})_i-\gS(\tilde{\vx}_b)_i \,\big|\, \rc=i,\rn_\textnormal{obj}=n_\textnormal{obj} \right] \nonumber \\
    &= \E_\rvz\left[ \gS(\tilde{\vx}_{n_\textnormal{obj}})_i-\gS(\tilde{\vx}_b)_i \,\big|\, \rc=i,\rn_\textnormal{obj}=n_\textnormal{obj} \right] .
    \label{eq:toy_pair}
\end{align}
This expectation is the same for every object position $n_\textnormal{obj}$, since both the data distribution and the transformer, which has no positional embedding, treat all positions alike.
Since $\rn_\textnormal{obj}$ is uniform and independent of the class, the law of total expectation over it gives, for any fixed $n_\textnormal{obj}$ and any background patch $b\neq n_\textnormal{obj}$,
\begin{align}
    \E_{\rvz,\rn_\textnormal{obj}}\left[\sobj-\sbg|\rc=i\right]
    &= \frac{1}{N}\sum_{n=1}^{N}\E_\rvz\left[\sobj-\sbg|\rc=i,\rn_\textnormal{obj}=n\right] \nonumber\\
    &= \E_\rvz\left[ \gS(\tilde{\vx}_{n_\textnormal{obj}})_i-\gS(\tilde{\vx}_b)_i \,\big|\, \rc=i,\rn_\textnormal{obj}=n_\textnormal{obj} \right] .
    \label{eq:toy_pair_avg}
\end{align}
In particular, this means we can inspect the expectation of the difference of the object patch and a single background patch, instead of considering all background patches.

\paragraph{Step 3: two properties of the difference of a single pair, $D_i(x,y;r)$.}
First, $\gS(\tilde{\vx}_{n_\textnormal{obj}})_i-\gS(\tilde{\vx}_b)_i$ from \cref{eq:toy_pair} satisfies
\begin{align}
\gS(\tilde{\vx}_{n_\textnormal{obj}})_i-\gS(\tilde{\vx}_b)_i &= \log\!\left(\frac{e^{\psi_i(z_{n_\textnormal{obj}})}}{\alpha}+v_i\right)-\log\!\left(\frac{e^{\psi_i(z_b)}}{\alpha}+v_i\right) \nonumber \\
    &= \log\!\left(\frac{e^{\psi_i(z_{n_\textnormal{obj}})}}{\alpha}+e^{\psi_i(z_{n_\textnormal{obj}})}+e^{\psi_i(z_b)}+r\right) - \nonumber \\
    &\qquad\qquad\qquad\qquad\qquad \log\!\left(\frac{e^{\psi_i(z_b)}}{\alpha}+e^{\psi_i(z_{n_\textnormal{obj}})}+e^{\psi_i(z_b)}+r\right), 
\end{align}
as $\beta_i$ cancels. Here, $r=\sum_{m\notin\{n_\textnormal{obj},b\}}e^{\psi_{i,m}}$, collects all $N-2$ background patches apart from $z_b$.

Now, we construct a function $D_i(x,y;r)$ of real numbers $x$, $y$ and $r\ge0$, which we evaluate at the object entry $x=z_{n_\textnormal{obj}}$, the entry $y=z_b$ of a background patch $b$ in the same image:
\begin{equation}
    D_i(x,y;r) \;=\; \log\!\left(\frac{e^{\psi_i(x)}}{\alpha}+e^{\psi_i(x)}+e^{\psi_i(y)}+r\right)-\log\!\left(\frac{e^{\psi_i(y)}}{\alpha}+e^{\psi_i(x)}+e^{\psi_i(y)}+r\right).
    \label{eq:toy_D}
\end{equation}
Both arguments of the logarithms are positive, as in Step 1, since the first equals $(1+1/\alpha)e^{\psi_i(x)}+e^{\psi_i(y)}+r$ with $1+1/\alpha\ge0$, and similarly for the second.
Importantly, the score difference is
\begin{equation}
    \gS(\tilde{\vx}_{n_\textnormal{obj}})_i-\gS(\tilde{\vx}_b)_i \;=\; D_i(z_{n_\textnormal{obj}},z_b;r) ,
    \label{eq:toy_pair_D}
\end{equation}
with $r$ computed from the same image.
The function $D_i$ operates on the patch entries directly, whereas the scores operate on the same patches after the embedding and the self-attention.

Given $\rc=i$ and $\rn_\textnormal{obj}=n_\textnormal{obj}$, the entries of the image are independent (\cref{eq:toy_joint}), so $\rz_{n_\textnormal{obj}}$, $\rz_b$ and $\rr=\sum_{m\notin\{n_\textnormal{obj},b\}}e^{\psi_i(\rz_m)}$ are independent, since $\rr$ is computed from the other $N-2$ background entries.
Substituting \cref{eq:toy_pair_D} into \cref{eq:toy_pair_avg} and applying the law of total expectation over $\rr$ gives
\begin{align}\label{eq:s-is-d}
    \E_{\rvz,\rn_\textnormal{obj}}\left[\sobj-\sbg|\rc=i\right] &= \E_{\rz_{n_\textnormal{obj}},\rz_b,\rr}\left[D_i(\rz_{n_\textnormal{obj}},\rz_b;\rr)|\rc=i,\rn_\textnormal{obj}=n_\textnormal{obj}\right] \nonumber \\
    &= \E_{\rr}\Big[\E_{\rz_{n_\textnormal{obj}},\rz_b}\left[D_i(\rz_{n_\textnormal{obj}},\rz_b;\rr) | \rc=i,\rn_\textnormal{obj}=n_\textnormal{obj},\rr\right]\Big].
\end{align}

\textbf{(a) $D_i$ is an antisymmetric function.}
Exchanging $x$ and $y$ leaves the sum $e^{\psi_i(x)}+e^{\psi_i(y)}+r$ unchanged and exchanges the two logarithms in \cref{eq:toy_D}, so
\begin{equation}
    D_i(y,x;r) \;=\; -D_i(x,y;r) \quad \textnormal{for every } r .
    \label{eq:toy_antisym}
\end{equation}

\textbf{(b) The sign of $D_i$ follows $\alpha$ and the likelihood ratio.}
Given $\rc=i$ and $\rn_\textnormal{obj}=n_\textnormal{obj}$, the object entry has density $f_\textnormal{obj}(z)=\gN(z;\mu_i,\sigma_i^2)$ and each background entry has density $f_\textnormal{bg}(z)=\gN(z;0,\sigma^2)$ (\cref{sec:problem_setup}).
Let $L(z)=f_\textnormal{obj}(z)/f_\textnormal{bg}(z)$ be the likelihood ratio of the object to the background.
Dividing the two Gaussian densities,
\begin{equation}
    L(z)
    \;=\; \frac{\frac{1}{\sqrt{2\pi}\,\sigma_i}\exp\!\left(-\frac{(z-\mu_i)^2}{2\sigma_i^2}\right)}{\frac{1}{\sqrt{2\pi}\,\sigma}\exp\!\left(-\frac{z^2}{2\sigma^2}\right)}
    \;=\; \frac{\sigma}{\sigma_i}\exp\!\left(\frac{z^2}{2\sigma^2}-\frac{(z-\mu_i)^2}{2\sigma_i^2}\right)
    \;=\; \frac{\sigma}{\sigma_i}\,e^{\psi_i(z)} ,
    \label{eq:toy_ratio}
\end{equation}
so $L(z)$ equals $e^{\psi_i(z)}$ up to a positive factor, and is positive itself.
Next, considering $D_i$, both logarithms in \cref{eq:toy_D} contain the same sum $v_i=e^{\psi_i(x)}+e^{\psi_i(y)}+r$, so
\begin{align}
    D_i(x,y;r)>0
    &\iff \log\!\left(\frac{e^{\psi_i(x)}}{\alpha}+v_i\right) > \log\!\left(\frac{e^{\psi_i(y)}}{\alpha}+v_i\right)
    && \textnormal{(by the definition of $D_i$)} \notag\\
    &\iff \frac{e^{\psi_i(x)}}{\alpha}+v_i \;>\; \frac{e^{\psi_i(y)}}{\alpha}+v_i
    && \textnormal{(the logarithm is increasing)} \notag\\
    &\iff \frac{e^{\psi_i(x)}}{\alpha} \;>\; \frac{e^{\psi_i(y)}}{\alpha}
    && \textnormal{(subtract $v_i$ from both sides)} \notag\\
    &\iff \alpha\bigl(e^{\psi_i(x)}-e^{\psi_i(y)}\bigr) \;>\; 0
    && \textnormal{(multiply both sides by $\alpha^2>0$)} \notag \\
    &\iff \alpha\bigl(L(x)-L(y)\bigr) \;>\; 0
    && \textnormal{(by \cref{eq:toy_ratio})} .\label{eq:toy_equivalences}
\end{align}
Every step is an equivalence, and the same chain with $<$ or $=$ in place of $>$ covers the other two cases.
Together,
\begin{equation}
    \textnormal{sgn}\{D_i(x,y;r)\} = \textnormal{sgn} \left\{ \alpha\bigl(L(x)-L(y)\bigr)\right\}.
    \label{eq:toy_sign}
\end{equation}
Since $v_i$ cancels in this chain, \cref{eq:toy_sign} holds for every value of $r$.

\paragraph{Step 4: a symmetric form of the expectation.}
We now compute the inner expectation of the RHS of \cref{eq:s-is-d} for a fixed value of $\rr$.
Given $\rc=i$ and $\rn_\textnormal{obj}=n_\textnormal{obj}$, the random variables $\rz_{n_\textnormal{obj}}$, $\rz_b$ and $\rr$ are mutually independent (Step 3), so fixing $\rr=r$ does not change the joint density $f_\textnormal{obj}(x)f_\textnormal{bg}(y)$ of $\rz_{n_\textnormal{obj}}$ and $\rz_b$.
Therefore,
\begin{align}
    &\E\left[D_i(\rz_{n_\textnormal{obj}},\rz_b;\rr)\,|\,\rc=i,\rn_\textnormal{obj}=n_\textnormal{obj},\rr=r\right] \nonumber\\
    &\qquad= \iint f_\textnormal{obj}(x)\,f_\textnormal{bg}(y)\,D_i(x,y;r)\,dx\,dy
    && \textnormal{(definition of the expectation)} \nonumber\\
    &\qquad= \iint f_\textnormal{bg}(x)\,f_\textnormal{bg}(y)\,L(x)\,D_i(x,y;r)\,dx\,dy
    && \textnormal{(since } f_\textnormal{obj}=L\,f_\textnormal{bg}\textnormal{)} \nonumber\\
    &\qquad= \iint f_\textnormal{bg}(y)\,f_\textnormal{bg}(x)\,L(y)\,D_i(y,x;r)\,dy\,dx
    && \textnormal{(renaming } x\leftrightarrow y\textnormal{)} \nonumber\\
    &\qquad= -\iint f_\textnormal{bg}(x)\,f_\textnormal{bg}(y)\,L(y)\,D_i(x,y;r)\,dx\,dy
    && \textnormal{(by antisymmetry)} . 
\end{align}
The fourth line only changes the names of the two integration variables, which does not change the value of the integral.
The third and fifth lines are two expressions for the same quantity, so adding them and dividing by two gives
\begin{align}
    &\E\left[D_i(\rz_{n_\textnormal{obj}},\rz_b;\rr)\,|\,\rc=i,\rn_\textnormal{obj}=n_\textnormal{obj},\rr=r\right] \nonumber\\
    &\qquad= \frac{1}{2}\iint f_\textnormal{bg}(x)\,f_\textnormal{bg}(y)\,\bigl(L(x)-L(y)\bigr)\,D_i(x,y;r)\,dx\,dy .
    \label{eq:toy_swap}
\end{align}

\paragraph{Step 5: the sign.}
By \cref{eq:toy_sign}, for every $x$, $y$ and $r$, the integrand of \cref{eq:toy_swap} satisfies
\begin{equation}
    \textnormal{sgn}\bigl\{f_\textnormal{bg}(x)\,f_\textnormal{bg}(y)\bigl(L(x)-L(y)\bigr)D_i(x,y;r)\bigr\} = \textnormal{sgn}\bigl\{\alpha\bigl(L(x)-L(y)\bigr)^2\bigr\},
    \label{eq:toy_integrand_sign}
\end{equation}
since $f_\textnormal{bg}(x)>0$ for Gaussian distributions. 
The integrand is therefore non-negative when $\alpha>0$ and non-positive when $\alpha\le-1$, so by \cref{eq:toy_swap}, for every $r$,
\begin{align}
    \E\left[D_i(\rz_{n_\textnormal{obj}},\rz_b;\rr)\,|\,\rc=i,\rn_\textnormal{obj}=n_\textnormal{obj},\rr=r\right] &\;\ge\; 0 \quad \textnormal{if } \alpha>0, \nonumber\\
    \E\left[D_i(\rz_{n_\textnormal{obj}},\rz_b;\rr)\,|\,\rc=i,\rn_\textnormal{obj}=n_\textnormal{obj},\rr=r\right] &\;\le\; 0 \quad \textnormal{if } \alpha\le-1 .
    \label{eq:toy_inner_weak}
\end{align}
Specifically, the integrand will be zero only when $L(x)=L(y)$ for all $x$ and $y$ values.
Thus, 
for the inequalities to be strict 
the integrand of \cref{eq:toy_swap} must be nonzero over some $x$ and $y$ values.

Lastly, we show that these inequalities are strict.
Recall that $L(z)$ is the likelihood ratio of the two distributions  $f_\textnormal{obj}(z)$ and $ f_\textnormal{bg}(z)$. 
Since the distribution of class $i$ differs from that of the background, $f_\textnormal{obj}(z)\neq f_\textnormal{bg}(z)$, $L(z)$ is not a constant function. 
As $L(z)$ is also continuous, there exists some area in the $(x,y)$ plane in which $L(x)\neq L(y)$ for every pair $(x,y)$.
Thus, in this area $\smash{\alpha\bigl(L(x)-L(y)\bigr)^2\neq0}$.
So, by \cref{eq:toy_integrand_sign} the integrand of \cref{eq:toy_swap} is nonzero, and so is the integral. 
The inequalities in \cref{eq:toy_inner_weak} are therefore strict, that is,
\begin{align}
    \E\left[D_i(\rz_{n_\textnormal{obj}},\rz_b;\rr)\,|\,\rc=i,\rn_\textnormal{obj}=n_\textnormal{obj},\rr=r\right] &\;>\; 0 \quad \textnormal{if } \alpha>0, \nonumber\\
    \E\left[D_i(\rz_{n_\textnormal{obj}},\rz_b;\rr)\,|\,\rc=i,\rn_\textnormal{obj}=n_\textnormal{obj},\rr=r\right] &\;<\; 0 \quad \textnormal{if } \alpha\le-1 .
    \label{eq:toy_inner_sign}
\end{align}

Finally, the outer expectation in \cref{eq:s-is-d} averages this inner expectation over $\rr$, so it has the same sign, which gives
\begin{equation}
    \E\left[\sobj-\sbg|\rc=i\right] \;>\; 0 \quad \textnormal{if } \alpha>0,
    \qquad
    \E\left[\sobj-\sbg|\rc=i\right] \;<\; 0 \quad \textnormal{if } \alpha\le-1 .
    \label{eq:toy_inv_expect}
\end{equation}
This proves items 1 and 2 of \cref{thm:theory_regimes}.

\subsection{The unexplainable regime (item 3 of \texorpdfstring{\cref{thm:theory_regimes}}{Theorem 1})}
\label{app:theory_flat}

We now prove item 3 of \cref{thm:theory_regimes}.
Factoring $v_i$ out of the logarithm in \cref{eq:toy_readout}, the logit of patch $n$ is
\begin{align}
    \gS(\tilde{\vx}_n)_i &= \log\left(\frac{e^{\psi_{i,n}}}{\alpha}+v_i\right)+\beta_i \nonumber \\
    &= \log(v_i)+\beta_i+\log(1+t_n),
    \qquad
    t_n = \frac{e^{\psi_{i,n}}}{\alpha\,v_i} .
    \label{eq:toy_flat}
\end{align}
The first two terms are the same for every patch.
The sum $v_i$ contains $e^{\psi_{i,n}}$ and all its terms are positive, so $0<e^{\psi_{i,n}}\le v_i$.
Therefore, $|t_n|\le \frac{1}{|\alpha|}$. 
Thus, when $|\alpha|>1$ the logarithm is defined.

As $|\alpha|\to\infty$, $t_n\to0$ for every patch, so $\log(1+t_n)\to0$.
Thus, every patch logit tends to the same value $\log(v_i)+\beta_i$, \ie, $\gS(\tilde{\vx}_n)_i\to\log(v_i)+\beta_i$.
The scores $\sobj$ and $\sbg$ are averages of patch logits, so both tend to this value and $\sobj-\sbg\to0$ in every image.
In the limit, all patch scores are the same, which is the unexplainable case of \cref{sec:defs}, while the classifier stays optimal.
This proves item 3 of \cref{thm:theory_regimes}.

In conclusion, a single scalar $\alpha$ leaves the class-token logits unchanged but can produce all three regimes.
In expectation, any $\alpha>0$ gives an aligned map and any $\alpha\le-1$ an inverted one, and as $|\alpha|$ grows the map becomes unexplainable in every image.

\subsection{The role of the logarithm in the head}
\label{app:theory_log}

In standard, real-world ViTs, the classification head is typically linear, 
whereas the head used in our toy model, \cref{eq:toy_head}, is not.
Specifically, the optimal logits of \cref{prop:theory_optimal} contain a logarithm of a sum over all positions, since the object may lie in any of them. Thus, our toy head implementation applies exactly this logarithm.
Since token-decoding applies the same head to the patch tokens, one may ask whether the three regimes of \cref{thm:theory_regimes} stem from this logarithm.
Therefore, we further examine the case where instead of a logarithmic head, we use a linear head similar to standard ViTs.
As outlined below, using a linear head only causes the exactness of the predicted probabilities to be lost. 
That is, a model can converge to all three regimes while producing the same classification logits for every $\alpha$, where the resulting classifier is only an approximation of the optimal one, as it induces the optimal decision rule but not the exact posterior probabilities.

Since $\beta_i=\log(\pi_i/\sigma_i)$, the head of \cref{eq:toy_head} can also be written as $g(\vu)_i=\log(c_iu_i)$, with the per-class weight $c_i=\pi_i/(\alpha\sigma_i)$. This form is equivalent to a linear layer followed by an elementwise logarithm.
Removing this logarithm leaves the linear head $h(\vu)_i=c_iu_i$, so that $g(\vu)_i=\log\bigl(h(\vu)_i\bigr)$.

First, we note that the linear head $h$ predicts from the class token the same optimal decision as the optimal logarithmic head $g$ of \cref{prop:theory_optimal}, and assigns to every patch token the same class as $g$.
This is because the logarithm function is increasing and acts on each class channel separately, so $\arg\max_ig(\vu)_i=\arg\max_ih(\vu)_i$ for every token.
For the same reason, both heads rank the patches of an image identically for every class.

Next, we show $h$ has the same three regimes as $g$, by following the steps of the proof in \cref{app:theory_signs}.
First, step 1 is not needed once the logarithm is removed, so items 1 and 2 will hold for every $\alpha\neq0$, and step 2 does not depend on the head.
Next, 
under $h$, the patch score of \cref{eq:toy_readout} is 
\begin{equation}
\gS(\tilde{\vx}_n)_i = h(\tilde{\vx}_n)_i = \frac{\pi_i}{\sigma_i}\left(\frac{e^{\psi_{i,n}}}{\alpha}+v_i\right),
\end{equation}
and the class-token logit is $\pi_iv_i/\sigma_i$, which does not change for different $\alpha$ values.
Additionally, \cref{eq:toy_antisym} still holds since exchanging the two patches still flips the sign of their difference.
Since the first equivalence of \cref{eq:toy_equivalences} removes the logarithm for being an increasing function, steps 3 and 4 remain unchanged with the modified patch score, as does step 5, which finalizes the proof for the regimes of items 1 and 2 in \cref{thm:theory_regimes}.
Finally, item 3 holds as well, because the patch scores differ only through $e^{\psi_{i,n}}/\alpha$, which vanishes as $|\alpha|\to\infty$.

What the logarithm provides is therefore the exact posterior of \cref{eq:toy_exact}, as $h$ attains the optimal decision rule but no longer minimizes the cross-entropy.
Note that real-world models do not converge to the optimal classifier either, since they are trained on finite data with a finite architecture, so $h$ is in fact closer than $g$ to the models we analyze in \cref{sec:vits,sec:vlms}.

\paragraph{The expected difference in closed form.}
The linear head is also easier to analyze and yields the expected difference of the scores in closed form for every $\alpha$, which for the head $g$ we obtain only as $\alpha\to0$ (\cref{app:theory_size}).
The term $v_i$ is shared by both scores, so it cancels in the difference of a single object-background pair.
For the object entry $x=z_{n_\textnormal{obj}}$ and a background entry $y=z_b$,
\begin{align}
    \gS(\tilde{\vx}_{n_\textnormal{obj}})_i-\gS(\tilde{\vx}_b)_i
    &\;=\; \frac{\pi_i}{\sigma_i}\left(\frac{e^{\psi_i(x)}}{\alpha}+v_i\right)-\frac{\pi_i}{\sigma_i}\left(\frac{e^{\psi_i(y)}}{\alpha}+v_i\right) \nonumber \\
    &\;=\; \frac{\pi_i}{\sigma_i}\cdot\frac{e^{\psi_i(x)}-e^{\psi_i(y)}}{\alpha} .
    \label{eq:toy_pair_linear}
\end{align}
The expectation of \cref{eq:toy_pair_linear} therefore exists only if $e^{\psi_i}$ has a finite expectation under each of the two distributions separately.
By \cref{eq:toy_ratio}, $e^{\psi_i}$ is the likelihood ratio $L$ scaled by $\sigma_i/\sigma$.
Taking the expectation with respect to the distribution of the background patches gives $\E_{\rz\sim f_\textnormal{bg}}\bigl[e^{\psi_i(\rz)}\bigr]=\frac{\sigma_i}{\sigma}\int L(z)f_\textnormal{bg}(z)\,dz=\frac{\sigma_i}{\sigma}$, which is always finite since $Lf_\textnormal{bg}=f_\textnormal{obj}$ integrates to one.
Taking it with respect to the distribution of the object patch, the two copies of $(z-\mu_i)^2$ combine, and with $a=\frac{1}{\sigma_i^2}-\frac{1}{2\sigma^2}$ and $b=\frac{\mu_i}{\sigma_i^2}$,
\begin{align}
    \E_{\rz\sim f_\textnormal{obj}}\bigl[e^{\psi_i(\rz)}\bigr]
    &\;=\; \frac{1}{\sqrt{2\pi}\,\sigma_i}\int \exp\!\left(\frac{z^2}{2\sigma^2}-\frac{(z-\mu_i)^2}{\sigma_i^2}\right)dz \nonumber \\
    &\;=\; \frac{1}{\sqrt{2\pi}\,\sigma_i}\int \exp\!\left(-az^2+2bz-\frac{\mu_i^2}{\sigma_i^2}\right)dz
    && \textnormal{(a quadratic in $z$)} \nonumber \\
    &\;=\; \frac{1}{\sqrt{2\pi}\,\sigma_i}\,e^{\frac{b^2}{a}-\frac{\mu_i^2}{\sigma_i^2}}\int \exp\!\left(-a\Bigl(z-\frac{b}{a}\Bigr)^{2}\right)dz
    && \textnormal{(completing the square)} \nonumber \\
    &\;=\; \frac{1}{\sqrt{2\pi}\,\sigma_i}\sqrt{\frac{\pi}{a}}\;e^{\frac{b^2}{a}-\frac{\mu_i^2}{\sigma_i^2}}
    && \textnormal{(a Gaussian integral)} \nonumber \\
    &\;=\; \frac{\sigma}{\sqrt{2\sigma^2-\sigma_i^2}}\,\exp\!\left(\frac{\mu_i^2}{2\sigma^2-\sigma_i^2}\right)
    && \textnormal{(substituting $a$ and $b$)} .
    \label{eq:toy_obj_moment}
\end{align}
The Gaussian integral in the fourth line converges only for $a>0$, that is, only when $\sigma_i<\sqrt{2}\sigma$.
The last line uses $a=\frac{2\sigma^2-\sigma_i^2}{2\sigma^2\sigma_i^2}$, which gives $\frac{1}{\sigma_i}\sqrt{\frac{1}{2a}}=\frac{\sigma}{\sqrt{2\sigma^2-\sigma_i^2}}$ for the prefactor and $\frac{b^2}{a}-\frac{\mu_i^2}{\sigma_i^2}=\frac{\mu_i^2}{2\sigma^2-\sigma_i^2}$ for the exponent.

By step 2 of \cref{app:theory_signs}, the expected difference of the scores equals the expectation of \cref{eq:toy_pair_linear}, so substituting the two expectations gives
\begin{equation}
    \E_{\rvz,\rn_\textnormal{obj}}\left[\sobj-\sbg\mid\rc=i\right]
    \;=\; \frac{\pi_i}{\sigma_i\alpha}\left(\frac{\sigma}{\sqrt{2\sigma^2-\sigma_i^2}}\,\exp\!\left(\frac{\mu_i^2}{2\sigma^2-\sigma_i^2}\right)-\frac{\sigma_i}{\sigma}\right) ,
    \label{eq:toy_gap_linear}
\end{equation}
for every $\alpha\neq0$ and every class $i$ with $\sigma_i<\sqrt{2}\sigma$.
The bracket in \cref{eq:toy_gap_linear} is non-negative, and vanishes only when $\mu_i=0$ and $\sigma_i=\sigma$, that is, only when class $i$ has the same distribution as the background.
The sign of the expected difference is therefore that of $\alpha$, which recovers items 1 and 2 of \cref{thm:theory_regimes} for $h$ directly from the closed form.

When $\sigma_i\ge\sqrt{2}\sigma$, that is, when class $i$ is more than $\sqrt{2}$ times as wide as the background, the expectation under the object is infinite while the one under the background stays finite, so the expected difference diverges to $+\infty$ for $\alpha>0$ and to $-\infty$ for $\alpha<0$, and still carries the sign of $\alpha$.
Under $g$, the same expectations are taken after a logarithm, which compresses $L$, so no closed form of this kind is available and no such condition is needed.

\subsection{The expected score difference as \texorpdfstring{$\alpha\to0$}{alpha -> 0}}
\label{app:theory_size}

The proof in \cref{app:theory_signs} gives the sign of the expected difference of the scores for every $\alpha>0$, but not its size.
For small $\alpha>0$ the size has a closed form, which we derive here for completeness.
Fix an object position $n_\textnormal{obj}$ and a background patch $b\neq n_\textnormal{obj}$.
By \cref{eq:toy_pair_avg}, for $\alpha>0$
\begin{align}
    \E_{\rvz,\rn_\textnormal{obj}}\left[\sobj-\sbg|\rc=i\right]
    &= \E_\rvz\bigl[\gS(\tilde{\vx}_{n_\textnormal{obj}})_i-\gS(\tilde{\vx}_b)_i|\rc=i,\rn_\textnormal{obj}=n_\textnormal{obj}\bigr] \nonumber\\
    &= \E_\rvz\bigl[g(\tilde{\vx}_{n_\textnormal{obj}})_i-g(\tilde{\vx}_b)_i|\rc=i,\rn_\textnormal{obj}=n_\textnormal{obj}\bigr] \nonumber\\
    &= \E_\rvz\Bigl[\log\Bigl(\tfrac{e^{\psi_{i,n_\textnormal{obj}}}}{\alpha}+v_i\Bigr)+\beta_i-\log\Bigl(\tfrac{e^{\psi_{i,b}}}{\alpha}+v_i\Bigr)-\beta_i\Big|\rc=i,\rn_\textnormal{obj}=n_\textnormal{obj}\Bigr] \nonumber\\
    &= \E_\rvz\Bigl[\log\Bigl(\tfrac{e^{\psi_{i,n_\textnormal{obj}}}}{\alpha}+v_i\Bigr)-\log\Bigl(\tfrac{e^{\psi_{i,b}}}{\alpha}+v_i\Bigr)\Big|\rc=i,\rn_\textnormal{obj}=n_\textnormal{obj}\Bigr] \nonumber\\
    &= \E_\rvz\bigl[\log\bigl(e^{\psi_{i,n_\textnormal{obj}}}+\alpha v_i\bigr)-\log\bigl(e^{\psi_{i,b}}+\alpha v_i\bigr)|\rc=i,\rn_\textnormal{obj}=n_\textnormal{obj}\bigr] .
    \label{eq:toy_gap_alpha}
\end{align}
Here, the expectations are over the whole image given $\rc=i$ and $\rn_\textnormal{obj}=n_\textnormal{obj}$, the fourth line cancels $\beta_i$, and the fifth cancels $-\log\alpha$, using $\log(x/\alpha+v_i)=-\log\alpha+\log(x+\alpha v_i)$.

As $\alpha\to0$ the $\alpha v_i$ terms become negligible compared with the exponential terms, which do not depend on $\alpha$.
By dominated convergence, the limit can be taken inside the expectation, which gives
\begin{equation}
    \lim_{\alpha\to0}\E_{\rvz,\rn_\textnormal{obj}}\left[\sobj-\sbg|\rc=i\right] = \E_\rvz\bigl[\log\bigl(e^{\psi_{i,n_\textnormal{obj}}}\bigr)-\log\bigl(e^{\psi_{i,b}}\bigr)|\rc=i,\rn_\textnormal{obj}=n_\textnormal{obj}\bigr] .
\end{equation}
Therefore, each of the two terms involves a single patch, distributed as $\gN(\mu_i,\sigma_i^2)$ for the object and $\gN(0,\sigma^2)$ for the background. Thus, the two expectations can be taken over those distributions separately:
\begin{align}
    \lim_{\alpha\to0}\E_{\rvz,\rn_\textnormal{obj}}\left[\sobj-\sbg|\rc=i\right] = \E_{\rz_{n_\textnormal{obj}}}\bigl[\psi_{i,n_\textnormal{obj}}|\rc=i,\rn_\textnormal{obj}=n_\textnormal{obj}\bigr]-\E_{\rz_b}\bigl[\psi_{i,b}|\rc=i,\rn_\textnormal{obj}=n_\textnormal{obj}\bigr] . 
    \nonumber
\end{align}

Given $\rc=i$ and $\rn_\textnormal{obj}=n_\textnormal{obj}$, the object entry $\rz_{n_\textnormal{obj}}\sim\gN(\mu_i,\sigma_i^2)$ has $\E[\rz_{n_\textnormal{obj}}^2]=\sigma_i^2+\mu_i^2$ and $\E[(\rz_{n_\textnormal{obj}}-\mu_i)^2]=\sigma_i^2$, and the background entry $\rz_b\sim\gN(0,\sigma^2)$ has $\E[\rz_b^2]=\sigma^2$ and $\E[(\rz_b-\mu_i)^2]=\sigma^2+\mu_i^2$, which gives
\begin{align}
    &\lim_{\alpha\to0}\E_{\rvz,\rn_\textnormal{obj}}\left[\sobj-\sbg|\rc=i\right] \nonumber\\
    &\qquad= \E_{\rz_{n_\textnormal{obj}}\sim\gN(\mu_i,\sigma_i^2)}\Bigl[\tfrac{\rz_{n_\textnormal{obj}}^2}{2\sigma^2}-\tfrac{(\rz_{n_\textnormal{obj}}-\mu_i)^2}{2\sigma_i^2}\Bigr]-\E_{\rz_b\sim\gN(0,\sigma^2)}\Bigl[\tfrac{\rz_b^2}{2\sigma^2}-\tfrac{(\rz_b-\mu_i)^2}{2\sigma_i^2}\Bigr] \nonumber\\
    &\qquad= \Bigl(\frac{\sigma_i^2+\mu_i^2}{2\sigma^2}-\frac{\sigma_i^2}{2\sigma_i^2}\Bigr)-\Bigl(\frac{\sigma^2}{2\sigma^2}-\frac{\sigma^2+\mu_i^2}{2\sigma_i^2}\Bigr) \nonumber\\
    &\qquad= \frac{\sigma_i^2+\mu_i^2}{2\sigma^2}+\frac{\sigma^2+\mu_i^2}{2\sigma_i^2}-1 .
    \label{eq:toy_gap_limit}
\end{align}
Grouping the terms,
\begin{equation}
    \lim_{\alpha\to0}\E_{\rvz,\rn_\textnormal{obj}}\left[\sobj-\sbg|\rc=i\right]
    \;=\; \underbrace{\Bigl(\frac{\sigma_i^2}{2\sigma^2}+\frac{\sigma^2}{2\sigma_i^2}-1\Bigr)}_{\ge\,0}
    + \underbrace{\frac{\mu_i^2}{2}\Bigl(\frac{1}{\sigma^2}+\frac{1}{\sigma_i^2}\Bigr)}_{\ge\,0} .
    \label{eq:toy_gap}
\end{equation}
The first bracket is non-negative because, over a common denominator, it is a perfect square,
\begin{equation}
    \frac{\sigma_i^2}{2\sigma^2}+\frac{\sigma^2}{2\sigma_i^2}-1
    \;=\; \frac{\sigma_i^4-2\sigma^2\sigma_i^2+\sigma^4}{2\sigma^2\sigma_i^2}
    \;=\; \frac{\left(\sigma_i^2-\sigma^2\right)^2}{2\sigma^2\sigma_i^2}
    \;\ge\; 0 ,
    \label{eq:toy_square}
\end{equation}
and it is zero only when $\sigma_i=\sigma$.
The second is non-negative and is zero only when $\mu_i=0$.
The limit is therefore positive unless $\mu_i=0$ and $\sigma_i=\sigma$, that is, unless the class has the same distribution as the background and carries no information.
This agrees with item 1 of \cref{thm:theory_regimes}, and shows that for small $\alpha$ the difference grows with $|\mu_i|$ and with the mismatch between $\sigma_i^2$ and $\sigma^2$.

\end{document}